\documentclass[]{alaya}
\usepackage{makecell}
\usepackage{wrapfig}
\usepackage{tabularx}
\usepackage{textcomp}
\usepackage{stfloats}
\usepackage{url}
\usepackage{verbatim}
\usepackage{titlesec}
\usepackage{tocloft}
\usepackage{adjustbox}
\usepackage{multirow}
\usepackage{pifont}
\usepackage[sc]{mathpazo}
\usepackage{tikz}
\usepackage{comment}
\usepackage{amsmath,amssymb}
\usepackage{colortbl}
\usepackage{natbib}
\usepackage{color}
\usepackage{booktabs} 
\usepackage{hyperref}
\usepackage{graphicx}
\usepackage{subcaption}
\RequirePackage{xspace}
\makeatletter
\DeclareRobustCommand\onedot{\futurelet\@let@token\@onedot}
\def\@onedot{\ifx\@let@token.\else.\null\fi\xspace}
\usepackage[most]{tcolorbox}
\usepackage{array}
\usepackage{siunitx}
\usepackage[table]{xcolor}
\usepackage{caption}
\definecolor{headerpurple}{HTML}{d8d2fc}
\definecolor{rowgray}{gray}{0.95}
\usepackage{CJKutf8}

\makeatother

\definecolor{adptorange}{RGB}{248, 205, 172}
\definecolor{cmpblue}{RGB}{189, 215, 238}

\definecolor{our_red}{RGB}{232,157,160}
\definecolor{our_blue}{RGB}{136,206,230}
\definecolor{our_orange}{RGB}{246,200,168}
\definecolor{our_green}{RGB}{178,211,164}

\definecolor{attn_code0}{RGB}{247,215,200}
\definecolor{attn_code1}{RGB}{238,169,139}
\definecolor{mlp_code0}{RGB}{204,201,221}
\definecolor{mlp_code1}{RGB}{102,95,153}
\definecolor{mygray}{HTML}{f0f0f0}

\definecolor{token_blue}{RGB}{84, 120, 140}

\usepackage{bbding}
\usepackage{fontawesome}
\usepackage{float}

\newcommand{\cmark}{\textcolor{green!70!black}{\ding{51}}}
\newcommand{\xmark}{\textcolor{red!70!black}{\ding{55}}}
\newcommand{\pmark}{\textcolor{orange!80!black}{$\boldsymbol{\sim}$}}

\newlength\savewidth\newcommand\shline{\noalign{\global\savewidth\arrayrulewidth \global\arrayrulewidth 1pt}\hline\noalign{\global\arrayrulewidth\savewidth}}
\newcommand{\tablestyle}[2]{\setlength{\tabcolsep}{#1}\renewcommand{\arraystretch}{#2}\centering\footnotesize}

\newcolumntype{x}[1]{>{\centering\arraybackslash}p{#1pt}}
\newcolumntype{y}[1]{>{\raggedright\arraybackslash}p{#1pt}}
\newcolumntype{z}[1]{>{\raggedleft\arraybackslash}p{#1pt}}

\renewcommand{\paragraph}[1]{\vspace{1.25mm}\noindent\textbf{#1}}

\usepackage{algorithm}
\usepackage{listings}

\definecolor{codeblue}{rgb}{0.25, 0.5, 0.5}
\definecolor{codekw}{rgb}{0.35, 0.35, 0.75}
\lstdefinestyle{Pytorch}{
    language = Python,
    backgroundcolor = \color{white},
    basicstyle = \fontsize{9pt}{8pt}\selectfont\ttfamily\bfseries,
    columns = fullflexible,
    aboveskip=1pt,
    belowskip=1pt,
    breaklines = true,
    captionpos = b,
    commentstyle = \color{codeblue},
    keywordstyle = \color{codekw},
}

\definecolor{green}{HTML}{009000}
\definecolor{red}{HTML}{ea4335}

\newcommand{\dataset}{Sekai2\xspace}

\title{\dataset: From World Exploration to Interactive World Modeling}
\author[1,2,3]{Kang He}
\author[4]{Wenshuo Peng}
\author[1]{Zihui Gao}
\author[1]{Jiaming Tan}
\author[1,*]{Kaipeng Zhang}
\author[1,*]{Yongtao Ge}

\affiliation[1]{Alaya Lab}
\affiliation[2]{Shanghai Innovation Institute}
\affiliation[3]{Wuhan University}
\affiliation[4]{Tsinghua University}

\abstract{

Video world models must capture how scenes evolve over time and across viewpoints. Training them for long-horizon generation and camera control therefore benefits from long videos paired with camera trajectories and temporally grounded semantics. Existing corpora rarely offer the three together: large-scale web video provides broad visual diversity but no trajectories or time-aligned text, while pose-annotated datasets are typically short-range or reconstruction-oriented.
We introduce \dataset, a multi-source real-world video dataset that carries the world-exploration footage of Sekai toward interactive world modeling. The release contains $128{,}892$ clips totaling $2{,}826$ hours from $10{,}428$ source videos across $113$ countries or regions, and is deliberately weighted toward sustained observation: under a common $120$-second decomposition, $43{,}594$ segments reach the full two minutes and account for $51.4\%$ of all footage.
Every clip includes a released camera trajectory and hierarchical annotations disentangling subject motion, environment dynamics, static scene content, and camera behavior, resulting in $649{,}597$ temporally grounded segments. Crucially, we further introduce $982$ panoramic sequences captured along non-linear trajectories with loops and revisits. These revisits provide repeated observations of the same locations across time and viewpoints, offering essential supervision for learning persistent scene representations, long-term spatial memory, and geometrically consistent world models.
Corpus-scale analyses demonstrate complete pose-and-caption coverage, broad geographic and semantic diversity, varied camera trajectories, and highly non-redundant temporal descriptions. Together, these properties make \dataset a scalable resource for long-horizon video generation, camera-controllable synthesis, and interactive world-model pre-training.
}

\github{\url{https://kangverse.github.io/sekai2-project/}} 
\Code{\url{https://github.com/kangverse/Sekai2-Dataset}} 
\Corresponding{kaipeng.zhang@shanda.com,yongtao.ge@shanda.com}
\date{\today}

\begin{document}
\maketitle

\begin{figure}[!h]
    \centering
    \includegraphics[width=\linewidth]{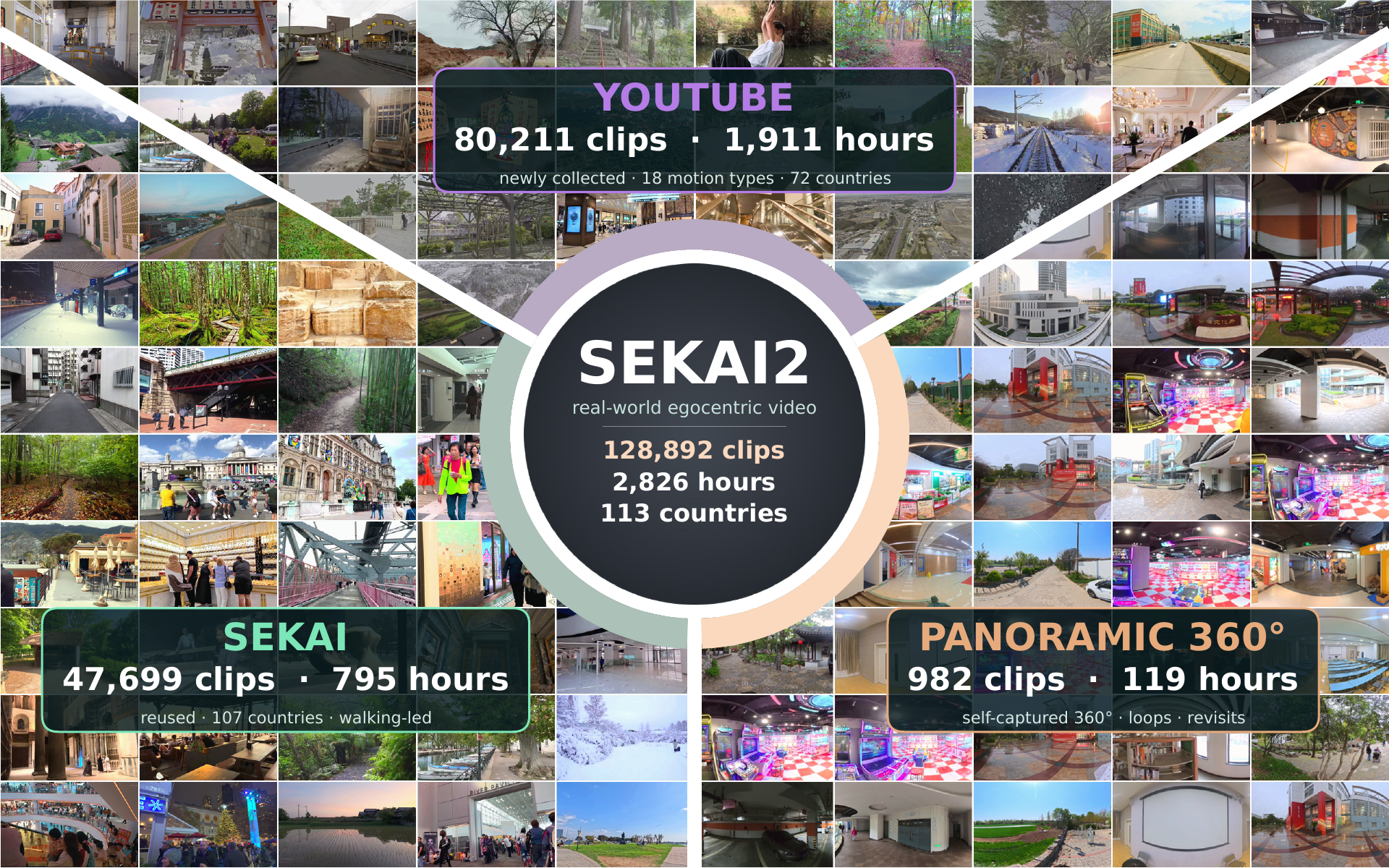}\vspace{5pt}
    \captionof{figure}{\textbf{Composition and visual coverage of \dataset.}
    Each tile shows a frame from a released clip, grouped into three source sectors: the newly collected YouTube subset ($80{,}211$ clips; $1{,}912$\,h), the inherited Sekai subset ($47{,}699$ clips; $795$\,h), and self-captured panoramic videos ($982$ clips; $119$\,h). Panoramic tiles display the horizon band of the native equirectangular frame. The resulting corpus comprises $128{,}892$ clips and  $2{,}826$ hours across $113$ countries or regions. Every clip is paired with a released camera trajectory and structured global and temporally grounded annotations.}
    \label{fig:teaser}
\end{figure}

\section{Introduction}
\label{sec:intro}

Video generation is rapidly evolving from synthesizing short visual clips toward building \emph{interactive world models}: generative systems that maintain persistent scene states, produce coherent observations over extended horizons, and respond to changes in viewpoint or user control~\cite{bruce2024genie,ball2025genie3,gao2026infinite,mao2026yume1}. This shift places stronger demands on training data. Beyond modeling local appearance and short-term motion, long-horizon generation requires models to preserve world state as the observer moves, distinguish environmental changes from viewpoint changes, and maintain spatial consistency over extended trajectories. Training data for interactive world modeling therefore benefits from the joint availability of \emph{temporal continuity}, \emph{camera-pose supervision}, and \emph{temporally localized semantics}.

Existing video datasets provide only part of this supervision. Large-scale video--text corpora such as WebVid~\cite{bain2021frozen}, Panda-70M~\cite{chen2024panda}, OpenVid-1M~\cite{nan2025openvid}, and MiraData~\cite{ju2024miradata} offer broad visual and semantic diversity, but generally lack explicit camera trajectories and temporally aligned descriptions. Spatial datasets such as RealEstate10K~\cite{zhou2018stereo}, DL3DV-10K~\cite{ling2024dl3dv}, SpatialVID~\cite{wang2026spatialvid}, and OmniWorld~\cite{zhou2025omniworld} provide camera poses, depth, or other spatial supervision, yet are largely designed for reconstruction, view synthesis, or short-range camera control rather than sustained world evolution over minute-scale trajectories. Action-centric datasets and interactive game pipelines provide richer control signals~\cite{Chen_2026_CVPR,che2025gamegen,he2025matrix,wang2026matrix}, but often focus on local actions or are predominantly derived from game and simulated environments. Consequently, \textbf{long-horizon real-world video with both explicit camera trajectories and temporally grounded semantics remains scarce.}

A further challenge emerges when long-horizon trajectories revisit previously observed locations. During open-ended exploration, an observer may leave a scene, follow a non-linear route, and later return from a different direction or after a long temporal interval. Such revisits directly expose whether a world model preserves previously observed scene structure rather than relying only on recent visual context. However, controlled revisit trajectories are uncommon in ordinary web video, providing little direct supervision for studying long-range revisit consistency. We therefore argue that real-world data for interactive world modeling should be represented not merely as independent video--text pairs, but as \emph{temporally evolving observations along continuous trajectories}, with visual content, camera motion, and localized semantic changes aligned on a shared timeline, and with revisit-rich trajectories explicitly included where possible.

Sekai~\cite{li2026sekai} took an important step in this direction by providing geographically diverse first-person and aerial videos for real-world exploration. Building on this foundation, we introduce \dataset to move from \emph{world exploration} toward \emph{\textbf{interactive world modeling}}. This extension is not a simple addition of new data sources: we re-curate Sekai videos and process them together with newly collected videos under a unified quality-control, trajectory-processing, and hierarchical annotation pipeline. This places heterogeneous videos under a consistent supervision scheme that aligns scene content, local dynamics, and viewpoint changes over time. Importantly, we further introduce a \emph{revisit-rich panoramic subset} deliberately captured along long non-linear routes containing loops and repeated visits to previously observed locations. Panoramic capture preserves surrounding visual context throughout these trajectories, while the resulting revisit structure provides long-range observations that are difficult to obtain reliably from ordinary web video.

The resulting \dataset release contains $128{,}892$ clips totaling \textbf{2,826 hours}, collected from $10{,}428$ source videos across $113$ countries or regions. Long temporal continuity is a defining property of the corpus: under a common $120$-second analysis decomposition, \textbf{43,594 segments reach the full two-minute duration}, contributing \textbf{1,453 hours} and accounting for $51.4\%$ of the total footage, while segments of at least one minute cover $92.3\%$ of the total duration. Every released video is paired with a camera-pose trajectory and structured semantic annotations, yielding $649{,}597$ temporally grounded segments that separately characterize \emph{\textbf{subject motion}}, \emph{\textbf{environmental dynamics}}, \emph{\textbf{static scene content}}, and \emph{\textbf{camera behavior}}. These contiguous temporal units can be aggregated according to the target training duration, providing multi-scale supervision under a unified annotation schema. Of these clips, $982$ are panoramic sequences totaling $119$ hours, preserving extended loops, non-linear routes, and revisits within complete trajectories.

To construct \dataset at scale, we develop a unified data engine that integrates long-video construction, quality filtering, camera-trajectory processing, and hierarchical semantic annotation. For revisit-rich panoramic sequences, geometrically verified loop closures are further incorporated to reduce accumulated trajectory drift while preserving repeated observations of the same locations. We conduct corpus-scale validation of visual quality, temporal dynamics, camera trajectories, semantic annotations, and cross-modal alignment, establishing \dataset as a scalable real-world data foundation for long-horizon video generation, camera-controllable synthesis, and interactive world-model pre-training.

Our contributions are summarized as follows:
\begin{itemize}

\item We introduce \dataset, a real-world video dataset that jointly provides long-horizon temporal continuity, explicit camera trajectories, and fine-grained temporally grounded semantics across $2{,}826$ hours of geographically diverse video.

\item We provide a unified hierarchical annotation schema that factorizes persistent scene content, subject and environmental dynamics, and camera behavior, together with a dedicated revisit-rich panoramic subset containing long non-linear trajectories, loops, and repeated observations for studying long-range spatial consistency.

\item We develop a reproducible multi-source data engine with trajectory-quality control and panoramic loop-closure refinement, and conduct corpus-scale validation of visual quality, camera trajectories, semantic annotations, and cross-modal alignment, providing a reliable data foundation for camera-controllable generation and interactive world modeling.

\end{itemize}

\section{Related Work}
\label{sec:related}

\subsection{Video Datasets for World Modeling}

Large-scale video--text datasets provide the data foundation for modern video generation. WebVid~\cite{bain2021frozen}, InternVid~\cite{wang2024internvid}, Panda-70M~\cite{chen2024panda}, OpenVid-1M~\cite{nan2025openvid}, and VidGen-1M~\cite{tan2024vidgen} substantially scale open-domain video--text pairs, while ShareGPT4Video~\cite{chen2024sharegpt4video}, Vript~\cite{yang2024vript}, and MiraData~\cite{ju2024miradata} improve caption detail, structure, and temporal extent. Their supervision, however, remains predominantly semantic, typically without explicit camera trajectories or temporally localized descriptions aligned with scene and viewpoint changes.

Spatial video datasets provide complementary geometric supervision. RealEstate10K~\cite{zhou2018stereo}, DL3DV-10K~\cite{ling2024dl3dv}, TartanAir~\cite{wang2020tartanair}, and SpatialVID~\cite{wang2026spatialvid} provide camera poses, depth, or other spatial annotations, while OmniWorld~\cite{zhou2025omniworld} further integrates multi-domain 4D data. In particular, SpatialVID scales real-world dynamic video with camera poses, depth, structured captions, and motion instructions, but this line of work primarily emphasizes spatial reconstruction, view synthesis, and short-range camera control rather than sustained world evolution over minute-scale trajectories. Action100M~\cite{Chen_2026_CVPR}, VideoEspresso~\cite{han2025videoespresso}, and Leader360V~\cite{zhang2026leader360v} enrich temporal action, video reasoning, and panoramic perception supervision, respectively, but do not jointly provide long-form real-world video, camera trajectories, and fine-grained temporally grounded semantics.

Sekai~\cite{li2026sekai} is the closest predecessor to our work, providing geographically diverse first-person and aerial exploration videos with rich metadata and camera trajectories. \dataset builds on this foundation but reorganizes the data for long-horizon world modeling: selected Sekai videos are reprocessed together with newly collected data under a unified quality-control, trajectory-processing, and hierarchical annotation pipeline. Beyond video-level exploration supervision, \dataset pairs released videos with camera trajectories and dense temporal annotations that explicitly localize scene dynamics and viewpoint changes.

\subsection{Long-Horizon Interactive World Models}

Interactive world models predict future observations conditioned on visual history and user or agent controls. UniSim~\cite{yang2023learning} and GAIA-1~\cite{hu2023gaia} study action-conditioned simulation in robotics and driving, while GameNGen~\cite{valevski2025diffusion}, Oasis~\cite{decart2024oasis}, MineWorld~\cite{guo2025mineworld}, and related systems~\cite{menapace2024promptable,kanervisto2025world} demonstrate controllable generation in game environments. Genie~\cite{bruce2024genie} and its successors~\cite{parker2024genie,ball2025genie3} explore latent actions and promptable world generation, while recent systems further extend interaction horizons, camera control, and real-time generation~\cite{he2025matrix,wang2026matrix,che2025gamegen,li2025hunyuan,tang2025hunyuan,mao2026yume1,hyworld2025,team2026advancing,gao2026infinite}.

As interaction horizons grow, maintaining memory of previously observed environments becomes increasingly important. WorldMem~\cite{xiao2026worldmem} introduces memory-based world simulation to recover previously observed scenes across large temporal and viewpoint gaps, while Video World Models with Long-term Spatial Memory~\cite{wu2026video} explicitly identifies scene forgetting during revisits as a major failure mode of autoregressive world models. More recently, Infinite-World~\cite{wu2026infinite} highlights the scarcity of viewpoint revisits in natural real-world videos and employs revisit-dense fine-tuning to improve long-range loop-closure behavior. PanoWorld~\cite{li2026panoworld} similarly explores panoramic representations for long-range memory and spatial consistency. These studies demonstrate that revisit consistency is becoming a central challenge for long-horizon world modeling.

Despite this progress, existing work remains largely model-centric, and the corresponding training data are often proprietary, simulation-based, or specifically curated for individual systems. Controlled revisits are particularly uncommon in ordinary web videos. \dataset complements these approaches with long-horizon real-world trajectories and a dedicated revisit-rich panoramic subset containing non-linear routes, loops, and repeated observations, providing reusable data for studying long-range spatial consistency.

\subsection{Camera and Temporally Grounded Video Supervision}

Controllable video generation requires separating changes in the world from changes in viewpoint. Camera-conditioned methods represent control through camera poses, rays, Pl"ucker coordinates, or positional embeddings~\cite{he2024cameractrl,wang2024motionctrl,he2025cameractrl,xu2024camco,ren2025gen3c,wang2026bullettime,yu2025trajectorycrafter}, demonstrating the importance of reliable camera-trajectory supervision. In parallel, video annotation has progressed from single global captions toward structured and temporally grounded descriptions~\cite{chen2024sharegpt4video,yang2024vript,ju2024miradata,Chen_2026_CVPR}. However, camera and semantic supervision are commonly constructed independently or provided at different temporal granularities, making it difficult to associate what changes in a video with when it changes and whether that change originates from the scene or the camera.

\dataset organizes camera trajectories and semantic supervision on a shared timeline. Each released video is paired with a camera-pose trajectory and is annotated at both the overall and segment levels. The two levels share the same six textual fields: \emph{subject motion}, \emph{environment motion}, \emph{static scene}, \emph{camera description}, together with the training-oriented \texttt{full\_prompt} and \texttt{short\_prompt}. The \texttt{full\_prompt} integrates scene content, motion, and camera information, whereas the \texttt{short\_prompt} removes camera-specific descriptions and retains a compact scene-centric summary. Segment-level annotations further provide explicit temporal boundaries and a discrete \texttt{camera\_path} label, aligning scene content, subject and environmental motion, and viewpoint changes with specific temporal intervals. This hierarchical design provides unified, multi-granularity supervision for long-horizon video generation and camera-controllable world modeling.

\section{Dataset Construction}
\label{sec:dataset}

\subsection{Design Principles}
\label{sec:dataset_design}

\dataset is constructed for long-horizon world modeling from real-world video, following four design principles: \emph{temporal continuity}, such that each sample captures sustained evolution rather than isolated events; \emph{world coverage}, spanning diverse locations, scenes, viewpoints, and environmental conditions; \emph{camera supervision}, with a trajectory provided for every released clip; and \emph{temporally grounded semantics}, which distinguish persistent scene content, world dynamics, and viewpoint changes over time. Accordingly, our data engine comprises source acquisition, shot-aware long-clip construction, multi-stage filtering, camera-pose estimation, and hierarchical semantic annotation. Detailed algorithms, thresholds, and prompts are provided in Appendix~\ref{app:dataset_details}.

\subsection{Data Acquisition}
\label{sec:dataset_sources}

We assemble the source pool from three complementary channels.

\paragraph{Inherited Sekai data.}
We reprocess the original $5{,}000{+}$-hour Sekai collection~\cite{li2026sekai}, whose first-person and aerial videos provide broad coverage of exploration-oriented scenes. All inherited videos are screened and annotated under the same continuity, visual-quality, and annotation criteria as the newly collected data. Their camera geometry, however, is not reconstructed identically: inherited clips reuse their existing per-frame intrinsics, whereas newly collected clips are calibrated by GeoCalib during reconstruction. The cross-run evaluation in Sec.~\ref{sec:exp_pose} therefore re-estimates intrinsics for every subset, so that source-level pose comparisons are not confounded by this difference.

\paragraph{Newly collected YouTube videos.}
We collect walking tours, indoor and outdoor traversal, driving, rail travel, drone footage, and FPV videos from YouTube. The initial crawl contains $5{,}199$ source videos. Source-level matching against Sekai identifies $110$ overlaps, leaving $5{,}089$ newly collected videos. After shot-aware decomposition but before downstream technical filtering, these videos yield $173{,}303$ candidate clips totaling $3{,}957.4$ hours.

\paragraph{Revisit-rich panoramic captures.}
We additionally record $1{,}283$ panoramic videos, totaling approximately $160$ hours, using Insta360 cameras across urban streets, residential communities, campuses, parks, scenic areas, and indoor spaces. These recordings preserve complete azimuthal observations along non-linear walking routes containing sharp turns, revisits, and loop-like trajectories, thereby capturing off-axis context that is typically absent from conventional perspective video.
Unlike ordinary web videos, these sequences are deliberately captured along non-linear routes containing loops and repeated visits to previously observed locations, providing long-range trajectory structures for studying revisit consistency.

Together the three channels supply more than $9{,}100$ hours of candidate footage: the $5{,}000{+}$-hour inherited collection, the $3{,}957.4$ hours of newly crawled clips obtained after shot-aware decomposition, and $160$ hours of panoramic capture. Only the middle term is already segmented; all three precede downstream curation and quality filtering.

\subsection{Long-Clip Curation and Filtering}
\label{sec:dataset_curation}

Editing transitions introduce discontinuous state changes that may be incorrectly learned as physical dynamics. We therefore use OmniShotCut~\cite{wang2026omnishotcut} to detect hard cuts and gradual transitions, remove a margin around internal boundaries, and partition continuous perspective shots into clips of at most $120$ seconds. Short residual segments are discarded, and stream copy is used whenever possible to avoid re-encoding. Self-collected panoramic videos are retained as complete sequences to preserve their long-range loop and revisit structure.

We then apply a manifest-driven filtering pipeline. Optical-flow statistics identify temporally degenerate clips, while media-quality and OCR checks remove malformed encodings, severe exposure failures, black borders, persistent HUD elements, subtitles, watermarks, and large on-screen text. Camera-pose quality and semantic-annotation validity are evaluated as independent gates, and only samples passing all required checks are retained. Because the motion gate alone would remove stationary observation entirely, clips rejected by it but passing every other check enter a separately curated low-motion branch. It contributes $5{,}608$ of the $80{,}211$ newly collected clips ($7.0\%$), and we refer to it as the \emph{static supplement} in the trajectory evaluation of Sec.~\ref{sec:exp_pose}.
Stage-wise retention statistics and implementation details are provided in Sec.~\ref{sec:statistics_scale} and Appendix~\ref{app:filtering_framework}, respectively.

\subsection{Camera-Pose Annotation}
\label{sec:dataset_pose}

\paragraph{Scalable pose estimation.}
We use ViPE~\cite{huang2025vipe} to estimate camera intrinsics and per-frame extrinsics. For corpus-scale processing, we adopt a pose-only deployment that preserves ViPE's geometric reconstruction pipeline while omitting dense-depth export and visualization. GeoCalib~\cite{veicht2024geocalib} initializes the camera intrinsics, dynamic foregrounds are masked before DROID-SLAM optimization~\cite{teed2021droid}, and a monocular depth prior is used to improve reconstruction stability. Each released trajectory file contains frame indices and the corresponding camera-to-world transformations, and the estimated intrinsics are released as a companion artifact, so that the two together can be converted directly into ray or Pl\"ucker embeddings for camera-conditioned video generation.

\paragraph{Long-horizon panoramic trajectory refinement.}
Long panoramic sequences are particularly susceptible to accumulated drift. We therefore retrieve loop-closure candidates between temporally distant keyframes and geometrically verify whether their matched spherical rays are consistent with a single relative rotation, filtering out appearance-based matches between distinct locations. Verified closures are incorporated into pose-graph optimization as camera-center constraints, allowing the same location to be revisited under different viewing directions. This procedure refines $1{,}051$ of the $1{,}283$ panoramic trajectories ($81.9\%$); sequences without reliable loop closures retain their original ViPE estimates. Implementation details and qualitative results are provided in Appendix~\ref{sec:panoramic_captures} and Appendix~\ref{app:panoramic_reconstruction_cases}, respectively.

\subsection{Hierarchical Semantic Annotation}
\label{sec:dataset_annotation}

Single-sentence captions often entangle subject motion, environmental dynamics, and camera motion, while providing little indication of when these changes occur. To provide fine-grained supervision for long-horizon world modeling, we use Kimi-K2.6~\cite{team2026kimi} to annotate timestamped video frames at two complementary levels: \textit{clip-level global descriptions} and \textit{interval-level local descriptions}.

At the clip level, structured attributes cover geographic context, scene type, environmental conditions, viewpoint, camera properties, and visual quality. Free-form descriptions further decompose the video into \emph{\textbf{subject motion}}, \emph{\textbf{environment motion}}, \emph{\textbf{static scene content}}, and \emph{\textbf{camera behavior}}, reducing semantic entanglement among different sources of visual change.

At the interval level, temporal boundaries are determined by meaningful changes in scene content, illumination, surface type, or camera behavior, with all intervals jointly covering the full clip. Each interval contains the same structured fields as the global annotation, a discrete \texttt{camera\_path} label, a \texttt{full\_prompt} describing both scene content and camera motion, and a \texttt{short\_prompt} with camera-related information removed. Camera descriptions in the \texttt{full\_prompt} are explicitly enclosed by \texttt{<camera>...</camera>}, enabling camera conditions to be dropped, replaced, or controlled independently during training. 

In total, \dataset contains $649{,}597$ temporally grounded semantic segments, providing dense supervision at the timescale of observable scene changes. The complete annotation schema, controlled vocabularies, prompt design, and validation rules are provided in Appendix~\ref{app:caption_schema}, while a complete video--pose--caption example is shown in Appendix~\ref{app:multimodal_case}.

\begin{table*}[t]
    \centering
    \caption{\textbf{Comparison with representative video datasets.}
    ``Units'' reports released clips unless otherwise specified.
        ``Pose'' denotes publicly available camera trajectories.
    ``Long obs.'' indicates substantial minute-scale continuous observations rather than predominantly short clips.
        ``Temp. sem.'' denotes semantic supervision localized to temporal intervals,
    while ``G+S'' indicates semantic annotations at both global and temporally grounded segment levels.
        ``Revisit'' denotes explicitly collected or annotated loop/revisit trajectories rather than incidental repeated views.
    A partial mark indicates limited, indirect, or subset-level availability.
    For Action100M, the reported scale corresponds to hierarchical temporal segments derived from long source videos; its segments are not disjoint, so an average segment duration is not reported.}
    \label{tab:dataset_comparison}
    \vspace{3pt}
    \tablestyle{3.5pt}{1.10}
    \resizebox{0.99\textwidth}{!}{%
    \begin{tabular}{lrrrrcccccc}
    \shline
    Dataset & Units & Hours & Avg. (s) & Countries
    & Real & Pose & Long obs. & Temp. sem. & G+S & Revisit \\
    \shline

    WebVid~\cite{bain2021frozen}
    & 10.7M & 52K & 18.0 & --
    & \cmark & \xmark & \xmark & \xmark & \xmark & \xmark \\

    Panda-70M~\cite{chen2024panda}
    & 70.8M & 167K & 8.5 & --
    & \cmark & \xmark & \xmark & \xmark & \xmark & \xmark \\

    OpenVid-1M~\cite{nan2025openvid}
    & 1.0M & 2.1K & 7.2 & --
    & \cmark & \xmark & \xmark & \xmark & \xmark & \xmark \\

    MiraData~\cite{ju2024miradata}
    & 330K & -- & 72.1 & --
    & \pmark & \xmark & \cmark & \pmark & \pmark & \xmark \\

    Ego4D~\cite{grauman2022ego4d}
    & -- & 3.7K & -- & 9
    & \cmark & \pmark & \cmark & \cmark & \pmark & \xmark \\

    Action100M~\cite{Chen_2026_CVPR}
    & $\sim$100M seg. & $\sim$128K$^\dagger$ & -- & --
    & \cmark & \xmark & \pmark & \cmark & \cmark & \xmark \\

    SpatialVID~\cite{wang2026spatialvid}
    & 2.7M & 7.1K & 9.5 & --
    & \cmark & \cmark & \xmark & \pmark & \xmark & \xmark \\

    OmniWorld~\cite{zhou2025omniworld}
    & 96K & 214 & 8.0 & --
    & \xmark & \cmark & \xmark & \cmark & \pmark & \xmark \\

    Sekai~\cite{li2026sekai}
    & -- & 5K+ & 60 & 100+
    & \cmark & \pmark & \cmark & \xmark & \xmark & \xmark \\

    \shline
    \rowcolor{mygray}
    \textbf{\dataset (ours)}
    & \textbf{128,892} & \textbf{2,826} & \textbf{78.9} & \textbf{113}
    & \cmark & \cmark & \cmark & \cmark & \cmark & \cmark \\
    \shline
\end{tabular}}
\vspace{2pt}

{\footnotesize
$^\dagger$Action100M reports approximately 14.6 years of underlying instructional video.
Its $\sim$100M temporally localized annotations form a hierarchical, potentially overlapping
segmentation and are therefore not directly comparable with disjoint released clips.}

\end{table*}

\subsection{Dataset Quality Assurance}
\label{sec:dataset_quality}

We ensure dataset quality through both filtering-stage quality gates and systematic post-annotation validation. For camera trajectories, we check validity, degeneracy, discontinuities, local jitter, and long-term drift. For semantic annotations, we validate schema compliance, enumerated fields, required content, temporal coverage, prompt formatting, and duplicated descriptions to remove invalid or inconsistent outputs.

Because camera poses and semantic descriptions are generated through independent pipelines, we further compare pose-derived camera motion with the annotated \texttt{camera\_path} labels and use their agreement as a cross-modal quality diagnostic. The corresponding consistency analysis and trajectory-quality evaluation are reported in Sec.~\ref{sec:exp_grounding} and Sec.~\ref{sec:exp_pose}, respectively, with detailed metrics and decision rules provided in Appendix~\ref{app:quality_assurance}.

\begin{figure*}[t]
    \centering
    \includegraphics[width=\linewidth]{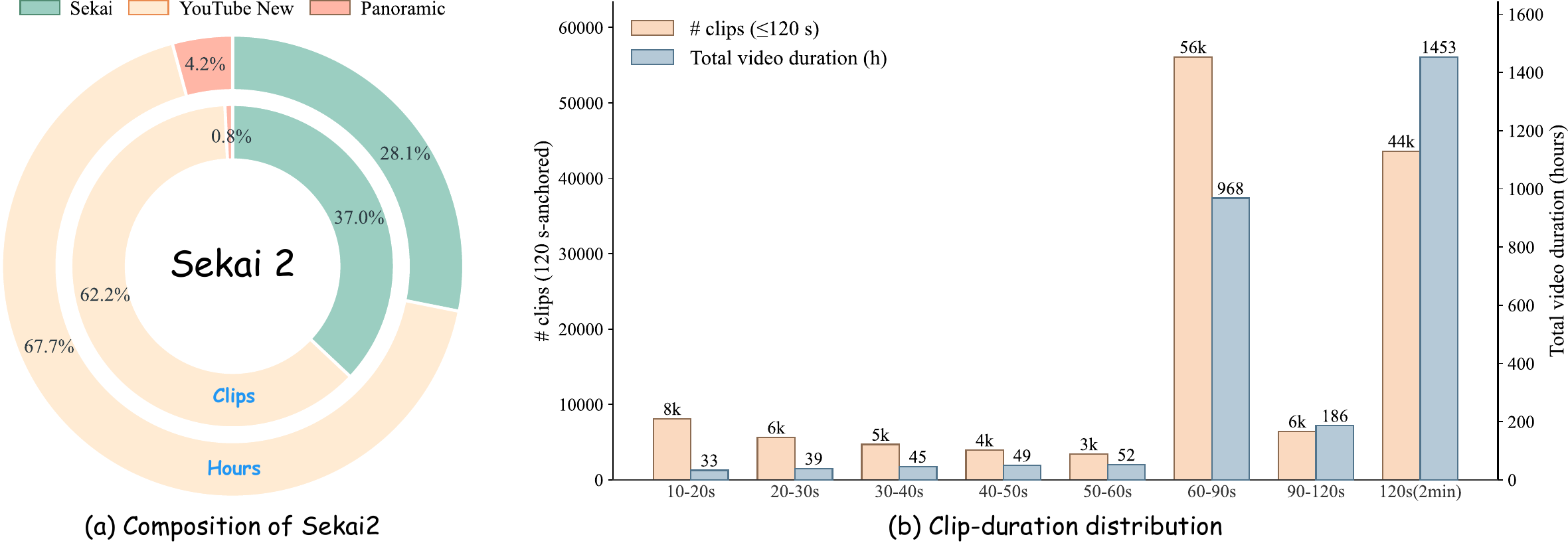}
    \vspace{4pt}
    \caption{\textbf{Scale, composition, and temporal extent of \dataset.}
    (a) Source composition by released clip count (inner ring) and footage
    duration (outer ring). (b) Duration distribution after representing all
    released videos as non-overlapping analysis segments of at most
    $120$\,s. The left and right axes show segment counts and aggregate
    duration, respectively.}
    \label{fig:duration_composition}
\end{figure*}

\section{Dataset Statistics and Analysis}
\label{sec:statistics}

\subsection{Comparison with Existing Datasets}
\label{sec:statistics_comparison}

Table~\ref{tab:dataset_comparison} compares existing datasets along dimensions relevant to interactive world modeling. Large-scale video--text corpora such as WebVid, Panda-70M, and OpenVid-1M offer substantially greater scale, but are dominated by short clips and typically lack explicit camera trajectories and fine-grained temporal semantic supervision. MiraData, Ego4D, and Action100M provide richer long-form or temporally localized information, yet camera-pose supervision and temporally grounded semantics are rarely available together.

Recent spatial video datasets such as SpatialVID and OmniWorld further introduce camera poses, depth, and other spatial supervision, but mainly focus on short video segments or synthetic environments. In contrast, \dataset emphasizes sustained real-world video and long-range trajectory evolution: its average video duration is $78.9$ seconds, and segments of at least $60$ seconds account for $92.3\%$ of the total footage duration. Meanwhile, every released clip is paired with a camera trajectory and temporally grounded semantic annotations.

As the successor to Sekai, \dataset does not simply inherit the original data. Selected Sekai videos are reprocessed through the same unified quality filtering, trajectory processing, and hierarchical semantic annotation pipeline, and are further complemented by newly collected videos and panoramic captures containing loops and revisits. In this way, \dataset extends Sekai from data for world exploration toward \emph{long-form real-world video with camera trajectories and temporally structured annotations}. Its distinguishing strength lies not in scale along any single dimension, but in jointly providing long-horizon continuity, camera trajectories, temporally grounded semantics, and revisit structure within a unified real-world corpus.

\subsection{Scale, Composition, and Temporal Extent}
\label{sec:statistics_scale}

\dataset contains $128{,}892$ released clips totaling $2{,}826$ hours from three complementary sources (Fig.~\ref{fig:duration_composition}a). Specifically, the inherited Sekai subset contains $47{,}699$ one-minute clips totaling $795$ hours, the newly collected perspective subset contains $80{,}211$ clips totaling $1{,}912$ hours, and the self-collected panoramic subset contains $982$ sequences totaling $119$ hours. Accordingly, the three sources contribute $28.1\%$, $67.7\%$, and $4.2\%$ of the total footage duration, respectively. Despite its small share of released clips, the panoramic subset contributes disproportionately by duration because its long, uncut sequences are retained to preserve complete routes, loops, and revisits.

To characterize temporal extent on a common scale, we represent all released videos as non-overlapping analysis segments capped at $120$\,s. Among them, $43{,}594$ segments reach the full $120$-second duration, contributing $1{,}453$ hours and accounting for $51.4\%$ of the corpus duration (Fig.~\ref{fig:duration_composition}b). Segments of at least $60$ seconds collectively contribute $92.3\%$ of all footage, demonstrating that \dataset is dominated by sustained real-world video rather than short isolated clips. Detailed source-wise and duration-bin statistics are provided in Appendix~\ref{app:scale_duration}.

\subsection{Environment and Scene Diversity}
\label{sec:statistics_scene}

\begin{figure*}[t]
    \centering
    \includegraphics[width=0.98\linewidth]{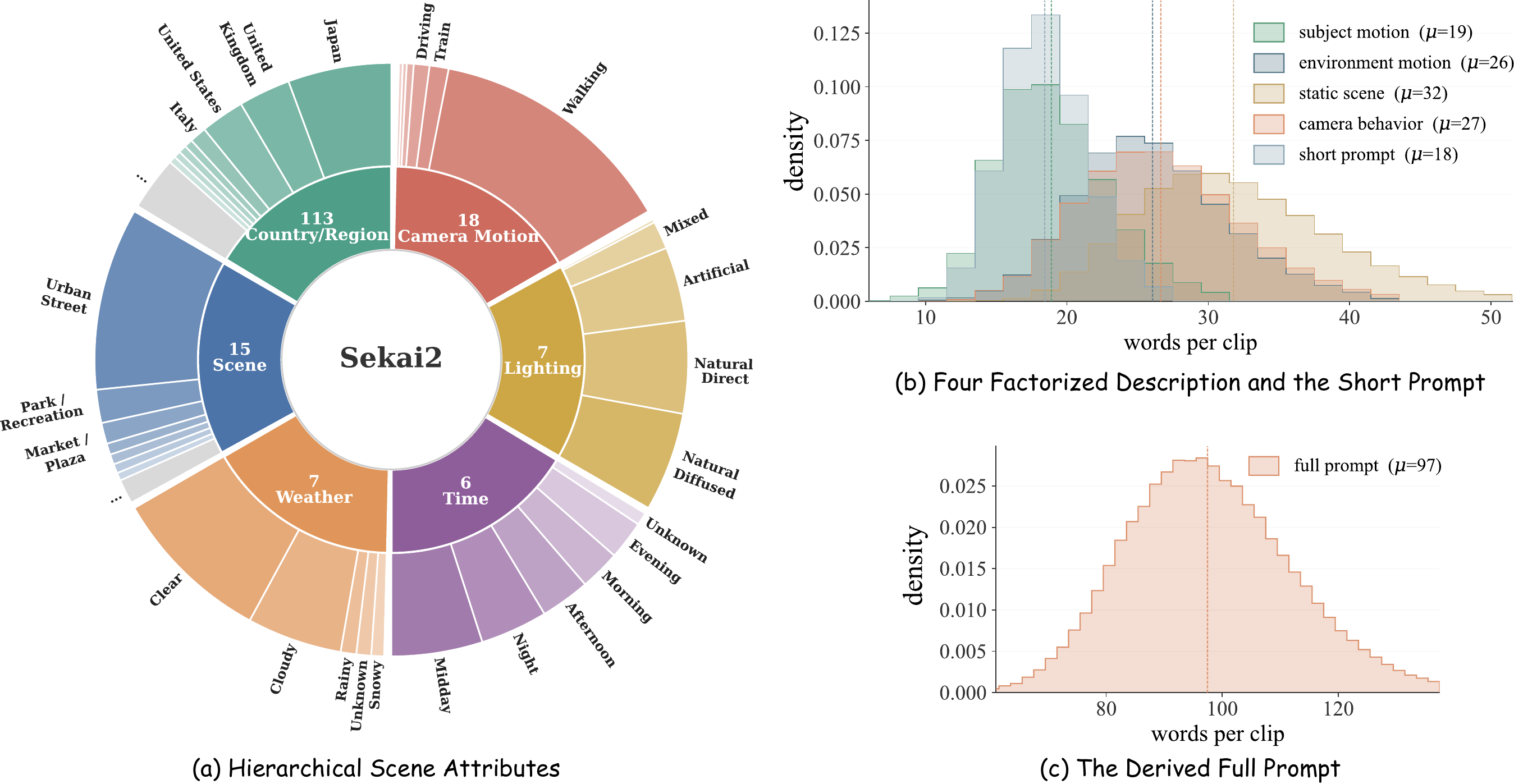}
    \vspace{4pt}
    \caption{\textbf{Scene attributes and caption lengths.} (a) Distribution of six controlled clip-level attributes. The inner ring reports the number of distinct values for each attribute, and the outer ring shows their empirical frequencies. (b) Word-count distributions of the four factorized descriptions and the short prompt. (c) Word-count distribution of the derived full prompt. Dashed lines indicate the means. Panels (b) and (c) are discussed in Sec.~\ref{sec:statistics_caption}.}
    \label{fig:scene_caption}
\end{figure*}

Each clip carries ten controlled clip-level attributes, of which Fig.~\ref{fig:scene_caption}a shows six: country or region, camera motion, lighting, time of day, weather, and scene type, comprising $113$, $18$, $7$, $6$, $7$, and $15$ distinct values, respectively. The remaining four---perspective, video style, camera height, and season---are used in the dependence analysis of Appendix~\ref{app:attribute_dependence}. As expected for exploration-oriented web video, urban streets constitute the largest scene category ($60.9\%$), followed by parks and recreational areas ($11.1\%$), markets and plazas ($7.0\%$), mountains ($3.7\%$), indoor spaces ($3.4\%$), and forests ($3.1\%$). The corpus nevertheless retains substantial variation in environmental conditions: $35.3\%$ of clips are recorded in the evening or at night, $41.5\%$ under non-clear weather, and $34.0\%$ under artificial or mixed lighting.

The newly collected footage further broadens scene coverage beyond the inherited Sekai subset. In particular, the share of urban-street scenes decreases from $70.3\%$ to $56.9\%$, while natural environments---including mountains, forests, parks, grasslands, water, snow, and deserts---increase from $10.8\%$ to $28.7\%$. Detailed distributions, normalized entropy, and cross-attribute dependence are reported in Appendix~\ref{app:attribute_dependence}.

\subsection{Geographic Coverage}
\label{sec:statistics_geo}

\begin{figure}[t]
    \centering
    \includegraphics[width=0.92\linewidth]{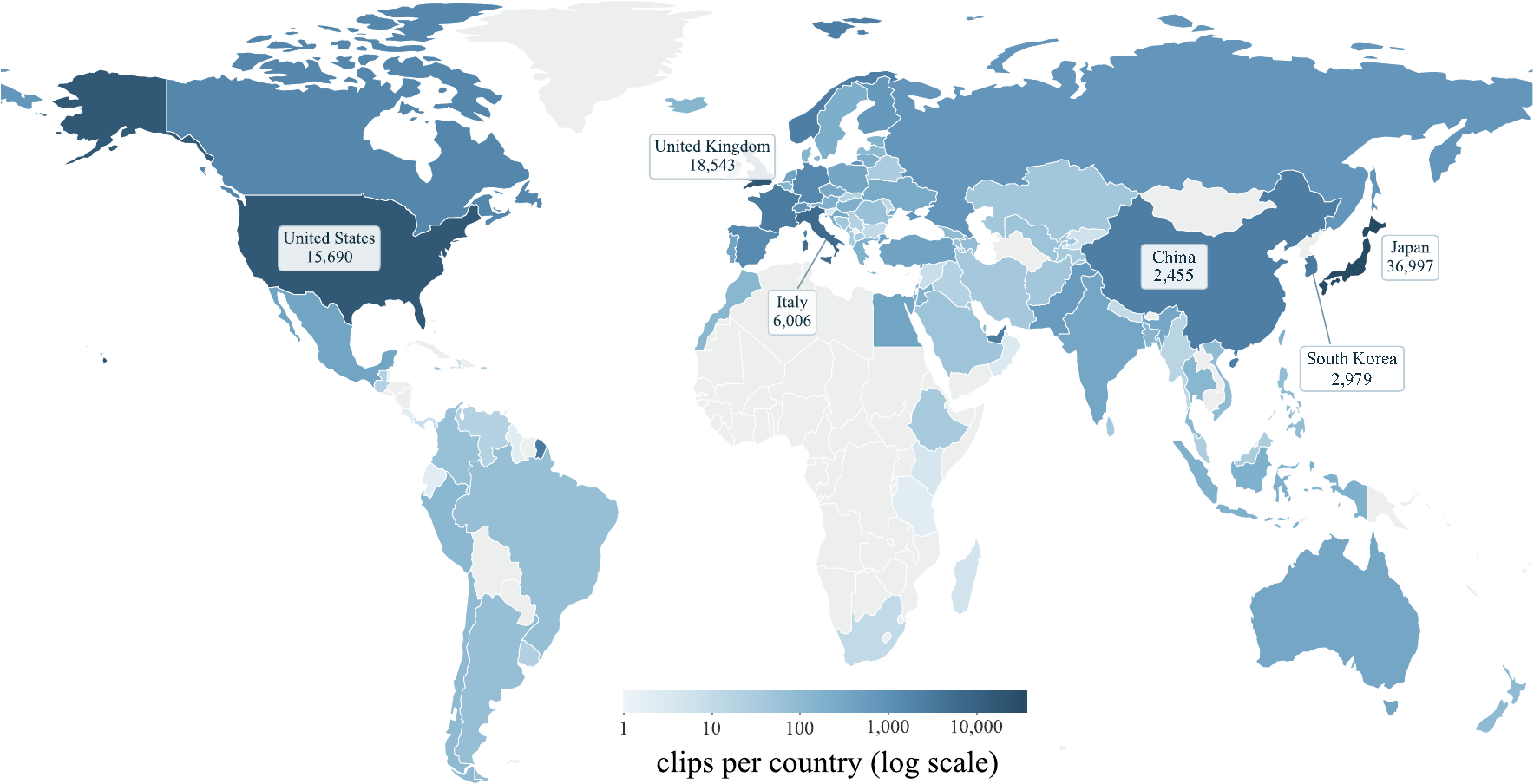}
    \vspace{4pt}
    \caption{\textbf{Geographic coverage of \dataset.} The map shows the number of clips per country or region on a $\log_{10}$ scale, with grey indicating no available clips. The five largest contributors are labelled, together with China, where the panoramic subset was captured.}
    \label{fig:world_map}
\end{figure}

\dataset covers $113$ countries or regions after geographic-label normalization (Fig.~\ref{fig:world_map}). As expected for web-sourced data, the geographic distribution is broad but long-tailed: Japan ($36{,}997$ clips), the United Kingdom ($18{,}543$), and the United States ($15{,}690$) together account for approximately two thirds of geographically resolved clips. At the continental level, Asia, Europe, and North America contribute $44.1\%$, $36.8\%$, and $16.5\%$, respectively. Nevertheless, coverage extends well beyond these dominant regions: the top $14$ countries account for $90\%$ of the resolved clips, while the remaining countries and regions form a geographically diverse tail spanning all inhabited continents, including South America, Africa, and Southeast Asia.

\subsection{Camera-Trajectory Diversity}
\label{sec:statistics_camera}

\begin{figure*}[t]
    \centering
    \includegraphics[width=\linewidth]{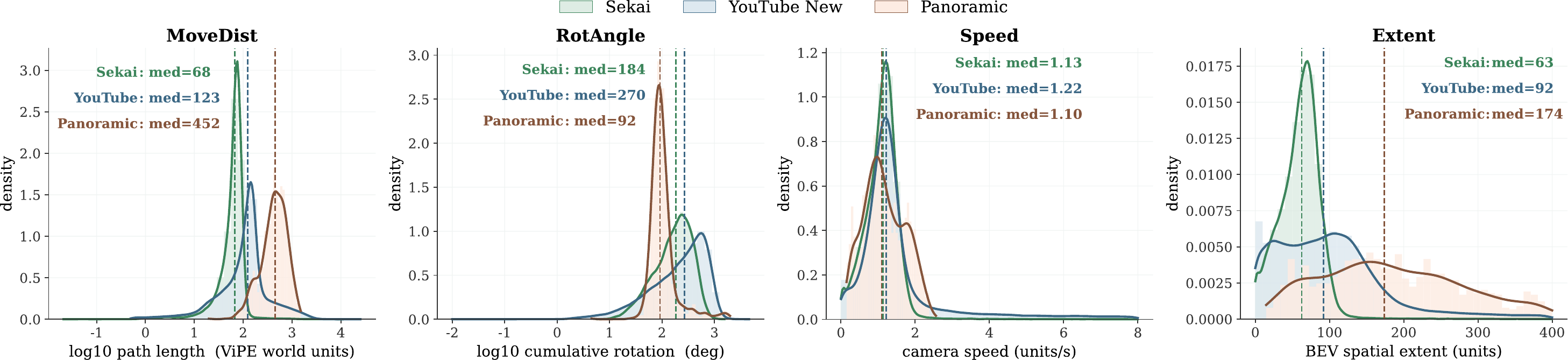}
    \vspace{5pt}
    \caption{\textbf{Camera-motion statistics across data sources.} Kernel-density distributions of four pose-derived trajectory descriptors over all released clips, with dashed lines indicating medians. Path length and cumulative rotation are shown on a $\log_{10}$ scale due to their heavy-tailed distributions. Translation-related quantities are expressed in ViPE world units, whose scale is estimated independently for each clip; they therefore characterize distributional differences rather than metric distances.}
    \label{fig:camera_motion}
\end{figure*}

Because every released clip is paired with a camera trajectory, we compute motion statistics over the full corpus rather than a sampled subset. As shown in Fig.~\ref{fig:camera_motion}, the three sources exhibit complementary trajectory profiles. Newly collected YouTube clips have greater median path length than the inherited Sekai subset ($123$ vs.\ $68$ ViPE units) and accumulate more rotation ($270^\circ$ vs.\ $184^\circ$), reflecting their longer and more varied traversal patterns. Panoramic sequences exhibit the broadest spatial coverage, with a median path length of $452$ units and a median bird's-eye-view extent of $174$ units, compared with $63$ units for Sekai. Their lower cumulative rotation ($92^\circ$) is consistent with omnidirectional capture, which preserves observations in all viewing directions without requiring equivalent camera-body rotation. In contrast, median camera speed remains similar across sources ($1.10$--$1.22$ units/s), suggesting that their differences arise primarily from trajectory extent and geometry rather than traversal pace.

\subsection{Caption Density and Lexical Structure}
\label{sec:statistics_caption}

\begin{figure*}[t]
    \centering
    \includegraphics[width=0.78\textwidth]{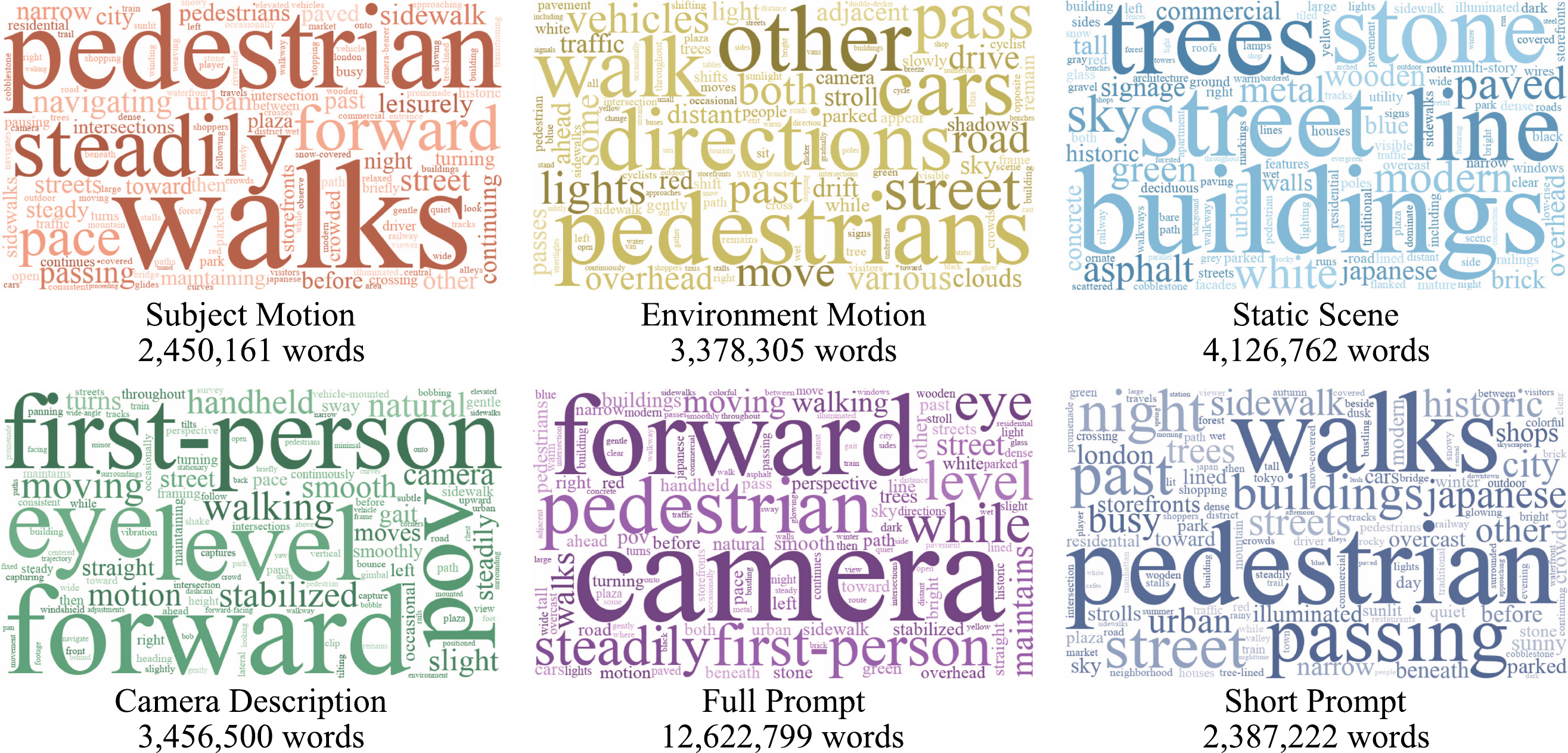}
    \caption{\textbf{Lexical structure of the clip-level annotations.} Word clouds visualize the most frequent content words in six annotation components, with word size proportional to within-component frequency. Colors distinguish components, and the values below each cloud report their total word counts. Quantitative analyses of their semantic complementarity are provided in Sec.~\ref{sec:exp_caption}.}
    \label{fig:word_cloud}
\end{figure*}

Counting all annotation fields, the caption corpus contains about $135$ million words. It comprises $649{,}597$ temporally grounded segments, corresponding to an average of $5.04$ segments per clip and a median segment duration of $15$ seconds. At the clip level, \emph{subject motion}, \emph{environment motion}, \emph{static scene content}, \emph{camera behavior}, and the camera-free short prompt contain an average of $19$, $26$, $32$, $27$, and $18$ words, respectively (Fig.~\ref{fig:scene_caption}b). The full prompt integrates these complementary fields into a generation-ready description, averaging $97$ words at the clip level and $79.6$ words at the segment level (Fig.~\ref{fig:scene_caption}c).

Fig.~\ref{fig:word_cloud} further reveals a clear semantic division across annotation fields. \texttt{subject\_motion} emphasizes agents and actions, \texttt{environment\_motion} captures independently moving objects and ambient changes, \texttt{static\_scene} describes persistent geometry and appearance, and \texttt{camera\_description} focuses on viewpoint and acquisition behavior. The \texttt{full\_prompt} combines scene, motion, and camera information into a detailed camera-aware narrative, whereas the \texttt{short\_prompt} provides a more compact, scene-centric description. Detailed vocabulary-growth, uniqueness, and duplication statistics are reported in Appendix~\ref{app:caption_statistics}, while semantic complementarity and caption quality are evaluated in Sec.~\ref{sec:exp_caption}.

\section{Experiments}
\label{sec:experiments}

Beyond dataset scale and coverage, reliable world-modeling supervision requires high-quality visual observations, camera geometry, and semantic annotations. We therefore evaluate \dataset along these three dimensions. We first compare its intrinsic visual quality and local temporal dynamics with representative video corpora under an identical evaluation pipeline. We then assess the validity and geometric consistency of the released camera trajectories, followed by the quality of the hierarchical captions. Finally, we examine the agreement between pose-derived camera motion and caption-level camera annotations. All evaluations are inference-only and require no additional training.

\subsection{Visual Quality and Temporal Dynamics}
\label{sec:exp_frame_quality}

\begin{figure*}[t]
    \centering
    \includegraphics[width=\linewidth]{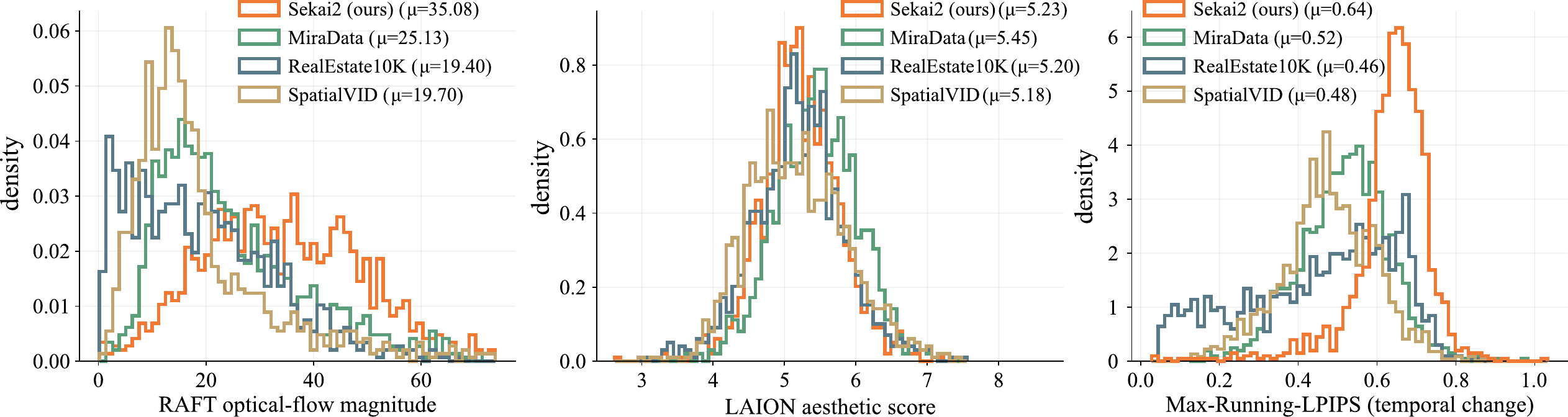}
    \vspace{4pt}
    \caption{\textbf{Intrinsic visual quality and temporal dynamics.} Per-clip distributions of RAFT optical-flow magnitude, LAION aesthetic score, and Max-Running-LPIPS for \dataset and three representative corpora. Each dataset contributes $1{,}200$ clips evaluated using the same models and preprocessing pipeline. Motion signals are computed over five adjacent frame pairs at a long-side resolution of $512$, while the aesthetic score is averaged over individual frames.}
    \label{fig:frame_quality}
\end{figure*}

We first examine whether the long-form, egocentric nature of \dataset compromises visual quality. We sample $1{,}200$ clips from \dataset and from each of MiraData~\cite{ju2024miradata}, RealEstate10K~\cite{zhou2018stereo}, and SpatialVID~\cite{wang2026spatialvid}, and evaluate all clips using the same aesthetic predictor, RAFT model, LPIPS network, image resolution, and sample size. Motion is measured over adjacent frame pairs rather than frames spread across the full clip, avoiding a duration-dependent bias when comparing minute-scale videos with substantially shorter baselines.

As shown in Fig.~\ref{fig:frame_quality}, the four datasets exhibit comparable aesthetic quality. \dataset achieves an average LAION aesthetic score of $5.23$, close to RealEstate10K ($5.20$) and SpatialVID ($5.18$), and moderately below MiraData ($5.45$). In contrast, \dataset contains substantially stronger local temporal variation: its mean optical-flow magnitude is $35.08$, compared with $25.13$, $19.40$, and $19.70$, while its Max-Running-LPIPS score is $0.64$, compared with $0.52$, $0.46$, and $0.48$. The separation is visible across the distributions rather than being driven only by a small number of highly dynamic clips: $81\%$ of \dataset clips exceed the median optical-flow magnitude of MiraData, and $91\%$ exceed its median Max-Running-LPIPS.

These results indicate that \dataset preserves frame-level visual quality comparable to existing corpora while providing richer local motion and visual change. The evaluation covers perspective videos only, since raw equirectangular panoramic frames are not directly comparable under the adopted models. Baseline results are computed on the dataset copies available to us rather than their complete public releases.

\subsection{Camera-Trajectory Quality}
\label{sec:exp_pose}

We assess the released camera trajectories from three complementary perspectives: corpus-wide numerical validity and smoothness, image-based epipolar consistency, and cross-run geometric consistency.

\paragraph{Corpus-wide validity and smoothness.}
We first audit every released trajectory using the finite-pose ratio, the rate of robust inter-frame translation outliers, and the median second difference of the inter-frame rotation angle. As shown in Table~\ref{tab:pose_intrinsic}, all three data sources achieve a median valid-pose ratio of $1.0$, together with near-zero translation-jump rates and low rotational jerk. These corpus-wide statistics indicate that the released trajectories are numerically complete and locally smooth across the full dataset.

\begin{table}[t]
    \centering
    \caption{\textbf{Validity and smoothness of all released trajectories.} Values are medians over clips. A jump is an inter-frame translation step exceeding the clip-specific median by five robust standard deviations. Rotational jerk is measured in degrees per frame squared. Lower is better except for validity.}
    \label{tab:pose_intrinsic}
    \vspace{5pt}
    \resizebox{0.52\linewidth}{!}{%
    \begin{tabular}{lrrrr}
        \toprule
        Source & Clips & Valid $\uparrow$ & Jump rate $\downarrow$ & Rot. jerk $\downarrow$ \\
        \midrule
        Inherited Sekai & $47{,}699$ & $1.000$ & $0.00000$ & $0.00318$ \\
        New YouTube & $80{,}211$ & $1.000$ & $0.00028$ & $0.01052$ \\
        Panoramic & $982$ & $1.000$ & $0.00010$ & $0.00427$ \\
        \midrule
        \textbf{All} & \textbf{$128{,}892$} & \textbf{$1.000$} & \textbf{$0.00000$} & \textbf{$0.00725$} \\
        \bottomrule
    \end{tabular}}
\end{table}

\begin{figure*}[t]
    \centering
    \includegraphics[width=0.80\linewidth]{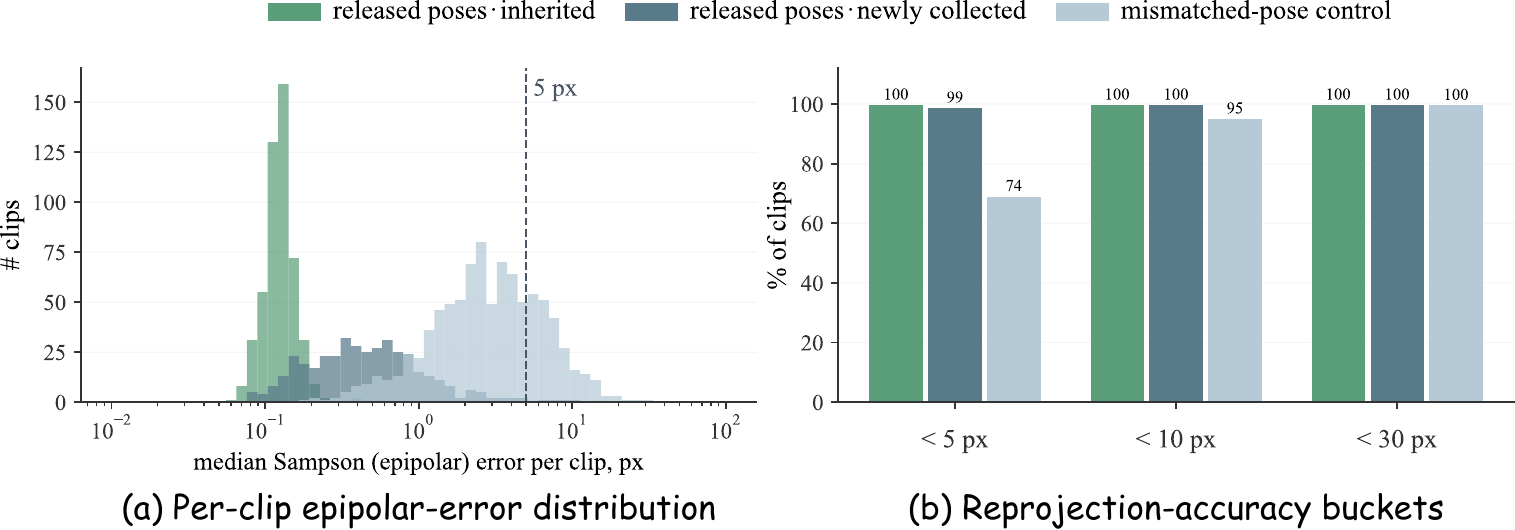}
    \vspace{5pt}
    \caption{\textbf{Ground-truth-free epipolar validation.} (a) Per-clip median Sampson-error distributions for the inherited and newly collected subsets, together with a mismatched-pose negative control that shifts the pose pair along the same trajectory while preserving all image-side inputs. (b) Fraction of clips below the $5$, $10$, and $30$\,px thresholds.}
    \label{fig:pose_epipolar}
    \vspace{-4pt}
\end{figure*}

\paragraph{Ground-truth-free epipolar consistency.}
Numerical validity alone does not guarantee that a trajectory is geometrically consistent with the observed video. We therefore evaluate the released poses against SIFT correspondences between frames separated by $0.5$,s. For each frame pair, we construct the fundamental matrix induced by the released poses and intrinsics and compute a trimmed Sampson error after removing the largest $20\%$ of match errors. As shown in Fig.~\ref{fig:pose_epipolar}, the inherited subset achieves a median error of $0.12$,px over $500$ clips, with all clips below $5$,px. For the $399$ newly collected clips, a  evaluation using a fixed prior focal length yields a median error of $0.43$,px, with $99.0\%$ of clips below $5$,px and all clips below $10$,px.

To verify that these low errors depend on correct pose--frame correspondence rather than image content alone, we construct a mismatched-pose negative control. Each aligned pose pair is replaced by another pair shifted by one third of the same trajectory, while the image pair, SIFT correspondences, temporal interval, intrinsics protocol, and trimming procedure remain unchanged. Under this mismatch, the median error increases to $2.83$,px, and the fraction of clips below $5$,px drops to $74.0\%$ (Fig.~\ref{fig:pose_epipolar}). This substantial degradation confirms that the observed epipolar consistency depends on the released pose--frame alignment. Additional protocol details, focal-length sensitivity, and motion-stratified analysis of the negative control are provided in Appendix~\ref{app:pose_epipolar_details}.

\paragraph{Cross-run geometric consistency.}
We further examine whether the released trajectories preserve their geometric structure under a separate reconstruction run. Specifically, we compare the released ViPE trajectories against DROID-SLAM reconstructions under a matched-intrinsics protocol. To ensure comparability across data sources, camera intrinsics are re-estimated from every input video using GeoCalib, while all subsets use the same DROID configuration, temporal sampling, and reconstruction settings. Corpus-level results are aggregated using the same source--motion weighting scheme. Sampling procedures, trajectory alignment, metric definitions, and calibration controls are provided in Appendix~\ref{app:matched_pose_details}.

As shown in Table~\ref{tab:matched_pose_evaluation}, the reconstructed trajectories exhibit strong cross-run geometric consistency over the weighted corpus distribution. The median rotation RPE is $0.103^\circ$ at one second and $0.414^\circ$ at five seconds, while the corresponding translation-direction errors are $1.88^\circ$ and $1.48^\circ$. The normalized ATE is $0.0176$, and the source--motion-weighted reconstruction success rate reaches $99.5\%$. The inherited Sekai and main-crawl subsets remain highly consistent in rotation, whereas the larger translation-direction error of the static supplement is concentrated primarily in near-static sequences, where translation direction becomes poorly conditioned as displacement approaches zero.

Matching the calibration source reduces part of the apparent variation across subsets but does not eliminate it. The remaining disagreement depends strongly on trajectory structure: straight motion exhibits the strongest cross-run agreement, while curved and winding motion is more challenging and more sensitive to intrinsic calibration. These motion-wise and source--motion effects are analyzed further in Appendix~\ref{app:matched_pose_details}. Finally, because ViPE itself employs a DROID-based geometric stack, this experiment should be interpreted as evidence of \emph{cross-run geometric consistency} across reconstruction runs and calibration settings, rather than as accuracy against independent metric ground truth.

\begin{table*}[t]
    \centering
    \caption{\textbf{Cross-run geometric consistency under matched intrinsics.}
    All subsets are reconstructed using GeoCalib-estimated intrinsics and identical DROID settings. Results are source--motion-weighted per-clip medians. Rotation RPE and translation-direction error are reported in degrees, while ATE is normalized by trajectory path length. Panoramic sequences are excluded because the reconstruction assumes pinhole imagery. Lower is better.}
    \vspace{3pt}
    \label{tab:matched_pose_evaluation}
    \resizebox{0.88\textwidth}{!}{%
    \begin{tabular}{lrrrrrrr}
        \toprule
        Source & $n$ & Success $\uparrow$ & Rot. $1$s $\downarrow$ & Rot. $5$s $\downarrow$
        & T-dir. $1$s $\downarrow$ & T-dir. $5$s $\downarrow$ & ATE$_\mathrm{norm}$ $\downarrow$ \\
        \midrule
        Inherited Sekai & $395$ & $99.9\%$ & $0.072^\circ$ & $0.281^\circ$
        & $1.09^\circ$ & $0.86^\circ$ & $0.0100$ \\
        New YouTube, main crawl & $480$ & $100.0\%$ & $0.134^\circ$ & $0.541^\circ$
        & $2.57^\circ$ & $2.19^\circ$ & $0.0281$ \\
        New YouTube, static supplement & $125$ & $89.3\%$ & $0.088^\circ$ & $0.313^\circ$
        & $13.14^\circ$ & $7.26^\circ$ & $0.0257$ \\
        \midrule
        \textbf{All (weighted)} & \textbf{$1{,}000$} & \textbf{$99.5\%$}
        & $\mathbf{0.103^\circ}$ & $\mathbf{0.414^\circ}$
        & $\mathbf{1.88^\circ}$ & $\mathbf{1.48^\circ}$ & $\mathbf{0.0176}$ \\
        \bottomrule
    \end{tabular}}
\end{table*}

\begin{table*}[t]
    \centering
    \caption{\textbf{Judge-based caption assessment.} VLM scores are means on a $1$--$5$ scale over $1{,}500$ clips, with $12$ sampled frames provided to the judge. Pairwise results compare our structured annotation with one-shot captions generated by the same VLM from identical frames over $800$ clips. Given a $71.6\%$ first-position bias and no ties, we report the conservative condition with our annotation shown second; the reverse order saturates at $100\%$.}
    \label{tab:caption_judge}
    \vspace{5pt}
    \resizebox{0.72\linewidth}{!}{%
    \begin{tabular}{llr}
        \toprule
        Protocol & Metric & Value \\
        \midrule
        \multirow{4}{*}{VLM judge (frames + caption, $n{=}1{,}500$)}
          & Coverage & $4.28$ \\
          & Scene correctness & $4.31$ \\
          & Motion correctness & $4.14$ \\
          & Camera correctness & $4.01$ \\
        \midrule
        \multirow{3}{*}{Pairwise vs.\ one-shot (same VLM, $n{=}800$)}
          & Ours preferred (shown second) & $60.9\%$ \\
          & Ours more informative (shown second) & $68.4\%$ \\
          & First-position preference & $71.6\%$ \\
        \bottomrule
    \end{tabular}}
\end{table*}

\subsection{Caption Quality}
\label{sec:exp_caption}

\paragraph{Judge-based assessment.} We assess caption quality using a frame-aware VLM judge provided with $12$ uniformly sampled frames and the corresponding annotation (Table~\ref{tab:caption_judge}). Across $1{,}500$ clips, our annotations achieve mean scores of $4.28$ for coverage, $4.31$ for scene correctness, $4.14$ for motion correctness, and $4.01$ for camera correctness. These results indicate strong overall visual grounding, while the comparatively lower scores for motion and camera behavior reflect the greater difficulty of inferring temporally extended dynamics from sparsely sampled frames. 

We further conduct a controlled pairwise evaluation on $800$ clips, comparing our structured annotations with conventional one-shot captions generated by the same VLM from identical visual inputs. Because the judge exhibits a pronounced positional bias, favoring the first response in $71.6\%$ of comparisons and never selecting a tie, we report only the conservative condition in which our annotation is presented second. Even under this unfavorable ordering, our annotations are preferred in $60.9\%$ of comparisons and judged more informative in $68.4\%$. By holding both the visual input and captioning model fixed, this comparison isolates the benefit of the structured annotation protocol from differences in model capacity.

\begin{figure*}[t]
    \centering
    \includegraphics[width=0.92\linewidth]{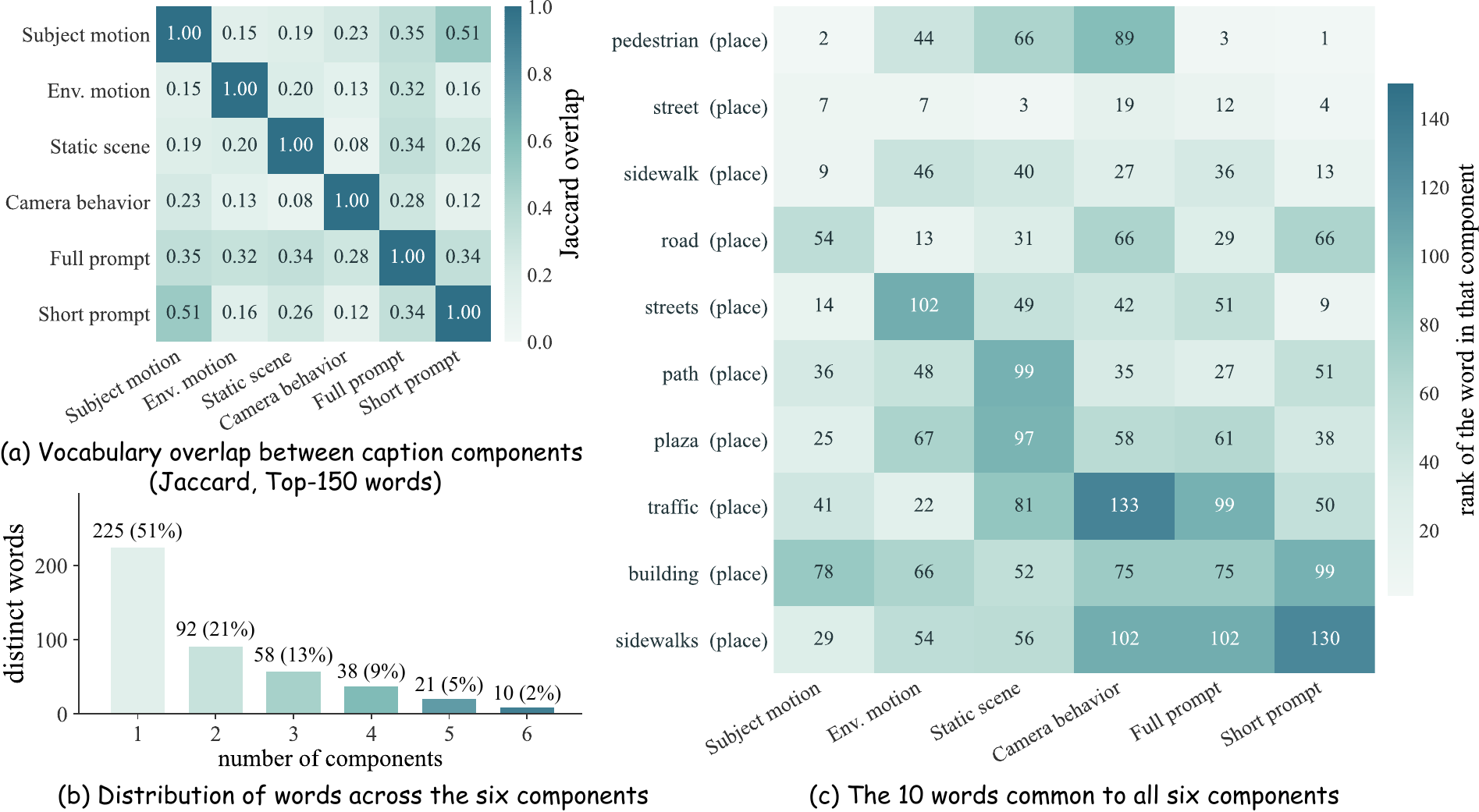}
    \vspace{4pt}
    \caption{\textbf{Complementarity within the clip-level annotation schema.} Each description type is represented by its $150$ most frequent content words. (a) Pairwise Jaccard overlap between description types. (b) Number of description types in which each word from the union appears. (c) Within-description ranks of the ten words shared by all six descriptions, illustrating common scene context with different descriptive emphases.}
    \label{fig:vocab_components_clip}
\end{figure*}

\paragraph{Complementary semantics across annotation views.}
A structured annotation schema is most useful when its descriptions emphasize complementary aspects of the video rather than repeatedly expressing the same information. To characterize this property, we represent each of the six clip-level annotation views by its $150$ most frequent content words and measure their pairwise vocabulary overlap (Fig.~\ref{fig:vocab_components_clip}). The mean Jaccard overlap is $0.24$ across all six views and decreases to $0.16$ among the four factorized descriptions. Of the $444$ distinct words in the combined vocabulary, $225$ ($51\%$) occur in only one view, whereas only $10$ ($2\%$) are shared across all six, indicating substantial differentiation among the descriptions.

The vocabulary profiles further reflect the intended roles of the annotation views. Camera behavior, environment motion, static scene, subject motion, and the short prompt contain $64$, $56$, $52$, $22$, and $29$ view-specific words, respectively. In contrast, the derived full prompt contains only two exclusive words, consistent with its role in integrating and reformulating the factorized descriptions. Its greater overlap with the other views is therefore expected, with the largest pairwise Jaccard score of $0.51$ observed between subject motion and the short prompt. The ten words shared across all six views are dominated by scene and location concepts, while their within-view ranks vary substantially, suggesting a shared scene context with distinct descriptive emphases. Overall, the schema captures subject motion, environmental dynamics, static scene content, and camera behavior through complementary descriptions while preserving the common context required for a coherent representation of the video. The corresponding segment-level analysis is provided in Appendix~\ref{app:vocab_segment}.

\begin{figure*}[t]
    \centering
    \includegraphics[width=0.82\linewidth]{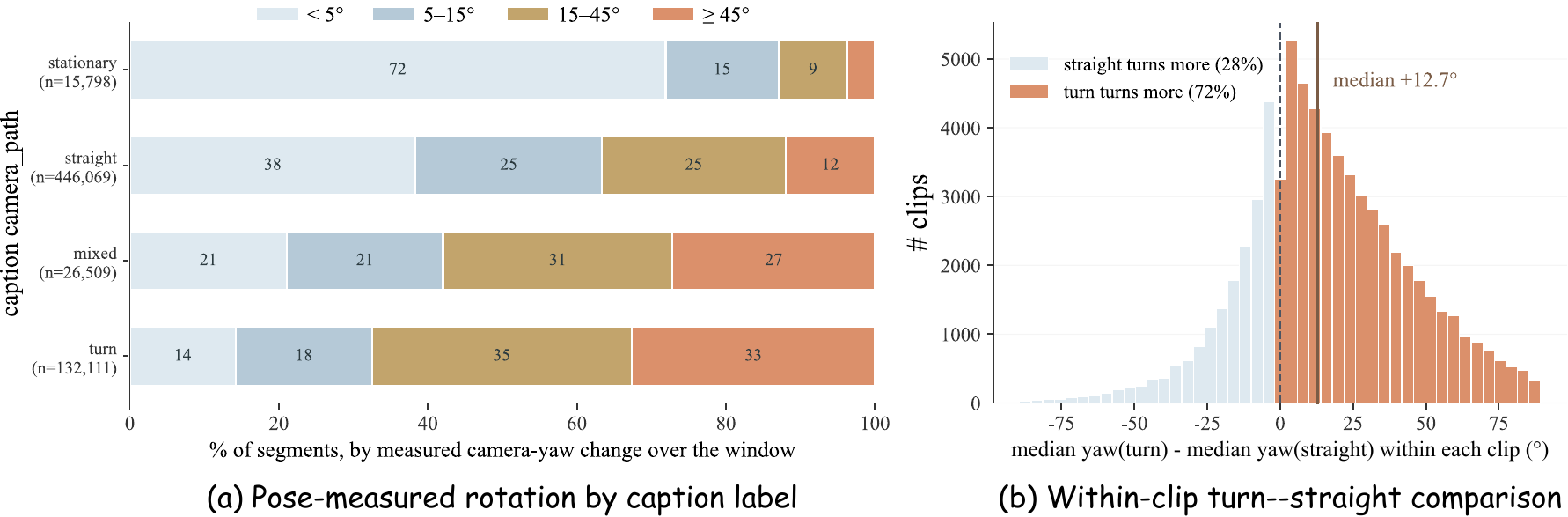}
    \vspace{4pt}
    \caption{\textbf{Segment-level pose--caption agreement.} (a) Distribution of pose-measured camera-yaw change within each caption \texttt{camera\_path} class over $620{,}487$ aligned segments. (b) Within-clip difference between the median yaw change of \texttt{turn} and \texttt{straight} segments for $74{,}437$ clips containing both labels. Positive values indicate stronger measured rotation for segments labeled \texttt{turn}.}
    \label{fig:segment_grounding}
\end{figure*}

\subsection{Cross-Modal Grounding}
\label{sec:exp_grounding}

Captions and camera trajectories are produced by independent pipelines, making their agreement an end-to-end test of cross-modal grounding. For the $620{,}487$ segments ($95.5\%$ of all annotated segments) whose temporal window could be aligned with the released trajectory, we compute the camera-yaw change from the trajectory over the temporal window and compare it with the \texttt{camera\_path} label (Fig.~\ref{fig:segment_grounding}). Measured rotation increases monotonically from \texttt{stationary} to \texttt{straight}, \texttt{mixed}, and \texttt{turn}, with median yaw changes of $1.3^\circ$, $8.5^\circ$, $20.6^\circ$, and $27.9^\circ$, respectively. Along the same ordering, the fraction of segments rotating less than $5^\circ$ decreases from $72\%$ to $14\%$, while the fraction exceeding $45^\circ$ increases from $4\%$ to $33\%$. The labels distinguish turning from non-turning segments with an AUC of $0.709$.

To control for clip-level differences in overall camera motion, capture style, and trajectory scale, we further compare labels within the same video. Among $74{,}437$ clips containing at least one \texttt{turn} segment and one \texttt{straight} segment, the median yaw change of the \texttt{turn} segments is larger in $71.8\%$ of clips, with a median within-clip difference of $12.7^\circ$. The distribution in Fig.~\ref{fig:segment_grounding}(b) is therefore shifted toward positive differences, showing that the ordering persists even after each clip serves as its own control. At the same time, the remaining $28.2\%$ of clips exhibit equal or stronger rotation for \texttt{straight} segments, indicating substantial but non-perfect overlap between the two semantic categories. This is expected for coarse discrete path labels that summarize local camera behavior rather than encode the exact magnitude of geometric rotation. Overall, these results demonstrate a consistent correspondence between the semantic camera labels and the independently estimated continuous camera geometry, providing direct evidence that the caption annotations are temporally grounded in the observed camera motion.


\section{Conclusion}
\label{sec:conclusion}

We introduce \dataset to advance real-world video data from \emph{world exploration} toward \emph{interactive world modeling}. Rather than scaling a single supervision dimension, \dataset jointly provides long-form real-world video, explicit camera trajectories, and temporally grounded hierarchical semantic annotations within a unified corpus, aligning scene content, subject and environmental motion, and viewpoint changes on a shared timeline. We further introduce a dedicated revisit-rich panoramic subset containing long non-linear trajectories, loops, and repeated observations of previously visited locations, complementing the revisit structures that are relatively scarce in ordinary web video. Together with a unified pipeline for data construction, trajectory processing, semantic annotation, and quality validation, \dataset brings long-horizon continuity, camera trajectories, fine-grained temporal semantics, and revisit structure into a common data framework. We hope these designs provide richer real-world supervision for long-horizon video generation and camera-controllable synthesis, while supporting research on viewpoint evolution, long-range spatial consistency, and revisit consistency during extended exploration.

\paragraph{Limitations and future work.}
Several limitations remain. First, the released camera trajectories are primarily estimated from monocular video and may therefore retain reconstruction errors and scale ambiguity despite extensive quality filtering and geometric validation. Second, the semantic annotations are automatically generated by vision-language models and may contain residual omissions, temporal-boundary errors, or semantic noise. Third, although camera trajectories provide an important control signal for viewpoint changes, \dataset does not include dense ground-truth user actions such as keyboard, mouse, or simulator controls, and therefore does not cover the complete action space required by general interactive agents. The data distribution also remains long-tailed across geography, weather conditions, capture devices, and motion patterns. In addition, the DROID-based evaluation measures geometric consistency across reconstruction runs rather than absolute accuracy against metric ground-truth poses. Future work will further improve camera calibration and trajectory uncertainty estimation, strengthen human or programmatic verification of difficult semantic annotations, expand underrepresented environments and interaction signals, and establish standardized benchmarks for long-horizon generation, camera control, and revisit consistency.

\paragraph{Panoramic data release.}
For the panoramic subset, we collected $1{,}283$ raw sequences totaling approximately $160$ hours and retained $982$ sequences totaling approximately $119$ hours after quality filtering and trajectory processing. Due to data-distribution agreements and licensing constraints, the current public release can include only approximately $20$ hours of this material. Accordingly, the panoramic statistics and analyses reported in this paper are based on the full $119$ hours of processed data, while the publicly downloadable portion is more limited. We will continue to pursue compliant release of additional panoramic sequences and expand real-world trajectories containing long-range loops and repeated visits to support research on revisit consistency and long-range spatial consistency.

\paragraph{Broader impact and responsible release.}
\dataset is intended as a research resource for long-horizon video generation, camera-controllable synthesis, and interactive world modeling. Because the corpus combines web-sourced and self-captured videos, its use and release should comply with applicable source licenses, attribution requirements, privacy regulations, and other data-distribution agreements. Downstream users should also apply appropriate safeguards for faces, license plates, and other potentially sensitive visual content. We provide source provenance, semantic annotations, camera-pose files, and processing documentation to support auditing of dataset composition and improve transparency and reproducibility under the applicable data-access conditions.

\bibliographystyle{abbrv}
\bibliography{main}

@String{CVPR    = {Proceedings of the IEEE/CVF Conference on Computer Vision and Pattern Recognition (CVPR)}}

@String{ICCV    = {Proceedings of the IEEE/CVF International Conference on Computer Vision (ICCV)}}

@inproceedings{bain2021frozen,
  title={Frozen in time: A joint video and image encoder for end-to-end retrieval},
  author={Bain, Max and Nagrani, Arsha and Varol, G{\"u}l and Zisserman, Andrew},
  booktitle={2021 IEEE/CVF International Conference on Computer Vision (ICCV)},
  pages={1708--1718},
  year={2021},
  organization={IEEE}
}

@inproceedings{wang2024internvid,
  title={Internvid: A large-scale video-text dataset for multimodal understanding and generation},
  author={Wang, Yi and He, Yinan and Li, Yizhuo and Li, Kunchang and Yu, Jiashuo and Ma, Xin and Li, Xinhao and Chen, Guo and Chen, Xinyuan and Wang, Yaohui and others},
  booktitle={International Conference on Learning Representations},
  volume={2024},
  pages={42055--42079},
  year={2024}
}

@inproceedings{chen2024panda,
  title={Panda-70m: Captioning 70m videos with multiple cross-modality teachers},
  author={Chen, Tsai-Shien and Siarohin, Aliaksandr and Menapace, Willi and Deyneka, Ekaterina and Chao, Hsiang-wei and Jeon, Byung Eun and Fang, Yuwei and Lee, Hsin-Ying and Ren, Jian and Yang, Ming-Hsuan and others},
  booktitle={2024 IEEE/CVF Conference on Computer Vision and Pattern Recognition (CVPR)},
  pages={13320--13331},
  year={2024},
  organization={IEEE}
}

@inproceedings{nan2025openvid,
  title={Openvid-1m: A large-scale high-quality dataset for text-to-video generation},
  author={Nan, Kepan and Xie, Rui and Zhou, Penghao and Fan, Tiehan and Yang, Zhenheng and Chen, Zhijie and Li, Xiang and Yang, Jian and Tai, Ying},
  booktitle={International Conference on Learning Representations},
  volume={2025},
  pages={1045--1064},
  year={2025}
}

@inproceedings{grauman2022ego4d,
  title={Ego4d: Around the world in 3,000 hours of egocentric video},
  author={Grauman, Kristen and Westbury, Andrew and Byrne, Eugene and Chavis, Zachary and Furnari, Antonino and Girdhar, Rohit and Hamburger, Jackson and Jiang, Hao and Liu, Miao and Liu, Xingyu and others},
  booktitle={Proceedings of the IEEE/CVF conference on computer vision and pattern recognition},
  pages={18995--19012},
  year={2022}
}

@article{tan2024vidgen,
  title={Vidgen-1m: A large-scale dataset for text-to-video generation},
  author={Tan, Zhiyu and Yang, Xiaomeng and Qin, Luozheng and Li, Hao},
  journal={arXiv preprint arXiv:2408.02629},
  year={2024}
}

@article{chen2024sharegpt4video,
  title={Sharegpt4video: Improving video understanding and generation with better captions},
  author={Chen, Lin and Wei, Xilin and Li, Jinsong and Dong, Xiaoyi and Zhang, Pan and Zang, Yuhang and Chen, Zehui and Duan, Haodong and Lin, Bin and Tang, Zhenyu and others},
  journal={Advances in Neural Information Processing Systems},
  volume={37},
  pages={19472--19495},
  year={2024}
}

@article{yang2024vript,
  title={Vript: A video is worth thousands of words},
  author={Yang, Dongjie and Huang, Suyuan and Lu, Chengqiang and Han, Xiaodong and Zhang, Haoxin and Gao, Yan and Hu, Yao and Zhao, Hai},
  journal={Advances in Neural Information Processing Systems},
  volume={37},
  pages={57240--57261},
  year={2024}
}

@article{ju2024miradata,
  title={Miradata: A large-scale video dataset with long durations and structured captions},
  author={Ju, Xuan and Gao, Yiming and Zhang, Zhaoyang and Yuan, Ziyang and Wang, Xintao and Zeng, Ailing and Xiong, Yu and Xu, Qiang and Shan, Ying},
  journal={Advances in Neural Information Processing Systems},
  volume={37},
  pages={48955--48970},
  year={2024}
}

@article{zhou2018stereo,
  title={Stereo magnification: Learning view synthesis using multiplane images},
  author={Zhou, Tinghui and Tucker, Richard and Flynn, John and Fyffe, Graham and Snavely, Noah},
  journal={arXiv preprint arXiv:1805.09817},
  year={2018}
}

@inproceedings{ling2024dl3dv,
  title={Dl3dv-10k: A large-scale scene dataset for deep learning-based 3d vision},
  author={Ling, Lu and Sheng, Yichen and Tu, Zhi and Zhao, Wentian and Xin, Cheng and Wan, Kun and Yu, Lantao and Guo, Qianyu and Yu, Zixun and Lu, Yawen and others},
  booktitle={2024 IEEE/CVF Conference on Computer Vision and Pattern Recognition (CVPR)},
  pages={22160--22169},
  year={2024},
  organization={IEEE}
}

@inproceedings{wang2020tartanair,
  title={Tartanair: A dataset to push the limits of visual slam},
  author={Wang, Wenshan and Zhu, Delong and Wang, Xiangwei and Hu, Yaoyu and Qiu, Yuheng and Wang, Chen and Hu, Yafei and Kapoor, Ashish and Scherer, Sebastian},
  booktitle={2020 IEEE/RSJ International Conference on Intelligent Robots and Systems (IROS)},
  pages={4909--4916},
  year={2020},
  organization={IEEE}
}

@inproceedings{wang2026spatialvid,
  title={Spatialvid: A large-scale video dataset with spatial annotations},
  author={Wang, Jiahao and Yuan, Yufeng and Zheng, Rujie and Lin, Youtian and Gao, Jian and Chen, Lin-Zhuo and Bao, Yajie and Zeng, Chang and Zhou, Yanxi and Long, Xiao-Xiao and others},
  booktitle={Proceedings of the IEEE/CVF Conference on Computer Vision and Pattern Recognition},
  pages={42592--42603},
  year={2026}
}

@article{zhou2025omniworld,
  title={Omniworld: A multi-domain and multi-modal dataset for 4d world modeling},
  author={Zhou, Yang and Wang, Yifan and Zhou, Jianjun and Chang, Wenzheng and Guo, Haoyu and Li, Zizun and Ma, Kaijing and Li, Xinyue and Wang, Yating and Zhu, Haoyi and others},
  journal={arXiv preprint arXiv:2509.12201},
  year={2025}
}

@InProceedings{Chen_2026_CVPR,
    author    = {Chen, Delong and Kasarla, Tejaswi and Bang, Yejin and Shukor, Mustafa and Chung, Willy and Yu, Jade and Bolourchi, Allen and Moutakanni, Theo and Fung, Pascale},
    title     = {Action100M: A Large-scale Video Action Dataset},
    booktitle = {Proceedings of the IEEE/CVF Conference on Computer Vision and Pattern Recognition (CVPR) Workshops},
    month     = {June},
    year      = {2026},
    pages     = {8832-8842}
}

@inproceedings{han2025videoespresso,
  title={Videoespresso: A large-scale chain-of-thought dataset for fine-grained video reasoning via core frame selection},
  author={Han, Songhao and Huang, Wei and Shi, Hairong and Zhuo, Le and Su, Xiu and Zhang, Shifeng and Zhou, Xu and Qi, Xiaojuan and Liao, Yue and Liu, Si},
  booktitle={2025 IEEE/CVF Conference on Computer Vision and Pattern Recognition (CVPR)},
  pages={26181--26191},
  year={2025},
  organization={IEEE}
}

@article{zhang2026leader360v,
  title={Leader360V: A Large-scale, Real-world 360 Video Dataset for Multi-task Learning in Diverse Environment},
  author={Zhang, Weiming and Xiao, Dingwen and Dai, Aobotao and Liu, Yexin and Pan, Tianbo and Wen, Shiqi and Chen, Lei and Wang, Lin},
  journal={Advances in Neural Information Processing Systems},
  volume={38},
  year={2026}
}

@article{li2026sekai,
  title={Sekai: A video dataset towards world exploration},
  author={Li, Zhen and Li, Chuanhao and Mao, Xiaofeng and Lin, Shaoheng and Li, Ming and Zhao, Shitian and Xu, Zhaopan and Li, Xinyue and Feng, Yukang and Sun, Jianwen and others},
  journal={Advances in Neural Information Processing Systems},
  volume={38},
  year={2026}
}

@article{yang2023learning,
  title={Learning interactive real-world simulators},
  author={Yang, Sherry and Du, Yilun and Ghasemipour, Kamyar and Tompson, Jonathan and Kaelbling, Leslie and Schuurmans, Dale and Abbeel, Pieter},
  journal={arXiv preprint arXiv:2310.06114},
  year={2023}
}

@article{hu2023gaia,
  title={Gaia-1: A generative world model for autonomous driving},
  author={Hu, Anthony and Russell, Lloyd and Yeo, Hudson and Murez, Zak and Fedoseev, George and Kendall, Alex and Shotton, Jamie and Corrado, Gianluca},
  journal={arXiv preprint arXiv:2309.17080},
  year={2023}
}

@inproceedings{valevski2025diffusion,
  title={Diffusion models are real-time game engines},
  author={Valevski, Dani and Leviathan, Yaniv and Arar, Moab and Fruchter, Shlomi},
  booktitle={International Conference on Learning Representations},
  volume={2025},
  pages={73754--73776},
  year={2025}
}

@article{decart2024oasis,
  title={Oasis: A universe in a transformer},
  author={Decart, Etched and McIntyre, Quinn and Campbell, Spruce and Chen, Xinlei and Wachen, Robert},
  journal={URL: https://oasis-model. github. io},
  volume={2},
  number={3},
  pages={6},
  year={2024}
}

@article{guo2025mineworld,
  title={Mineworld: a real-time and open-source interactive world model on minecraft},
  author={Guo, Junliang and Ye, Yang and He, Tianyu and Wu, Haoyu and Jiang, Yushu and Pearce, Tim and Bian, Jiang},
  journal={arXiv preprint arXiv:2504.08388},
  year={2025}
}

@article{menapace2024promptable,
  title={Promptable game models: Text-guided game simulation via masked diffusion models},
  author={Menapace, Willi and Siarohin, Aliaksandr and Lathuili{\`e}re, St{\'e}phane and Achlioptas, Panos and Golyanik, Vladislav and Tulyakov, Sergey and Ricci, Elisa},
  journal={ACM Transactions on Graphics},
  volume={43},
  number={2},
  pages={1--16},
  year={2024},
  publisher={ACM New York, NY, USA}
}

@article{kanervisto2025world,
  title={World and human action models towards gameplay ideation},
  author={Kanervisto, Anssi and Bignell, Dave and Wen, Linda Yilin and Grayson, Martin and Georgescu, Raluca and Valcarcel Macua, Sergio and Tan, Shan Zheng and Rashid, Tabish and Pearce, Tim and Cao, Yuhan and others},
  journal={Nature},
  volume={638},
  number={8051},
  pages={656--663},
  year={2025},
  publisher={Nature Publishing Group UK London}
}

@article{xiao2026worldmem,
  title={Worldmem: Long-term consistent world simulation with memory},
  author={Xiao, Zeqi and Lan, Yushi and Zhou, Yifan and Ouyang, Wenqi and Yang, Shuai and Zeng, Yanhong and Pan, Xingang},
  journal={Advances in Neural Information Processing Systems},
  volume={38},
  pages={49632--49652},
  year={2026}
}

@article{wu2026video,
  title={Video world models with long-term spatial memory},
  author={Wu, Tong and Yang, Shuai and Po, Ryan and Xu, Yinghao and Liu, Ziwei and Lin, Dahua and Wetzstein, Gordon},
  journal={Advances in Neural Information Processing Systems},
  volume={38},
  pages={49371--49393},
  year={2026}
}

@article{wu2026infinite,
  title={Infinite-world: Scaling interactive world models to 1000-frame horizons via pose-free hierarchical memory},
  author={Wu, Ruiqi and He, Xuanhua and Cheng, Meng and Yang, Tianyu and Zhang, Yong and Kang, Zhuoliang and Cai, Xunliang and Wei, Xiaoming and Guo, Chunle and Li, Chongyi and others},
  journal={arXiv preprint arXiv:2602.02393},
  year={2026}
}

@article{li2026panoworld,
  title={PanoWorld: Real-World Panoramic Generation},
  author={Li, Haoyuan and Zhang, Dizhe and Zhou, Yuemei and Zhang, Xiangkai and Feng, Haoran and Lin, Xiaofan and Jiang, Wenjie and Du, Bo and Yang, Ming-Hsuan and Qi, Lu},
  journal={arXiv preprint arXiv:2607.09661},
  year={2026}
}

@inproceedings{bruce2024genie,
  title={Genie: Generative interactive environments},
  author={Bruce, Jake and Dennis, Michael D and Edwards, Ashley and Parker-Holder, Jack and Shi, Yuge and Hughes, Edward and Lai, Matthew and Mavalankar, Aditi and Steigerwald, Richie and Apps, Chris and others},
  booktitle={Forty-first international conference on machine learning},
  year={2024}
}

@misc{parker2024genie,
  title={Genie 2: A large-scale foundation world model},
  author={Parker-Holder, Jack and Ball, Philip and Bruce, Jake and Dasagi, Vibhavari and Holsheimer, Kristian and Kaplanis, Christos and Moufarek, Alexandre and Scully, Guy and Shar, Jeremy and Shi, Jimmy and others},
  year={2024},
  howpublished={\url{https://deepmind.google/discover/blog/genie-2-a-large-scale-foundation-world-model}}
}

@misc{ball2025genie3,
  title={Genie 3: A new frontier for world models},
  author={Ball, Philip and Bauer, Jakob and others},
  year={2025},
  howpublished={\url{https://deepmind.google/discover/blog/genie-3-a-new-frontier-for-world-models/}}
}

@article{he2025matrix,
  title={Matrix-game 2.0: An open-source real-time and streaming interactive world model},
  author={He, Xianglong and Peng, Chunli and Liu, Zexiang and Wang, Boyang and Zhang, Yifan and Cui, Qi and Kang, Fei and Jiang, Biao and An, Mengyin and Ren, Yangyang and others},
  journal={arXiv preprint arXiv:2508.13009},
  year={2025}
}

@article{wang2026matrix,
  title={Matrix-game 3.0: Real-time and streaming interactive world model with long-horizon memory},
  author={Wang, Zile and Liu, Zexiang and Li, Jiaxing and Huang, Kaichen and Xu, Baixin and Kang, Fei and An, Mengyin and Wang, Peiyu and Jiang, Biao and Wei, Yichen and others},
  journal={arXiv preprint arXiv:2604.08995},
  year={2026}
}

@inproceedings{che2025gamegen,
  title={Gamegen-x: Interactive open-world game video generation},
  author={Che, Haoxuan and He, Xuanhua and Liu, Quande and Jin, Cheng and Chen, Hao},
  booktitle={International Conference on Learning Representations},
  volume={2025},
  pages={37546--37593},
  year={2025}
}

@article{li2025hunyuan,
  title={Hunyuan-gamecraft: High-dynamic interactive game video generation with hybrid history condition},
  author={Li, Jiaqi and Tang, Junshu and Xu, Zhiyong and Wu, Longhuang and Zhou, Yuan and Shao, Shuai and Yu, Tianbao and Cao, Zhiguo and Lu, Qinglin},
  journal={arXiv preprint arXiv:2506.17201},
  volume={2},
  number={3},
  pages={6},
  year={2025}
}

@article{tang2025hunyuan,
  title={Hunyuan-gamecraft-2: Instruction-following interactive game world model},
  author={Tang, Junshu and Liu, Jiacheng and Li, Jiaqi and Wu, Longhuang and Yang, Haoyu and Zhao, Penghao and Gong, Siruis and Yuan, Xiang and Shao, Shuai and Zhang, Linfeng and others},
  journal={arXiv preprint arXiv:2511.23429},
  year={2025}
}

@inproceedings{mao2026yume1,
  title={Yume1. 5: A text-controlled interactive world generation model},
  author={Mao, Xiaofeng and Li, Zhen and Li, Chuanhao and Xu, Xiaojie and Ying, Kaining and Zhang, Kaipeng},
  booktitle={Proceedings of the IEEE/CVF Conference on Computer Vision and Pattern Recognition},
  pages={7752--7761},
  year={2026}
}

@article{hyworld2025,
  title={HY-World 1.5: A Systematic Framework for Interactive World Modeling with Real-Time Latency and Geometric Consistency},
  author={Team HunyuanWorld},
  journal={arXiv preprint},
  year={2025}
}

@article{team2026advancing,
  title={Advancing open-source world models},
  author={Team, Robbyant and Gao, Zelin and Wang, Qiuyu and Zeng, Yanhong and Zhu, Jiapeng and Cheng, Ka Leong and Li, Yixuan and Wang, Hanlin and Xu, Yinghao and Ma, Shuailei and others},
  journal={arXiv preprint arXiv:2601.20540},
  year={2026}
}

@article{gao2026infinite,
  title={Infinite Worlds with Versatile Interactions},
  author={Gao, Zelin and Wang, Qiuyu and Zhu, Jiapeng and Chen, Jingye and Liu, Zichen and Bai, Qingyan and Wang, Jiahao and Yuan, Yufeng and Wang, Hanlin and Lu, Yichong and others},
  journal={arXiv preprint arXiv:2607.07534},
  year={2026}
}

@article{he2024cameractrl,
  title={Cameractrl: Enabling camera control for text-to-video generation},
  author={He, Hao and Xu, Yinghao and Guo, Yuwei and Wetzstein, Gordon and Dai, Bo and Li, Hongsheng and Yang, Ceyuan},
  journal={arXiv preprint arXiv:2404.02101},
  year={2024}
}

@inproceedings{wang2024motionctrl,
  title={Motionctrl: A unified and flexible motion controller for video generation},
  author={Wang, Zhouxia and Yuan, Ziyang and Wang, Xintao and Li, Yaowei and Chen, Tianshui and Xia, Menghan and Luo, Ping and Shan, Ying},
  booktitle={ACM SIGGRAPH 2024 Conference Papers},
  pages={1--11},
  year={2024}
}

@inproceedings{he2025cameractrl,
  title={Cameractrl ii: Dynamic scene exploration via camera-controlled video diffusion models},
  author={He, Hao and Yang, Ceyuan and Lin, Shanchuan and Xu, Yinghao and Wei, Meng and Gui, Liangke and Zhao, Qi and Wetzstein, Gordon and Jiang, Lu and Li, Hongsheng},
  booktitle={2025 IEEE/CVF International Conference on Computer Vision (ICCV)},
  pages={13416--13426},
  year={2025},
  organization={IEEE}
}

@article{xu2024camco,
  title={Camco: Camera-controllable 3d-consistent image-to-video generation},
  author={Xu, Dejia and Nie, Weili and Liu, Chao and Liu, Sifei and Kautz, Jan and Wang, Zhangyang and Vahdat, Arash},
  journal={arXiv preprint arXiv:2406.02509},
  year={2024}
}

@inproceedings{ren2025gen3c,
  title={Gen3c: 3d-informed world-consistent video generation with precise camera control},
  author={Ren, Xuanchi and Shen, Tianchang and Huang, Jiahui and Ling, Huan and Lu, Yifan and Nimier-David, Merlin and M{\"u}ller, Thomas and Keller, Alexander and Fidler, Sanja and Gao, Jun},
  booktitle={2025 IEEE/CVF Conference on Computer Vision and Pattern Recognition (CVPR)},
  pages={6121--6132},
  year={2025},
  organization={IEEE}
}

@inproceedings{wang2026bullettime,
  title={Bullettime: Decoupled control of time and camera pose for video generation},
  author={Wang, Yiming and Zhang, Qihang and Cai, Shengqu and Wu, Tong and Ackermann, Jan and Kuang, Zhengfei and Zheng, Yang and Raji{\v{c}}, Frano and Tang, Siyu and Wetzstein, Gordon},
  booktitle={Proceedings of the IEEE/CVF Conference on Computer Vision and Pattern Recognition},
  pages={18319--18330},
  year={2026}
}

@inproceedings{yu2025trajectorycrafter,
  title={Trajectorycrafter: Redirecting camera trajectory for monocular videos via diffusion models},
  author={Yu, Mark and Hu, Wenbo and Xing, Jinbo and Shan, Ying},
  booktitle={2025 IEEE/CVF International Conference on Computer Vision (ICCV)},
  pages={100--111},
  year={2025},
  organization={IEEE}
}

@article{wang2026omnishotcut,
  title={OmniShotCut: Holistic Relational Shot Boundary Detection with Shot-Query Transformer},
  author={Wang, Boyang and Xu, Guangyi and Zhang, Jiahui and Tang, Zhipeng and Cheng, Zezhou},
  journal={arXiv preprint arXiv:2604.24762},
  year={2026}
}

@article{huang2025vipe,
  title={Vipe: Video pose engine for 3d geometric perception},
  author={Huang, Jiahui and Zhou, Qunjie and Rabeti, Hesam and Korovko, Aleksandr and Ling, Huan and Ren, Xuanchi and Shen, Tianchang and Gao, Jun and Slepichev, Dmitry and Lin, Chen-Hsuan and others},
  journal={arXiv preprint arXiv:2508.10934},
  year={2025}
}

@inproceedings{veicht2024geocalib,
  title={Geocalib: Learning single-image calibration with geometric optimization},
  author={Veicht, Alexander and Sarlin, Paul-Edouard and Lindenberger, Philipp and Pollefeys, Marc},
  booktitle={European Conference on Computer Vision},
  pages={1--20},
  year={2024},
  organization={Springer}
}

@article{teed2021droid,
  title={Droid-slam: Deep visual slam for monocular, stereo, and rgb-d cameras},
  author={Teed, Zachary and Deng, Jia},
  journal={Advances in neural information processing systems},
  volume={34},
  pages={16558--16569},
  year={2021}
}

@article{team2026kimi,
  title={Kimi k2. 5: Visual agentic intelligence},
  author={Team, Kimi and Bai, Tongtong and Bai, Yifan and Bao, Yiping and Cai, SH and Cao, Yuan and Charles, Y and Che, HS and Chen, Cheng and Chen, Guanduo and others},
  journal={arXiv preprint arXiv:2602.02276},
  year={2026}
}

@inproceedings{rublee2011orb,
  title={ORB: An efficient alternative to SIFT or SURF},
  author={Rublee, Ethan and Rabaud, Vincent and Konolige, Kurt and Bradski, Gary},
  booktitle={2011 International conference on computer vision},
  pages={2564--2571},
  year={2011},
  organization={Ieee}
}

@article{kabsch1976solution,
  title={A solution for the best rotation to relate two sets of vectors},
  author={Kabsch, Wolfgang},
  journal={Foundations of Crystallography},
  volume={32},
  number={5},
  pages={922--923},
  year={1976},
  publisher={International Union of Crystallography}
}

@article{fischler1981random,
  title={Random sample consensus: a paradigm for model fitting with applications to image analysis and automated cartography},
  author={Fischler, Martin A and Bolles, Robert C},
  journal={Communications of the ACM},
  volume={24},
  number={6},
  pages={381--395},
  year={1981},
  publisher={ACM New York, NY, USA}
}

@inproceedings{kummerle2011g,
  title={g 2 o: A general framework for graph optimization},
  author={K{\"u}mmerle, Rainer and Grisetti, Giorgio and Strasdat, Hauke and Konolige, Kurt and Burgard, Wolfram},
  booktitle={2011 IEEE international conference on robotics and automation},
  pages={3607--3613},
  year={2011},
  organization={IEEE}
}

@article{grisetti2010tutorial,
  title={A tutorial on graph-based SLAM},
  author={Grisetti, Giorgio and K{\"u}mmerle, Rainer and Stachniss, Cyrill and Burgard, Wolfram},
  journal={IEEE Intelligent Transportation Systems Magazine},
  volume={2},
  number={4},
  pages={31--43},
  year={2010},
  publisher={IEEE}
}

@inproceedings{liu2024grounding,
  title={Grounding dino: Marrying dino with grounded pre-training for open-set object detection},
  author={Liu, Shilong and Zeng, Zhaoyang and Ren, Tianhe and Li, Feng and Zhang, Hao and Yang, Jie and Jiang, Qing and Li, Chunyuan and Yang, Jianwei and Su, Hang and others},
  booktitle={European conference on computer vision},
  pages={38--55},
  year={2024},
  organization={Springer}
}

@inproceedings{kirillov2023segment,
  title={Segment anything},
  author={Kirillov, Alexander and Mintun, Eric and Ravi, Nikhila and Mao, Hanzi and Rolland, Chloe and Gustafson, Laura and Xiao, Tete and Whitehead, Spencer and Berg, Alexander C and Lo, Wan-Yen and others},
  booktitle={2023 IEEE/CVF international conference on computer vision (ICCV)},
  pages={3992--4003},
  year={2023},
  organization={IEEE}
}

@inproceedings{teed2020raft,
  title={Raft: Recurrent all-pairs field transforms for optical flow},
  author={Teed, Zachary and Deng, Jia},
  booktitle={European conference on computer vision},
  pages={402--419},
  year={2020},
  organization={Springer}
}

\appendix


\section{Detailed Dataset Statistics}
\label{app:dataset_statistics}

\subsection{Released Dataset Statistics}
\label{app:release_stats}

The newly collected perspective-video pipeline begins with $173{,}303$ segmented candidates. Motion and media-quality filtering retains $142{,}844$ records, after which deduplication produces an intermediate pool of $134{,}880$ clips totaling $3{,}163.7$ hours. OCR filtering retains $88{,}835$ clips, while the independent pose-quality gate retains $109{,}846$ of the $134{,}169$ available trajectories. Intersecting the OCR, pose, and valid-caption pass lists yields $74{,}603$ dynamic clips. A separately curated low-motion branch contributes an additional $5{,}608$ clips, resulting in $80{,}211$ newly collected perspective clips.

Together with $47{,}699$ inherited Sekai clips and $982$ panoramic sequences, the final processed corpus contains $128{,}892$ video units totaling $2{,}826$ hours. These units preserve the native temporal structure of each source: perspective videos are represented as continuous clips of up to $120$ seconds, whereas panoramic recordings remain uncut and may substantially exceed this duration. Due to data-distribution agreements and licensing constraints, only approximately $20$ hours of the $119$-hour processed panoramic subset can be included in the current public release; the panoramic statistics reported in this paper are based on the full processed subset.

\subsection{Source Composition}
\label{app:scale_sources}

The three data sources exhibit complementary contributions in terms of both sample count and footage duration. The inherited Sekai subset accounts for $37.0\%$ of the released video units and $28.1\%$ of the total duration, with each clip lasting exactly $60$ seconds. The newly collected perspective subset contributes $62.2\%$ of the released units and $67.7\%$ of the total duration, with an average clip length of $85.8$ seconds.

The panoramic subset represents only $0.8\%$ of the released units but contributes $4.2\%$ of the total footage duration. Its disproportionately larger contribution by duration results from retaining the recordings as long, continuous sequences, with an average duration of $436$ seconds. Preserving these sequences without temporal truncation maintains complete routes, loops, and revisits that would otherwise be disrupted by conventional clip segmentation.

\subsection{Duration Distribution}
\label{app:scale_duration}

Because the released panoramic sequences can exceed the $120$-second limit used for perspective clips, we construct a common analysis representation for comparing temporal extent across the corpus. Specifically, all released videos are decomposed into non-overlapping analysis segments capped at $120$ seconds, yielding $131{,}876$ segments in total. This decomposition is used only for duration statistics and does not alter the released video units.

Among these analysis segments, $43{,}594$ reach the full $120$-second cap. They constitute $33.1\%$ of the segment count but contribute $1{,}453$ hours, corresponding to $51.4\%$ of the total corpus duration. More broadly, segments lasting at least $60$ seconds account for $92.3\%$ of all footage hours. In contrast, segments shorter than $60$ seconds represent $19.5\%$ of the segment count but contribute only $7.7\%$ of the total duration. The average duration of a released video unit is $78.9$ seconds. These statistics show that the temporal scale of \dataset is driven primarily by sustained observations rather than by the accumulation of short video snippets.

\begin{figure}[t!]
\centering
    \includegraphics[width=0.88\linewidth]{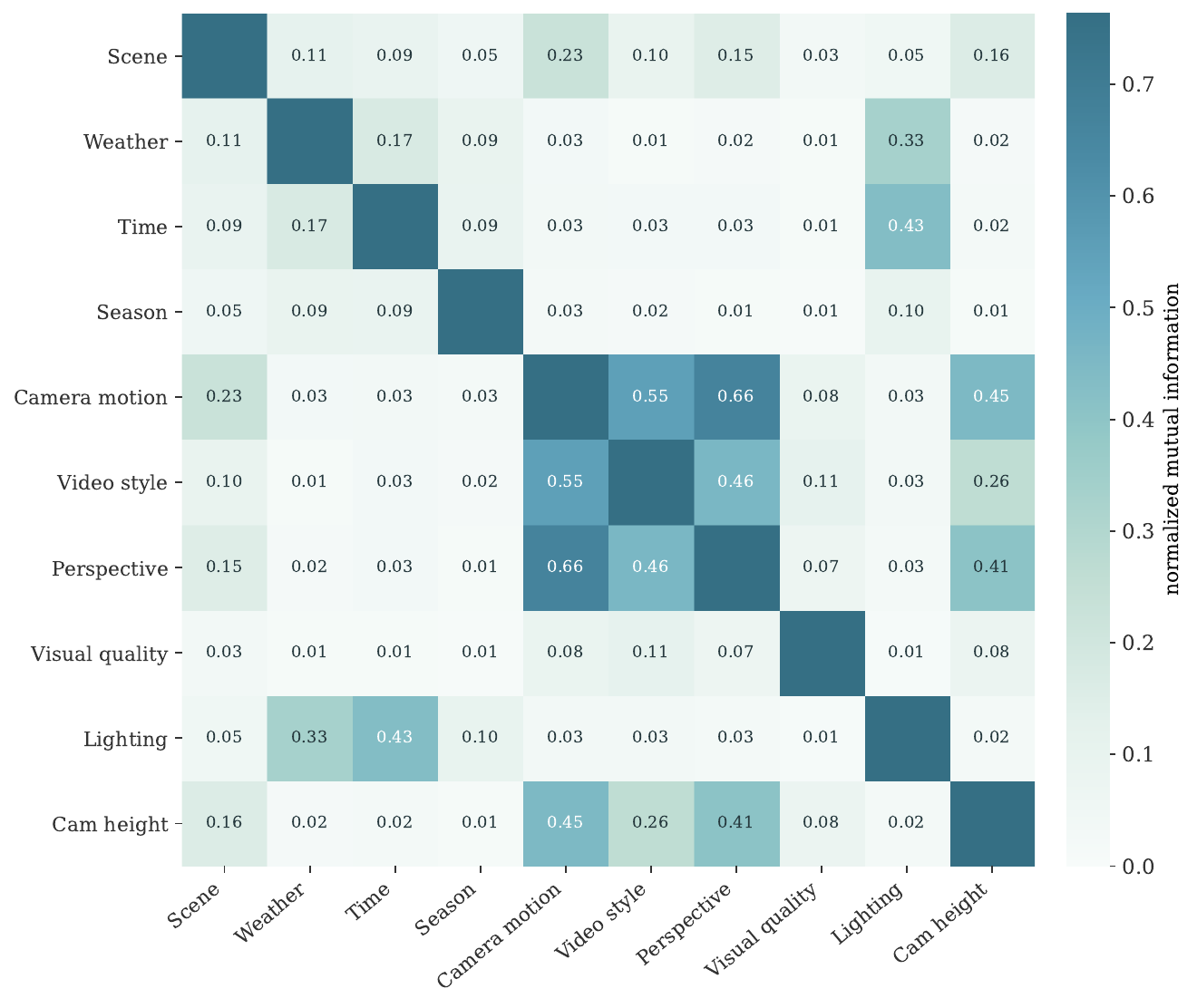}
    \vspace{3pt}
    \caption{\textbf{Attribute dependence.} Normalized mutual information between every pair of the ten clip-level attributes; the diagonal is self-information ($1$ by definition) and is drawn saturated. The colour scale follows the off-diagonal range so that the informative cells retain contrast.}
    \label{fig:attribute_mi}
\end{figure}

\subsection{Attribute Dependence}
\label{app:attribute_dependence}

Marginal diversity alone does not guarantee compositional diversity: a dataset may contain many distinct attribute values while realizing only a limited set of combinations if those attributes are strongly correlated. We therefore complement the marginal statistics in Sec.~\ref{sec:statistics_scene} by measuring normalized mutual information (NMI; $0$ denotes independence and $1$ indicates that one attribute determines the other) between every pair of the ten clip-level attributes. The analysis covers the $128{,}891$ released clips for which all ten attributes are resolved (Fig.~\ref{fig:attribute_mi}).

Across the $45$ unique attribute pairs, the mean NMI is $0.13$ and the median is $0.05$, with $31$ pairs ($69\%$) below $0.1$. Thus, most annotated properties vary largely independently: for example, knowing the scene type provides little information about a clip's weather, lighting, or camera behavior. The marginal attribute values therefore compose into a substantially richer set of realized conditions than their individual counts alone would suggest.
Two groups exhibit stronger dependence, both of which are semantically expected. Viewpoint-related attributes are mutually informative, with camera motion showing NMI values of $0.66$ with perspective, $0.55$ with video style, and $0.45$ with camera height, yielding a mean of $0.46$ within this group. This dependence reflects the acquisition process itself: aerial footage is typically captured from higher viewpoints, follows characteristic camera motions, and is associated with a distinct video style. Illumination-related attributes form a weaker second group, led by time of day with lighting ($0.43$) and weather with lighting ($0.33$), with a mean NMI of $0.20$.

Importantly, the viewpoint and illumination groups remain nearly independent of each other. Across the $16$ cross-group pairs, the mean NMI is only $0.024$ and the maximum is $0.034$. This indicates that viewpoint, motion, and appearance conditions are combined across the corpus rather than being restricted to a small number of recurring acquisition patterns.

\section{Dataset Construction Details}
\label{app:dataset_details}

This appendix specifies the implementation used to construct \dataset. We report the stage order, decision rules, thresholds, output formats, and failure handling needed to reproduce the released manifest. The panoramic pipeline in Appendix~\ref{sec:panoramic_captures} is retained as a separate treatment because its camera model and global trajectory refinement differ from those of perspective video.

\subsection{Temporal Segmentation}
\label{app:segmentation}

The input loader first probes every source with ffprobe and rejects undecodable files or records with invalid duration, frame rate, resolution, or codec metadata. We then apply OmniShotCut~\cite{wang2026omnishotcut} to detect both hard cuts and gradual transitions. For each internal boundary, we remove a $6$-second buffer on either side; no buffer is removed at the physical start or end of a source video. Each remaining continuous interval is partitioned into clips of at most $120$ seconds, and residual intervals shorter than $10$ seconds are discarded. The cutter stores the source identifier, absolute frame range, local temporal offsets, frame rate, resolution, and media path in the manifest. Clip materialization uses ffmpeg stream-copy when the source codec permits exact cutting; re-encoding is used only for codec normalization or frame-accurate boundaries. This stage applies to the perspective sources only; self-captured panoramic videos are single continuous takes without editing cuts, and are carried through the pipeline uncut so that their loops and revisits remain within one sequence (Appendix~\ref{sec:panoramic_captures}).

This procedure is intentionally \emph{shot aware but duration preserving}: it avoids training across edits while preferentially retaining the longest continuous interval allowed by the $120$-second training budget. The resulting file name encodes the absolute start and end frames, which provides a deterministic key for deduplication and for joining video, pose, caption, and filter artifacts.

\subsection{Filtering Framework}
\label{app:filtering_framework}

The Prism filtering framework is manifest driven. A stage consumes a manifest row, writes cached scores and a versioned verdict, and never mutates the source video. File-system caches permit interrupted jobs to resume and allow thresholds to be swept without recomputing optical flow or OCR. Hard gates determine inclusion; auxiliary stages retain scores for later analysis and sampling. The released perspective subset applies the gates described below and takes their intersection.

\paragraph{Full-clip motion estimation.}
We uniformly sample at most $30$ frames across the \emph{entire} clip, rather than taking only an initial window. Frames are resized so that the long side is $256$ pixels and passed to RAFT-Small~\cite{teed2020raft}. For each adjacent sampled pair $i$, we compute the mean optical-flow magnitude $m_i$. A pair is considered static when $m_i<1.0$, and the clip-level statistic is
\begin{equation}
    r_{\mathrm{static}}=\frac{1}{T-1}\sum_{i=1}^{T-1}
    \mathbb{I}[m_i<1.0].
\end{equation}
A clip fails the strict motion gate when $r_{\mathrm{static}}\geq0.5$. The same RAFT forward pass also caches the mean flow vector, temporal variance, directional variance, peak magnitude, active-pair fraction, luminance, frame differences, and border summaries. These features feed subsequent rules without decoding the video again.

\paragraph{Media integrity.}
The format gate requires a height of at least $720$ pixels, a frame rate in $[24,65]$ fps, and an H.264, HEVC/H.265, or AV1 codec. On uniformly sampled luminance frames, values below $16$ or above $240$ are marked as under- or over-exposed; a clip is rejected if either condition persists for more than $15$ sampled frames. Black borders are found from rows or columns with both low mean intensity and low variance. The maximum combined vertical or horizontal border ratio must not exceed $0.10$. Freeze statistics are recorded from consecutive-frame luminance differences, but the freeze rule is not a hard gate in the released configuration because it produced false positives on legitimate low-motion content.

\paragraph{HUD detection.}
Twelve frames are resized to $640\times360$. Let $\sigma(x)$ and $\mu(x)$ denote the temporal standard deviation and mean luminance at pixel $x$, and let $g(x)$ be the Sobel edge magnitude of the temporal mean image. A candidate overlay pixel satisfies $\sigma(x)<5$, $\mu(x)>32$, and $g(x)>30$. We apply a $3\times3$ morphological opening to suppress isolated responses and reject a clip when the surviving mask covers more than $4\%$ of the frame. This criterion targets temporally fixed UI elements while avoiding textured but moving scene content.

\paragraph{OCR, subtitles, and watermarks.}
EasyOCR is run on eight uniformly sampled frames. Text boxes shorter than $2\%$ of the image height are treated as noise. A box centered in the lower $22\%$ of the frame and within the horizontal interval $[0.1W,0.9W]$ is classified as a subtitle; a box whose center lies within a $15\%$ corner margin is a watermark candidate. We reject a clip when subtitles occur in at least two sampled frames, when a corner box with intersection-over-union above $0.6$ persists for at least three frames, or when the detected text area exceeds $2\%$ of a frame. HUD and OCR depend only on the basic media gate, not on the motion verdict, so that clips entering the low-motion recovery branch are checked by the same overlay rules.

\paragraph{Auxiliary motion descriptors.}
The mean flow and directional variance assign a coarse \texttt{static}, \texttt{pan}, \texttt{mixed}, or \texttt{gameplay} descriptor. Directional variance is defined as one minus the mean resultant length of per-pair flow angles; values at most $0.25$ indicate consistent motion and values at least $0.55$ indicate highly varying motion. These labels, together with action saliency, temporal variance, and mean angular jitter, are retained for analysis and balanced sampling rather than used as additional hard gates.

\paragraph{Low-motion recovery.}
The strict motion gate is applied first to construct a motion-rich main set. We then consider clips whose only failed verdict is static motion and that pass all media, HUD, and OCR checks. Candidates are grouped by $r_{\mathrm{static}}$ in intervals of width $0.1$ over $[0.2,1.0]$. Selection proceeds round-robin across bins, with deterministic within-bin ordering obtained from a SHA-256 hash of the clip identifier and a fixed seed. The budget is measured in hours rather than clip count: for a desired final fraction $\rho$, the recovered duration is $H_{\mathrm{main}}\rho/(1-\rho)$; we use $\rho=0.06$. The recovered clips are subsequently subjected to pose and caption checks and are stored as a separately identifiable low-motion subset.

\subsection{Camera Pose Generation}
\label{app:pose_generation}

Perspective clips are annotated with the pose-only configuration of ViPE~\cite{huang2025vipe}. This configuration preserves the complete SLAM path while disabling dense-depth alignment and visualization, the two expensive outputs not required by camera-conditioned video training.

\paragraph{Calibration and dynamic masking.}
GeoCalib~\cite{veicht2024geocalib} estimates a shared pinhole field of view from three frames sampled near the start of the sequence. The implied $(f_x,f_y,c_x,c_y)$ parameters initialize the camera, and bundle adjustment refines intrinsics when calibration is not provided. Dynamic regions interfere with rigid-scene flow and are therefore removed before SLAM. Grounding DINO~\cite{liu2024grounding} detects a fixed prompt set including people, animals, and vehicles; SAM~\cite{kirillov2023segment} segments the detections; and a video tracker propagates masks between detection keyframes. Masks are eroded by five pixels to reduce boundary contamination, while sky is represented separately.

\paragraph{Trajectory optimization.}
DROID-SLAM~\cite{teed2021droid} performs sequential initialization, keyframe selection, local refinement, and global backend optimization. The deployed configuration uses an eight-frame warm-up, a keyframe threshold of $4.0$, a frontend window of $25$ frames, and $24$ global backend iterations. UniDepth supplies keyframe depth priors, which regularize geometry and scale without requiring dense depth to be saved. Source videos above $40$ fps are processed with stride two and other videos with stride one, producing poses aligned at approximately $30$ fps for the dominant source formats.

\paragraph{Outputs and distributed execution.}
Each successful clip writes \texttt{pose/<clip>.npz} and \texttt{intrinsics/<clip>.npz}. The pose array has shape $N\times4\times4$ and stores camera-to-world transformations; the accompanying index array records the source frames. Workers deterministically partition a globally sorted clip list into non-overlapping contiguous shards. Existing pose files are treated as completed outputs, allowing a failed or interrupted distributed run to resume without recomputing successful clips.

\subsection{Panoramic Captures}
\label{sec:panoramic_captures}

\paragraph{Panoramic Data Collection.}
We collect 1,283 panoramic videos using Insta360 cameras, amounting to approximately 160 hours of footage. The videos are captured primarily across Shanghai, Hangzhou, and Jinan (including the Zhangqiu district), with a small number collected in other locations. The captures span diverse indoor and outdoor environments, including urban streets and plazas, residential communities, campuses, parks, scenic areas, and shopping malls. They further exhibit varied camera motions and elevation changes, such as non-linear walking paths, sharp turns, obstacle avoidance, and stair traversal. Whenever feasible, the camera operator returns to the starting location near the end of each capture, producing an explicit start-to-end loop. Some videos additionally contain mid-trajectory revisits, such as repeated passages through the same area in shopping malls.

\paragraph{Initial Trajectory Estimation.}
The captured videos are equirectangular panoramas, which we downscale to $960\times480$ and process using the panorama mode of ViPE~\cite{huang2025vipe} to obtain per-frame camera poses as the initial trajectories. The sequences have an average duration of approximately 7.5 minutes; at roughly 30 frames per second, each sequence contains about 13K frames and covers a substantial spatial extent. Although the initial estimates generally preserve local camera motion, drift accumulated over such long sequences can cause substantial global inconsistency. In particular, frames captured at the same physical location may remain widely separated in the reconstructed trajectory. We therefore exploit the revisits introduced during data collection to refine the initial poses.

\paragraph{Loop-Closure Trajectory Refinement.}
We construct loop-closure candidates from temporally distant keyframe pairs. Candidate generation primarily relies on ORB feature matching~\cite{rublee2011orb} between keyframes from the beginning and ending portions of each video, with ambiguous correspondences filtered using Lowe's ratio test. We supplement these candidates with spatially proximal keyframe pairs from intermediate portions of the initial trajectory. A central challenge is that the camera may revisit the same location with an arbitrary heading. Although the complete azimuthal coverage of panoramic imagery facilitates appearance matching, visual similarity alone cannot reliably distinguish a genuine revisit from two nearby but distinct camera positions. We therefore geometrically verify each candidate in spherical space. Specifically, we convert the matched feature pairs from equirectangular image coordinates into unit viewing rays $(\mathbf{r}_k,\mathbf{r}'_k)$ and estimate the rotation that best aligns the two sets, which admits a closed-form solution via singular value decomposition~\cite{kabsch1976solution}:
\begin{equation}
    \mathbf{R}^{*}
    =
    \arg\min_{\mathbf{R}\in\mathrm{SO}(3)}
    \sum_k
    \left\|
        \mathbf{r}'_k-\mathbf{R}\mathbf{r}_k
    \right\|_2^2.
    \label{eq:panoramic_loop_alignment}
\end{equation}
We use the median angular residual $\theta_{\mathrm{med}} =\operatorname{median}_k \angle(\mathbf{R}^{*}\mathbf{r}_k,\mathbf{r}'_k)$ as the co-location measure. For a genuine co-location, the difference between the two panoramic observations can be explained predominantly by a single global rotation. In contrast, a change in camera center induces parallax that cannot be eliminated by rotation alone. We consequently accept a loop closure only when enough correspondences are geometrically consistent~\cite{fischler1981random} and the ray-alignment residual is low, thereby suppressing visually similar but geometrically inconsistent matches.

Given the verified closures, we refine each trajectory through pose-graph optimization (PGO)~\cite{kummerle2011g,grisetti2010tutorial}. The graph takes keyframe poses as nodes and uses the ViPE relative motions between adjacent keyframes as odometry factors. The rotational and translational residuals in these factors are assigned square-root information weights of 50 and 1, respectively, reflecting greater confidence in the local orientation estimates. Since revisiting a location does not imply returning with the same viewing direction, each loop-closure factor constrains the corresponding camera centers to coincide without forcing their orientations to match. After optimizing the keyframe poses, we propagate the resulting correction transformations to every video frame through interpolation on $\mathrm{SE}(3)$. Among the 1,283 trajectories, 1,051 (81.9\%) are refined using geometrically verified loop closures; those without a reliable closure retain their original ViPE poses to avoid spurious global constraints. The procedure also handles mid-trajectory revisits rather than only start-to-end returns, including 25 such sequences. Full-accumulation reconstruction examples are analyzed in Sec.~\ref{app:panoramic_reconstruction_cases}.

\subsection{VLM Captioning and Prompt Design}
\label{app:caption_schema}

Annotations are generated by Kimi-K2.6~\cite{team2026kimi} through an OpenAI-compatible service. We adopt a unified annotation schema and merged inference pipeline that produces global attributes and temporally grounded segments in a single request, avoiding redundant encoding of the same visual input.

\paragraph{Visual input construction.}
Videos are first resized to $360$p. We sample at a nominal rate of $1$ fps, capped at $64$ frames; when the cap is reached, indices are recomputed with \texttt{linspace} over the complete source interval so that the last sample remains close to the physical end of the clip. Consecutive samples are grouped in pairs and interleaved with their starting timestamps in \texttt{mm:ss} format. The prompt also states the measured duration and the first and last timestamps explicitly. These anchors prevent the model from terminating its segment description near one minute when processing a two-minute clip.

\paragraph{Global prompt.}
The global annotation prompt requests normalized attributes describing the scene, environment, and camera configuration, including weather, time of day, season, lighting, visibility, location type, geographic information, viewpoint, camera motion, camera height and stability, estimated speed, video style, and quality. Geographic attributes are inferred only when supported by evidence; otherwise, the model is instructed to return \texttt{unknown}. All categorical attributes are constrained to predefined vocabularies.

In addition, the prompt produces six global text fields: \texttt{subject\_motion}, \texttt{environment\_motion}, \texttt{static\_scene}, \texttt{camera\_description}, \texttt{full\_prompt}, and \texttt{short\_prompt}. The first four fields factorize the video content into subject dynamics, environmental dynamics, persistent scene context, and camera behavior, respectively. The \texttt{full\_prompt} integrates these components into a detailed camera-aware description, whereas the \texttt{short\_prompt} provides a compact scene-centric description with camera-specific information removed. Together, these fields support both structured supervision and training-ready conditioning at different levels of semantic and camera detail.

\paragraph{Temporal segmentation prompt.}
The model is asked to introduce a boundary when a salient element enters or leaves, the scene type or ground surface changes, illumination changes, or the camera changes its dominant behavior. A segment is not created for a single incidental pedestrian or a minor distant event. The target density is one segment per $10$--$20$ seconds, with a nominal range of $5$--$45$ seconds per segment. The first interval must start at zero, the last must reach the final timestamp, and consecutive intervals must tile the video without gaps. Before emitting JSON, the prompt asks the model to construct a timestamped element inventory as an internal scratchpad. This inventory encourages consistent object tracking across intervals but is not part of the released training schema.

For every segment, \texttt{subject\_motion} owns the main agent's activity; \texttt{environment\_motion} describes other moving entities and time-varying conditions; \texttt{static\_scene} records persistent layout, materials, and appearance; and \texttt{camera\_description} independently describes viewpoint, framing, path, and stability. The prompt explicitly warns that subject and camera motion need not agree---for example, a stationary subject may be observed by an orbiting camera.

\paragraph{Camera-path vocabulary.}
The discrete \texttt{camera\_path} field contains $16$ values. Translation comprises \texttt{straight}, \texttt{reverse}, \texttt{strafe\_left}, \texttt{strafe\_right}, \texttt{ascend}, and \texttt{descend}; coupled translation and rotation comprises \texttt{turn\_left}, \texttt{turn\_right}, \texttt{orbit\_left}, and \texttt{orbit\_right}; pure rotation comprises \texttt{pan\_left}, \texttt{pan\_right}, \texttt{tilt\_up}, and \texttt{tilt\_down}; and the two special states are \texttt{stationary} and \texttt{mixed}. The model selects the path occupying more than $70\%$ of an interval when one exists and uses \texttt{mixed} only when no single behavior dominates. Prompt-level visual heuristics distinguish yaw from lateral translation through common motion versus parallax, orbiting through a centered subject and moving background, and forward or reverse translation through scale change.

\paragraph{Training-prompt constraints.}
The \texttt{full\_prompt} must be coherent present-tense prose rather than a concatenation of field labels. All camera-specific sentences must occur inside exactly delimited \texttt{<camera>...</camera>} blocks. The \texttt{short\_prompt} summarizes only the subject, action, and setting; it excludes camera tags and camera-specific vocabulary. The prompt requests $15$--$45$ words for this compact description, while the automatic validator uses the more tolerant interval $8$--$60$ to avoid rejecting otherwise valid borderline outputs. This construction supports three training modes without regenerating captions: full semantic and camera conditioning, semantic conditioning with camera clauses removed, and compact caption conditioning.

\paragraph{Parsing, validation, and fault tolerance.}
The model emits global and segment JSON between explicit sentinel markers. The parser first extracts these spans, then falls back to a fenced JSON block and finally to the outermost balanced braces. Global validation checks every enum, required string, and numeric field. Segment validation checks a non-empty segment list, consecutive indices, numeric and increasing time ranges, a maximum one-second boundary mismatch, non-empty factorized fields, and membership in the camera-path vocabulary. It additionally rejects field-label prose, missing camera tags, camera terms outside the tags, camera leakage into the short prompt, and short prompts outside $8$--$60$ words.

We use temperature $0.1$ and a configured output budget of $16{,}384$ tokens; merged mode allocates additional headroom for the combined global and segment JSON. API failures are retried up to three times with exponential backoff and a $600$-second request timeout. File-based distributed locks prevent two workers from annotating the same clip, expire after two hours, and make completed outputs resumable. Each JSON stores the server-reported model identifier, resolution, sampling rate, number of VLM calls, elapsed time, processing host, and separate global/segment validation errors in \texttt{annotation\_meta}. Raw responses and the scratchpad inventory can be retained for auditing; downstream loaders can enforce stricter exclusion policies directly from these validation flags.

\begin{figure*}[t]
\centering
\begin{minipage}{\textwidth}
    \begin{lstlisting}[basicstyle=\fontsize{7.2pt}{8.2pt}\selectfont\ttfamily,
    breaklines=true,frame=single]
    ROLE
    You are annotating a video clip for a world-model training dataset. Watch the entire video once and return two JSON blocks: (1) overall attributes and (2) temporally grounded scene segments.
    
    INPUT
    Video duration: [DURATION] seconds. Sampled frames are paired with timestamps [TIMESTAMPED_FRAMES]. Inspect the complete interval, including the final timestamp.
    
    PART 1 -- OVERALL_JSON_START ... OVERALL_JSON_END
    Return controlled attributes for weather, time of day, season, lighting, visibility, location, camera perspective/motion/height/stability, speed, video style, and visual quality. Also return: subject_motion, environment_motion, static_scene, camera_description, full_prompt, short_prompt.
    Infer geography only from visible evidence; otherwise use "unknown".
    
    PART 2 -- SEGMENTS_JSON_START ... SEGMENTS_JSON_END
    Create a boundary when a salient element enters/exits, scene or surface changes, illumination changes, or dominant camera behavior changes.
    Target one segment per 10--20 s (normally 5--45 s each). Segments must tile [0, DURATION] without gaps. For each segment return:
    segment_index, time_range_s, subject_motion, environment_motion,
    static_scene, camera_description, camera_path, full_prompt, short_prompt.
    
    SEMANTIC FACTORIZATION
    
    * subject_motion: action of one consistently named protagonist.
    * environment_motion: other actors and time-varying world state.
    * static_scene: persistent layout, materials, objects, and appearance.
    * camera_description: viewpoint, framing, path, and stability, described independently from subject motion.
    * camera_path: exactly one of {straight, reverse, strafe_left, strafe_right, ascend, descend, turn_left, turn_right, orbit_left, orbit_right, pan_left, pan_right, tilt_up, tilt_down, stationary, mixed}.
    
    TRAINING-TEXT RULES
    
    * full_prompt: coherent present-tense prose. Put every camera-specific sentence inside <camera>...</camera>; keep camera language outside absent.
    * short_prompt: 1--2 sentences and 15--45 words covering subject, action, and scene only. Do not use camera/framing terms or camera tags.
    
    SELF-CHECK
    Use only legal enums; keep the POV agent consistent; distinguish subject from camera motion; reserve "mixed" for intervals with no dominant path; cover the complete video; emit valid JSON without Markdown fences.
    \end{lstlisting}
    \end{minipage}
    \vspace{3pt}
    \caption{\textbf{Caption annotation prompt used for hierarchical semantic annotation.}
    The prompt jointly produces clip-level attributes and temporally grounded segments under a unified schema. It factorizes subject motion, environmental dynamics, static scene content, and camera behavior, while providing both camera-aware \texttt{full\_prompt} and camera-free \texttt{short\_prompt} descriptions. The complete production prompt additionally includes the full controlled vocabularies, examples, and validation instructions.}
    \label{fig:caption_prompt}
\end{figure*}
\FloatBarrier

\subsection{Caption Annotation Prompt}
\label{app:caption_prompt}

For reproducibility, we summarize the production prompt used to generate the hierarchical annotations in Fig.~\ref{fig:caption_prompt}. Text in brackets denotes video-specific inputs inserted by the request builder. The summarized prompt captures the main annotation structure, including clip-level attributes, temporally grounded segments, semantic factorization, camera-path labeling, and the construction of camera-aware and camera-free training prompts.

The complete production prompt additionally specifies the full controlled vocabularies, field-level positive and negative examples, motion-disambiguation rules, an internal element-inventory scratchpad, sentinel syntax, and a self-verification checklist. We release the complete prompt together with the annotation code to support faithful regeneration of the caption corpus. Qualitative examples of the resulting hierarchical annotations are presented in Appendix~\ref{app:multimodal_case}.

\subsection{Annotation Quality Assurance}
\label{app:quality_assurance}

\paragraph{Pose validity and residual construction.}
Every NPZ file must contain a non-empty finite pose array under \texttt{data} or \texttt{cam\_c2w}. For pose $(\mathbf{R}_t,\mathbf{t}_t)$, we construct a forward reference point one unit along the optical axis,
\begin{equation}
    \mathbf{p}_t=\mathbf{R}_t[0,0,1]^{\mathsf T}+\mathbf{t}_t,
\end{equation}
and compare it with a coordinate-wise median-filtered trajectory $\widetilde{\mathbf{p}}_t$ using a $61$-frame window. The residual $e_t=\|\mathbf{p}_t-\widetilde{\mathbf{p}}_t\|_2$ responds to both translation and rotation while remaining insensitive to a smooth turn. A trajectory is declared degenerate when fewer than $90\%$ of the flattened forward-reference coordinates are distinct.

Residual peaks above $0.07$ are analyzed in $60$-frame neighborhoods. A neighborhood with a single extremum is treated as a plausible turn; repeated extrema indicate jitter. We report the largest residual $e_{\max}$, the residual median, and a normalized jitter rate $1800N_{\mathrm{jitter}}/N$. For sequences long enough to support it, a second $901$-frame median filter yields the $95$th-percentile drift residual. We also compute path length, endpoint displacement, and straightness from the translation component.

\paragraph{Pose decision rule.}
Four residual statistics are normalized between empirically chosen good and bad anchors: $e_{\max}$ between $0.15$ and $2.0$ on a logarithmic scale, jitter rate between $1$ and $15$, median residual between $0.002$ and $0.02$, and drift $p95$ between $2$ and $15$ on a logarithmic scale. Missing long-window drift is assigned a neutral sub-score of $0.5$. The final score is
\begin{equation}
    q_{\mathrm{pose}}=0.3q_{\mathrm{peak}}+0.3q_{\mathrm{jitter}}
    +0.2q_{\mathrm{median}}+0.2q_{\mathrm{drift}},
\end{equation}
and the clip is retained when $q_{\mathrm{pose}}<0.5$. Hard rules additionally reject $e_{\max}>5$, jitter rate above $40$, or path length below $0.5$. These are the calibrated thresholds used for the released real-world trajectories; alternative source domains should recalibrate the normalization anchors rather than transfer them blindly.

\paragraph{Caption structural audit.}
After online validation, we run a corpus-level audit over the serialized JSON. It checks UTF-8 decoding, schema version, the presence and type of every global field, a non-empty segment list, continuous time ranges, and non-empty values for all six text fields. It also verifies that camera tags are balanced, that camera-specific terms do not leak into the short prompt, and that no timestamp, segment index, or scratchpad marker leaks into training prose. Length checks flag truncation and abnormally short responses without forcing every valid caption to a fixed template.

We measure exact duplication of the global description and segment prompts across clips. Because exact matching alone can miss lightly edited templates, the audit also supports token-shingle Jaccard or MinHash checks for near duplicates. Operational failures such as timeouts, missing sentinel blocks, invalid enums, or partial segment output remain stored in \texttt{annotation\_meta}; the released manifest is obtained only after joining available caption artifacts and non-empty segments with the media, OCR, and pose pass lists.

\paragraph{Cross-modal audit.}
We derive a coarse motion shape independently from each pose trajectory and map the VLM-produced \texttt{camera\_path} labels into the same static, straight, curved, and winding taxonomy. Agreement is computed only when both modalities admit a valid mapping. This check detects systematic caption direction errors and pose degeneracy at corpus scale. It is deliberately reported separately from metric trajectory accuracy: monocular ViPE poses have an unknown global translation scale, and agreement between two predicted modalities is not equivalent to external ground truth.

\subsection{Configuration Summary}
\label{app:configuration_summary}

Table~\ref{tab:filter_configuration} consolidates the release configuration. Values are listed here to distinguish the actual release policy from optional stages and alternative settings retained in the codebase.

\begin{table*}[t]
    \centering
    \caption{\textbf{Release configuration for perspective-video curation.} ``Aux.'' denotes a recorded descriptor that is not itself a hard gate.}
    \label{tab:filter_configuration}
    \tablestyle{3.5pt}{1.12}
    \begin{tabular}{lll}
        \shline
        Component & Configuration & Decision \\
        \shline
        Segmentation & OmniShotCut; $6$ s boundary buffer & retain $10$--$120$ s intervals \\
        Motion & RAFT-Small; $30$ uniform frames; long side $256$ & reject $r_{\mathrm{static}}\geq0.5$ \\
        Format & height $\geq720$; fps $[24,65]$; H.264/HEVC/AV1 & hard gate \\
        Exposure & luminance $\leq16$ or $\geq240$ & reject run $>15$ samples \\
        Black border & low-mean, low-variance rows/columns & reject ratio $>0.10$ \\
        HUD & $12$ frames at $640\times360$ & reject stable-edge area $>0.04$ \\
        OCR & $8$ frames; EasyOCR & subtitle $\geq2$ / watermark $\geq3$ frames \\
        Flow descriptors & direction variance, saliency, temporal variance & Aux. \\
        Static recovery & duration budget; eight motion bins & target $6\%$ of final hours \\
        Pose score & peak/jitter/median/drift residuals & reject weighted score $\geq0.5$ \\
        Caption & Kimi-K2.6; $360$p; up to $64$ timestamped frames & schema + non-empty segments \\
        \shline
    \end{tabular}
\end{table*}

\subsection{Released Data Organization}
\label{app:data_organization}

\paragraph{Canonical manifest.}
The release is enumerated by a single CSV manifest with one row per final clip. The \texttt{dataset} field identifies the inherited, newly collected, recovered-static, or panoramic source group. \texttt{clip\_name} is the stable join key, and \texttt{video\_id} identifies the source video used for leakage-free splitting. \texttt{video\_ref}, \texttt{pose\_path}, and \texttt{caption\_path} locate the three modalities. All dataset statistics are generated by iterating this manifest rather than scanning storage directories, which prevents stale or intermediate artifacts from entering reported results.

\paragraph{Pose files.}
Each compressed NPZ contains \texttt{data}, an $N\times4\times4$ float array of camera-to-world transformations, and \texttt{inds}, the corresponding source-frame indices. Camera positions are given by \texttt{data[:, :3, 3]}; the $Y$ axis denotes the vertical direction, while the $X$--$Z$ plane is used for bird's-eye trajectory visualization. Because monocular reconstruction is defined only up to a global scale, the translation components should be interpreted as relative trajectory coordinates rather than metric distances, and trajectories should be aligned before metric comparison.

\paragraph{Caption files.}
Each UTF-8 JSON contains \texttt{version}, source identifiers, an \texttt{overall} object, a \texttt{segments} list, and \texttt{annotation\_meta}. A segment stores its index, temporal range, four factorized descriptions, discrete camera path, full prompt, and short prompt. The metadata records the actual server model identifier, processing resolution and sampling rate, VLM-call count, elapsed time, and validation-error arrays. Keeping these operational fields in the public schema allows users to impose stricter quality policies without regenerating annotations.

\subsection{Annotation Compute and Reproducibility}
\label{app:compute_reproducibility}

Caption annotation is performed using a distributed VLM inference service deployed on 48 compute nodes with 8 GPUs per node, for a total of 384 H20 GPUs. Global and temporally grounded annotations are generated within a merged inference request for each clip. The serving system uses dynamic request scheduling to maintain high accelerator utilization under variable generation lengths and occasional retries.

During production runs, the distributed service sustains an aggregate throughput of approximately $800$--$960$ clips per hour, with short-term variations caused by service latency, retrying invalid or incomplete outputs, and fluctuations in request scheduling. With the distributed deployment described above, completing the full annotation pipeline requires approximately $6$--$7$ days of wall-clock time under sustained operation. These measurements reflect end-to-end serving throughput rather than pure model computation.
All filtering stages store their runtime configuration, stage version, per-clip scores, and verdict reasons. Pose and caption jobs use deterministic sharding and recognize existing valid outputs, allowing interrupted distributed runs to resume without recomputing completed samples.

For reproducibility, we preserve four layers of provenance: (i)~immutable source and clip identifiers; (ii)~the exact processing configuration and software revision; (iii)~per-stage scores and raw model outputs; and (iv)~the final versioned intersection manifest. Threshold changes can therefore be applied by re-sweeping cached scores and generating a new manifest version rather than recomputing the complete pipeline.

\subsection{Known Failure Modes and Responsible Use}
\label{app:limitations_data}

The construction pipeline reduces common artifacts but cannot guarantee that every frame is free of personal information, copyrighted material, or small text missed by OCR. Public release should therefore preserve source provenance and follow the applicable source licenses, takedown procedures, and privacy review. Models trained on the corpus may inherit geographic and capture biases; in particular, country coverage is broad but not uniform, and walking videos dominate several motion and scene categories.

VLM annotations are observations inferred from sampled frames. Fine temporal events can be missed between samples, geographic fields can be uncertain, and apparently static scene properties may change outside the observed interval. Validation flags and \texttt{unknown} labels should be respected rather than treated as missing values to be automatically imputed. Similarly, ViPE trajectories are estimated rather than instrumented: dynamic foregrounds, low texture, reflections, rapid rotation, and long loops can still cause drift. Monocular scale ambiguity makes the poses suitable for relative camera control and geometric supervision but requires alignment for metric evaluation. The external pose and caption studies listed in \texttt{analysis/ADDITIONAL\_EXPERIMENTS.md} are designed to quantify these residual limitations.

\section{Additional Camera-Trajectory Evaluation}
\label{app:pose_cross_validation}

This appendix provides additional protocols and analyses for the camera-trajectory quality evaluation in Sec.~\ref{sec:exp_pose}. The corpus-wide validity and smoothness statistics are computed directly over all released trajectories and are fully reported in Table~\ref{tab:pose_intrinsic}. We therefore focus here on the two geometry-based evaluations: the ground-truth-free epipolar test and the cross-run comparison with DROID-SLAM~\cite{teed2021droid}. The former evaluates local pose--frame consistency directly from image correspondences, whereas the latter examines the stability of relative trajectory geometry across reconstruction runs, motion regimes, and intrinsic-calibration settings.

\subsection{Ground-Truth-Free Epipolar Validation}
\label{app:pose_epipolar_details}

\paragraph{Evaluation protocol.}
We evaluate whether the released poses explain image correspondences without requiring metric trajectory ground truth. For each sampled perspective clip, we extract SIFT matches between frames separated by $0.5$\,s, construct the fundamental matrix implied by the released poses and intrinsics, and compute the Sampson error. To reduce the influence of mismatches and independently moving objects, we discard the largest $20\%$ of errors within each frame pair before aggregation.

The inherited subset is evaluated using its released per-frame intrinsics. For the newly collected subset, the released intrinsics are obtained from the same GeoCalib estimates used during reconstruction. Directly reusing these intrinsics in the epipolar evaluation could therefore partially reuse the calibration being evaluated. We instead adopt a fixed prior focal length as the primary conservative protocol and report a per-clip focal sweep only as a sensitivity analysis.

\paragraph{Mismatched-pose negative control.}
To determine whether low Sampson errors genuinely depend on pose--frame alignment, we construct a negative control that changes only the associated pose pair. For an image pair indexed by $(a,b)$, we select another pose pair from the same trajectory:
\begin{equation}
    a'=
\left(
a+\left\lfloor N/3\right\rfloor
\right)
\bmod (N-1),
\qquad
b'=
\min\!\left(
a'+(b-a),N-1
\right).
\end{equation}

The control therefore retains the same two images, SIFT correspondences, intrinsics protocol, temporal separation, and robust trimming, but replaces $(\mathbf{T}_a,\mathbf{T}_b)$ with the temporally shifted pair $(\mathbf{T}_{a'},\mathbf{T}_{b'})$. The shifted poses remain valid poses from the same trajectory and preserve a comparable temporal baseline; only their correspondence with the evaluated image pair is broken.

\paragraph{Results.}
The inherited subset achieves a median per-clip Sampson error of $0.12$\,px over $500$ clips, with all clips below $5$\,px. Under the conservative fixed-focal protocol, the $399$ newly collected clips obtain a median error of $0.43$\,px, with $99.0\%$ below $5$\,px and all clips below $10$\,px.

Under the mismatched-pose control, the median error rises to $2.83$\,px and only $74.0\%$ of clips remain below $5$\,px. Relative to the correctly aligned results, the control error is approximately $23.6\times$ larger than that of the inherited subset and $6.6\times$ larger than the conservative fixed-focal result for the newly collected subset. Because all image-side inputs remain unchanged, this degradation isolates the effect of breaking pose--frame correspondence and confirms that the metric is sensitive to the released camera geometry.

\paragraph{Focal-length sensitivity.}
Allowing each newly collected clip to select its best focal length from a predefined sweep reduces the median Sampson error from $0.43$\,px to $0.14$\,px. Because per-clip selection can favor the focal value most compatible with the observed correspondences, we treat this result as a sensitivity analysis rather than the primary estimate. The fixed-prior result reported in the main paper is therefore intentionally more conservative.

\paragraph{Motion-dependent behavior of the negative control.}
The mismatched control is intentionally difficult rather than arbitrarily incorrect: it uses another valid pose pair from the same trajectory and preserves the original temporal interval. For trajectories with smooth and approximately stationary relative motion, shifting the pose pair along time may still produce a similar relative transformation. This is particularly common for aerial and driving videos with sustained forward motion and limited rotation. In contrast, walking and stationary-panning sequences contain stronger temporal variation in motion direction or orientation and are therefore more sensitive to the shift.

\begin{table}[t]
    \centering
    \caption{\textbf{Mismatched-pose control by motion type on newly collected clips.} The control shifts the pose pair along the same trajectory while preserving the image pair and temporal interval. Lower median error and higher threshold accuracy indicate that the shifted relative pose remains similar to the true one.}
    \label{tab:epipolar_control_motion}
    \vspace{4pt}
    \resizebox{0.45\linewidth}{!}{%
    \begin{tabular}{lrr}
        \toprule
        Motion type & Median error $\downarrow$ & Clips below $5$\,px $\uparrow$ \\
        \midrule
        Drone & $1.05$\,px & $100.0\%$ \\
        Driving & $1.95$\,px & $84.8\%$ \\
        Train & $2.33$\,px & $80.0\%$ \\
        Walking & $3.96$\,px & $62.2\%$ \\
        Stationary pan & $8.41$\,px & $33.3\%$ \\
        \bottomrule
    \end{tabular}}
\end{table}

As shown in Table~\ref{tab:epipolar_control_motion}, the shifted control remains relatively competitive for drone and driving clips, where motion is often smooth and approximately uniform. Its error becomes substantially larger for walking and stationary-panning clips, whose relative motion varies more strongly over time. The aggregate control is therefore conservative: some temporally shifted pose pairs remain geometrically similar to the correctly aligned pair, reducing the apparent separation between the released poses and the negative control.

\paragraph{Scope.}
The epipolar test evaluates local relative orientation, translation direction, camera intrinsics, and pose--frame synchronization. It does not constrain translation magnitude and is therefore insensitive to monocular scale. Because it operates on frame pairs separated by $0.5$\,s, it also does not directly measure long-horizon accumulated drift. These complementary aspects are examined by the cross-run geometric-consistency evaluation below using $\mathrm{Sim}(3)$ alignment, rotation RPE, translation-direction error, and normalized ATE. Loop-closure refinement and full-accumulation results for panoramic reconstruction are discussed separately in Appendix~\ref{app:panoramic_reconstruction_cases}.

\subsection{Cross-Run Geometric Consistency}
\label{app:matched_pose_details}

\paragraph{Sampling and reconstruction.}
We examine whether the released ViPE trajectories preserve their relative geometric structure when the same videos are reconstructed in a separate DROID-SLAM run~\cite{teed2021droid}. We attempt reconstruction on $1{,}040$ perspective clips stratified by data source and camera-motion regime: $400$ inherited Sekai clips, $480$ clips from the New YouTube main crawl, and $160$ clips from the static supplement. DROID successfully reconstructs and pairs $1{,}000$ clips with the corresponding ViPE trajectories, comprising $395$, $480$, and $125$ clips from the three subsets, respectively. These $1{,}000$ successfully paired trajectories constitute the evaluation set used for the geometric error metrics in Table~\ref{tab:matched_pose_evaluation}. Panoramic sequences are excluded because the DROID configuration assumes pinhole imagery.

To ensure a consistent comparison across data sources, camera intrinsics are re-estimated directly from every input video using GeoCalib, and all subsets are reconstructed with the same DROID configuration, sampling stride, frame budget, random seed, and runtime settings. The diagnostic sample is designed to provide approximately balanced coverage of four motion regimes---static, straight, curved, and winding---so that motion-dependent behavior can be analyzed reliably.

The reconstruction-success values reported in Table~\ref{tab:matched_pose_evaluation} are source--motion weighted over the attempted clips to reflect the released corpus distribution. They therefore differ from the unweighted raw success fraction obtained by simply dividing the $1{,}000$ successfully paired trajectories by the $1{,}040$ attempted reconstructions.

\paragraph{Trajectory alignment and metrics.}
Because monocular reconstruction is defined only up to a global similarity transform, each DROID trajectory is aligned to the corresponding ViPE trajectory by solving
\begin{equation}
\label{eq:pose_sim3_alignment}
\min_{s,\mathbf{Q},\mathbf{b}}
\sum_{i=1}^{N}
\left\|
s\mathbf{Q}\mathbf{p}^{\mathrm{D}}_i+\mathbf{b}
-\mathbf{p}^{\mathrm{V}}_i
\right\|_2^2,
\end{equation}
where $s$ is a global scale factor, $\mathbf{Q}\in\mathrm{SO}(3)$ is a global rotation, and $\mathbf{b}\in\mathbb{R}^{3}$ is a translation. The $\mathrm{Sim}(3)$ alignment in Eq.~\eqref{eq:pose_sim3_alignment} removes global scale and coordinate-frame ambiguity while preserving differences in relative trajectory geometry.

We evaluate cross-run consistency using rotation relative pose error (RPE), translation-direction error, and absolute trajectory error (ATE). Rotation RPE and translation-direction error are evaluated at one- and five-second horizons whenever valid frame pairs are available. To make ATE comparable across trajectories with different spatial extents, we normalize it by the path length of the corresponding ViPE trajectory:
\begin{equation}
\label{eq:pose_normalized_ate}
\mathrm{ATE}_{\mathrm{norm}}
=
\frac{
\mathrm{ATE}\!\left(
\mathbf{p}^{\mathrm{D}},
\mathbf{p}^{\mathrm{V}}
\right)
}{
\displaystyle
\sum_{i=2}^{N}
\left\|
\mathbf{p}^{\mathrm{V}}_i
-
\mathbf{p}^{\mathrm{V}}_{i-1}
\right\|_2
}.
\end{equation}

Corpus-level source and aggregate error statistics are reported as source--motion-weighted per-clip medians. Confidence intervals in the analyses below are obtained through stratified bootstrap resampling over clips.

\paragraph{Motion-wise pose consistency.}
We next examine how cross-run consistency varies with camera-motion regime. Table~\ref{tab:matched_by_motion} reports unweighted statistics over the motion-balanced diagnostic set. Unlike the weighted corpus-level results in Table~\ref{tab:matched_pose_evaluation}, these values are intended to characterize the geometric difficulty of each motion regime directly.

\begin{table}[t]
    \centering
    \caption{\textbf{Cross-run pose consistency by motion regime.}
    Results are unweighted per-clip medians over the motion-balanced diagnostic set, with $95\%$ bootstrap intervals. Rotation RPE and translation-direction error are reported in degrees, while ATE is normalized by trajectory path length.}
    \label{tab:matched_by_motion}
    \vspace{5pt}
    \resizebox{0.72\linewidth}{!}{%
    \begin{tabular}{lrrrrr}
    \toprule
    Motion & $n$ & Rot. $1$s $\downarrow$ & Rot. $5$s $\downarrow$
    & T-dir. $1$s $\downarrow$ & ATE$_{\mathrm{norm}}$ $\downarrow$ \\
    \midrule
    Straight & $250$ & $0.084^\circ$ \tiny$[0.077,0.096]$ & $0.332^\circ$ \tiny$[0.282,0.398]$
    & $1.24^\circ$ \tiny$[0.99,1.55]$ & $0.0137$ \tiny$[0.0112,0.0171]$ \\
    Curved & $245$ & $0.140^\circ$ \tiny$[0.115,0.162]$ & $0.614^\circ$ \tiny$[0.523,0.695]$
    & $5.13^\circ$ \tiny$[4.15,6.09]$ & $0.0345$ \tiny$[0.0270,0.0417]$ \\
    Static & $260$ & $0.156^\circ$ \tiny$[0.130,0.180]$ & $0.475^\circ$ \tiny$[0.375,0.653]$
    & $18.02^\circ$ \tiny$[14.41,24.18]$ & $0.0240$ \tiny$[0.0194,0.0277]$ \\
    Winding & $245$ & $0.203^\circ$ \tiny$[0.183,0.243]$ & $0.830^\circ$ \tiny$[0.730,1.019]$
    & $7.67^\circ$ \tiny$[5.87,8.86]$ & $0.0344$ \tiny$[0.0265,0.0413]$ \\
    \bottomrule
    \end{tabular}}
\end{table}

\textbf{Straight traversal.}
Straight trajectories exhibit the strongest cross-run consistency across all three geometric metrics. Their median one-second rotation RPE is $0.084^\circ$, translation-direction error is $1.24^\circ$, and normalized ATE is $0.0137$. All straight clips reconstruct successfully across the three data sources. This regime is also the most prevalent in the released corpus, carrying $72.3\%$ of the source--motion sampling weight, which explains why the weighted corpus-level statistics lie close to the straight-motion results.

\textbf{Curved and winding trajectories.}
Pose disagreement increases as trajectories involve stronger directional changes. One-second rotation RPE rises from $0.084^\circ$ for straight traversal to $0.140^\circ$ for curved trajectories and $0.203^\circ$ for winding trajectories. At five seconds, the same ordering is preserved, with errors of $0.332^\circ$, $0.614^\circ$, and $0.830^\circ$, respectively. Translation-direction error similarly increases from $1.24^\circ$ to $5.13^\circ$ and $7.67^\circ$.

Curved and winding trajectories are also more sensitive to intrinsic calibration. As shown in Table~\ref{tab:matched_calibration_by_motion}, replacing the original intrinsics with GeoCalib estimates increases one-second rotation RPE by factors of $3.0$ and $3.2$ for curved and winding trajectories, compared with $2.1\times$ for straight traversal. This suggests that calibration differences become more consequential as reconstruction must resolve stronger coupled rotation and translation.

\textbf{Static and near-static trajectories.}
Static trajectories require particular care because translation direction becomes mathematically unstable as displacement approaches zero. The one-second translation-direction error reaches $18.02^\circ$ over all static clips and $65.91^\circ$ within the static-supplement static subset. These large angular values do not necessarily imply correspondingly large pose errors: when both translation vectors are very short, small positional perturbations can induce large changes in their directions.

Limited motion also reduces reconstruction robustness. At the five-second horizon, only $112$ of the $260$ static clips retain sufficient valid-displacement pairs for translation-direction evaluation. Reconstruction failures are likewise concentrated entirely in static-motion cells. As shown in Table~\ref{tab:matched_source_x_motion}, the static supplement reconstructs only $45$ of $80$ static clips ($56.2\%$), and inherited Sekai reconstructs $95$ of $100$, whereas every non-static source--motion combination achieves $100\%$ reconstruction success.

Rotation RPE remains better conditioned because it does not depend on translation magnitude. Static trajectories reach a median one-second rotation RPE of $0.156^\circ$, comparable to the other non-straight motion regimes. We therefore interpret translation-direction error on near-static clips primarily as a diagnostic of metric conditioning rather than as a direct indicator of pose quality.

\paragraph{Source--motion interaction.}
To separate source effects from motion effects, Table~\ref{tab:matched_source_x_motion} further decomposes one-second consistency and reconstruction success jointly by source and motion class. Across all three sources, straight trajectories consistently yield lower rotation and translation-direction errors than curved or winding trajectories. This shared ordering indicates that a substantial part of the observed variation arises from motion regime rather than source identity alone. The most pronounced source-specific behavior occurs in the static supplement, whose translation-direction errors remain elevated even outside the static class, whereas its rotation consistency remains comparable to the other subsets.

\begin{table}[t]
    \centering
    \caption{\textbf{Sensitivity to intrinsic calibration across motion regimes.}
    One-second unweighted medians compare the original and GeoCalib-matched settings for the inherited and static-supplement subsets. The main crawl is excluded because it uses GeoCalib intrinsics in both settings.}
    \label{tab:matched_calibration_by_motion}
    \vspace{5pt}
    \resizebox{0.72\linewidth}{!}{%
    \begin{tabular}{lrrrrrr}
    \toprule
    Motion & $n$ & Original Rot. & Matched Rot. & Ratio & Original T-dir. & Matched T-dir. \\
    \midrule
    Static    & $140$ & $0.042^\circ$ & $0.098^\circ$ & $2.3\times$ & $18.92^\circ$ & $30.06^\circ$ \\
    Straight  & $130$ & $0.033^\circ$ & $0.067^\circ$ & $2.1\times$ & $0.40^\circ$  & $1.02^\circ$ \\
    Curved    & $125$ & $0.040^\circ$ & $0.121^\circ$ & $3.0\times$ & $0.98^\circ$  & $5.09^\circ$ \\
    Winding   & $125$ & $0.063^\circ$ & $0.202^\circ$ & $3.2\times$ & $3.46^\circ$  & $8.38^\circ$ \\
    \bottomrule
    \end{tabular}}
\end{table}

\begin{table}[t]
    \centering
    \caption{\textbf{Reconstruction success and one-second pose consistency by source and motion regime.} Success is measured over attempted reconstructions. All observed reconstruction failures occur in static-motion cells.}
    \label{tab:matched_source_x_motion}
    \vspace{5pt}
    \resizebox{0.72\linewidth}{!}{%
    \begin{tabular}{llrrrr}
    \toprule
    Source & Motion & $n$/attempted & Success $\uparrow$ & Rot. $1$s $\downarrow$ & T-dir. $1$s $\downarrow$ \\
    \midrule
    \multirow{4}{*}{Inherited Sekai}
    & Static    & $95/100$  & $95.0\%$  & $0.092^\circ$ & $20.93^\circ$ \\
    & Straight  & $100/100$ & $100.0\%$ & $0.065^\circ$ & $0.71^\circ$ \\
    & Curved    & $100/100$ & $100.0\%$ & $0.122^\circ$ & $3.59^\circ$ \\
    & Winding   & $100/100$ & $100.0\%$ & $0.208^\circ$ & $7.36^\circ$ \\
    \midrule
    \multirow{4}{*}{New YouTube, main crawl}
    & Static    & $120/120$ & $100.0\%$ & $0.198^\circ$ & $10.66^\circ$ \\
    & Straight  & $120/120$ & $100.0\%$ & $0.120^\circ$ & $1.39^\circ$ \\
    & Curved    & $120/120$ & $100.0\%$ & $0.156^\circ$ & $5.31^\circ$ \\
    & Winding   & $120/120$ & $100.0\%$ & $0.206^\circ$ & $6.62^\circ$ \\
    \midrule
    \multirow{4}{*}{New YouTube, static supplement}
    & Static    & $45/80$   & $56.2\%$  & $0.127^\circ$ & $65.91^\circ$ \\
    & Straight  & $30/30$   & $100.0\%$ & $0.082^\circ$ & $8.08^\circ$ \\
    & Curved    & $25/25$   & $100.0\%$ & $0.111^\circ$ & $13.74^\circ$ \\
    & Winding   & $25/25$   & $100.0\%$ & $0.185^\circ$ & $16.59^\circ$ \\
    \bottomrule
    \end{tabular}}
\end{table}

\paragraph{Motion-dependent calibration sensitivity.}
Finally, we examine whether sensitivity to camera intrinsics itself depends on trajectory type. Table~\ref{tab:matched_calibration_by_motion} compares the original and GeoCalib-matched settings for the inherited and static-supplement subsets, whose intrinsics change between the two evaluations.

The calibration effect is smallest for straight traversal and strongest for curved and winding trajectories. This trend is consistent with intrinsic errors becoming more consequential when substantial camera rotation must be distinguished from image motion induced by translation. The motion-wise results therefore reinforce the matched-calibration analysis: intrinsic calibration contributes measurably to cross-run disagreement, but its effect depends on trajectory structure and does not by itself account for the full source-level gap.

\paragraph{Interpretation and scope.}
Taken together, these analyses clarify the corpus-level results in Table~\ref{tab:matched_pose_evaluation}. Straight traversal exhibits the strongest and most stable cross-run agreement and dominates the released-data distribution. Curved and winding trajectories remain geometrically consistent but are more challenging and more sensitive to intrinsic calibration. Near-static sequences form a special case in which translation-direction metrics become poorly conditioned and DROID reconstruction itself may fail because of insufficient parallax.

The matched-intrinsics protocol and source--motion weighting remove two important procedural confounds, but they do not control for all differences in capture conditions. The main crawl contains predominantly two-minute driving, aerial, rail, and wide-FOV footage, whereas inherited Sekai is dominated by one-minute walking-oriented clips. Residual source-level differences should therefore not be interpreted directly as differences in absolute pose quality.

Finally, ViPE itself employs a DROID-based geometric stack. The cross-run experiment consequently measures stability across reconstruction runs, calibration settings, and capture regimes rather than accuracy against independent metric ground truth. Together with the image-based epipolar test, it provides complementary evidence that the released trajectories are geometrically consistent with the observed videos and remain stable under controlled re-reconstruction.

\section{Detailed Caption Statistics and Lexical Analysis}
\label{app:caption_statistics}

\subsection{Annotation Volume and Temporal Density}
\label{app:caption_volume}

The released annotation corpus contains approximately $135$ million words across all clip-level and segment-level fields. After aggregating each clip-level annotation with all of its temporally grounded segment annotations, each released clip contains a median of $973$ words and a mean of $1{,}050$ words. The segment-level \texttt{full\_prompt} is derived from the factorized descriptions and therefore partially restates their content in a generation-ready form. Excluding these derived segment-level full prompts, the corpus still contains approximately $85$ million words, corresponding to a median of $620$ and a mean of $657$ words per clip. We report both totals because the \texttt{full\_prompt} fields are included in the release and can be consumed directly during model training.

The dataset contains $649{,}597$ temporally grounded segments, corresponding to an average of $5.04$ segments per released clip. Their median duration is $15$ seconds, providing substantially denser temporal supervision than a single clip-level caption. Segment boundaries are aligned with meaningful changes in scene content, illumination, surface type, or camera behavior, so the number and duration of segments adapt to the temporal structure of each video rather than following a fixed partition.

\subsection{Field-Wise Caption Lengths}
\label{app:caption_lengths}

Table~\ref{tab:caption_field_statistics} summarizes the clip-level annotation fields. The four factorized descriptions occupy complementary length ranges: \texttt{subject\_motion}, \texttt{environment\_motion}, \texttt{static\_scene}, and \texttt{camera\_description} contain averages of $19$, $26$, $32$, and $27$ words, respectively. The camera-free \texttt{short\_prompt} averages $18$ words, while the derived \texttt{full\_prompt} averages $97$ words. At the segment level, the corresponding full prompt contains $79.6$ words on average. The distributions are shown in Fig.~\ref{fig:scene_caption}b,c.

\begin{table}[t]
    \centering
    \small
    \setlength{\tabcolsep}{7pt}
    \caption{\textbf{Statistics of the clip-level annotation fields.} Mean length reports the average number of words per released clip, while total words are accumulated over the complete corpus.}
    \label{tab:caption_field_statistics}
    \begin{tabular}{lrr}
        \toprule
        Annotation field & Mean words & Total words \\
        \midrule
        \texttt{subject\_motion}       & $19$ & $2{,}450{,}161$ \\
        \texttt{environment\_motion}   & $26$ & $3{,}378{,}305$ \\
        \texttt{static\_scene}         & $32$ & $4{,}126{,}762$ \\
        \texttt{camera\_description}   & $27$ & $3{,}456{,}500$ \\
        \texttt{short\_prompt}         & $18$ & $2{,}387{,}222$ \\
        \texttt{full\_prompt}          & $97$ & $12{,}622{,}799$ \\
        \bottomrule
    \end{tabular}
\end{table}

The difference between the two training-ready prompts is intentional. The \texttt{full\_prompt} integrates static scene content, subject motion, environmental dynamics, and camera behavior into a detailed camera-aware narrative. In contrast, the \texttt{short\_prompt} removes camera-specific clauses and retains a compact description of the observable scene and its dynamics. This design supports both camera-conditioned training and settings in which camera information is dropped, replaced, or controlled separately.

\subsection{Vocabulary Growth and Description Uniqueness}
\label{app:caption_vocabulary}

Across the full-prompt corpus, we observe $67{,}358$ distinct words over approximately $39$ million content-word tokens. To characterize vocabulary growth, we fit Heaps' law,

\begin{equation}
    V(N)=K N^{\beta},
\end{equation}
where $N$ is the number of observed tokens and $V(N)$ is the resulting vocabulary size. The fitted exponent is $\beta\approx0.43$, indicating that the vocabulary continues to expand as additional annotations are included rather than saturating at a small set of recurring expressions.

The annotations also exhibit little exact repetition. Among the $649{,}593$ segment-level short prompts included in the uniqueness analysis, $648{,}711$ are distinct, corresponding to a uniqueness rate of $99.86\%$. Across the factorized description fields, the exact description-level duplicate rate is $0.02\%$. These results suggest that the annotation volume is not produced by repeatedly applying a small collection of templates, but instead reflects substantial lexical variation across videos and temporal segments.

\subsection{Lexical Profiles within the Hierarchical Caption Schema}
\label{app:caption_lexical_structure}

The word clouds in Fig.~\ref{fig:word_cloud} reveal a clear division of semantic roles across the annotation fields. The \texttt{subject\_motion} field emphasizes agents and actions, with frequent terms such as \emph{pedestrian}, \emph{walks}, \emph{forward}, and \emph{steadily}. The \texttt{environment\_motion} field instead focuses on independently moving entities and ambient dynamics, including \emph{vehicles}, \emph{cars}, \emph{lights}, and \emph{traffic}. The \texttt{static\_scene} field describes persistent geometry, materials, and appearance through terms such as \emph{buildings}, \emph{trees}, \emph{stone}, and \emph{paved}. Finally, \texttt{camera\_description} concentrates on viewpoint and acquisition characteristics, including \emph{first-person}, \emph{eye-level}, \emph{smooth}, and \emph{stabilized}.

The two prompt fields combine these components at different levels of detail. The \texttt{full\_prompt} draws jointly from scene, agent, environment, and camera vocabularies to form a detailed generation-ready description. The \texttt{short\_prompt} retains a more concise scene-centric vocabulary and excludes explicit camera clauses. The distinct high-frequency terms across fields provide qualitative evidence that the factorized schema separates complementary aspects of the video rather than producing redundant paraphrases. Quantitative measurements of field complementarity are reported in Sec.~\ref{sec:exp_caption}.

\begin{figure*}[t]
    \centering
    \includegraphics[width=0.98\linewidth]{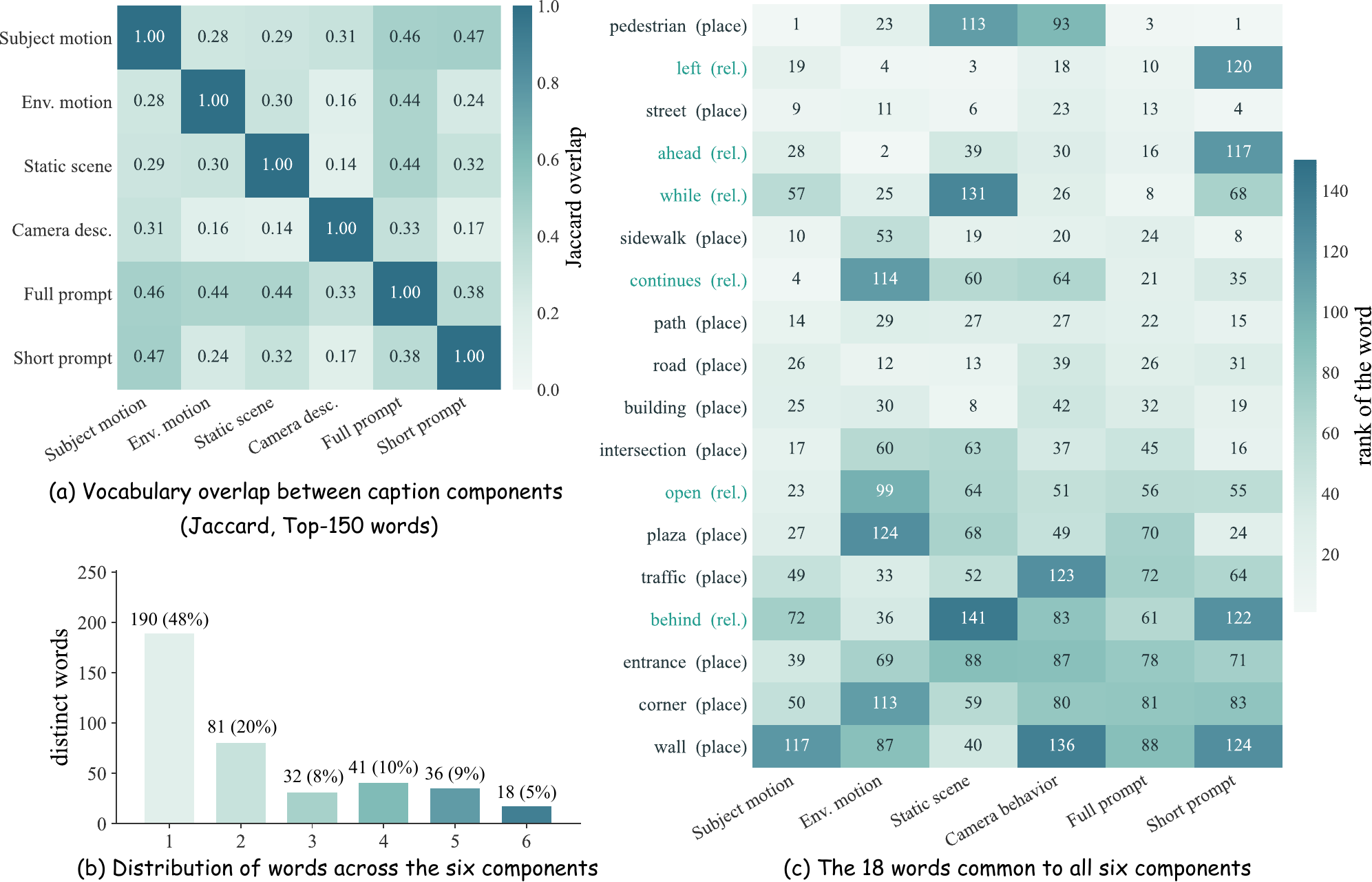}
    \vspace{4pt}
    \caption{\textbf{Vocabulary of the six segment-level caption components.} Panels follow Fig.~\ref{fig:vocab_components_clip}: (a) pairwise Jaccard overlap of the top-$150$ content words, (b) the number of components in which each word of the union appears, and (c) the words present in all six lists with their rank inside each component.}
    \label{fig:vocab_components_segment}
    \vspace{-5pt}
\end{figure*}

\subsection{Segment-Level Caption Vocabulary}
\label{app:vocab_segment}

Section~\ref{sec:exp_caption} measures component overlap at clip level; the same measurement at segment level is reported in Fig.~\ref{fig:vocab_components_segment}. The direction of the result is identical---the four factorized descriptions remain the most distinct group and the full prompt overlaps everything it restates---but the shared floor is wider: pairwise overlap among the four descriptions rises from $0.16$ to $0.25$, the union shrinks from $444$ to $398$ distinct words, the fraction occurring in a single component falls from $51\%$ to $48\%$, and the number of words common to all six grows from $10$ to $18$. The additional shared words are spatial relations and connectives (\emph{left}, \emph{ahead}, \emph{behind}, \emph{while}, \emph{continues}, \emph{open}) rather than new scene nouns, which follows from the annotation unit: consecutive segments re-describe one scene and must state where an action occurs relative to the preceding interval, whereas a clip-level description is written once. The full prompt again contributes only two words of its own at either level.

Overall, the annotations combine high textual density with fine temporal grounding and field-level semantic specialization. The corpus provides both structured factorized descriptions for disentangled supervision and derived prompts that can be used directly for generative training. Its vocabulary growth and low duplication rates further indicate that the released text captures diverse video content rather than relying on a limited set of recurring annotation templates.

\section{Qualitative Case Studies}
\label{app:case_studies}

This appendix consolidates qualitative examples of the released modalities and the panoramic trajectory-refinement pipeline. Separating these visual studies from the implementation details above makes it possible to reference each form of evidence directly from the corresponding method or analysis section.

\subsection{Visual Cases Across Data Sources}
\label{app:source_video_cases}

We begin with a source-level view of the videos themselves before examining their pose and annotation modalities. Starting from the $5$K+-hour Sekai collection, our quality pipeline retains $47{,}699$ clips ($795$ hours). We complement this inherited subset with $80{,}211$ newly curated perspective clips ($1{,}912$ hours), comprising $74{,}603$ clips from the main \dataset collection and $5{,}608$ recovered low-motion clips. Our self-captured panoramic collection initially contains $1{,}283$ videos (approximately $160$ hours), from which the released manifest retains $982$ videos ($119$ hours) with complete pose and caption annotations. Together, the three streams form the final release of $128{,}892$ clips totaling $2{,}826$ hours.

\begin{figure*}[t]
    \centering
    \includegraphics[width=\textwidth]{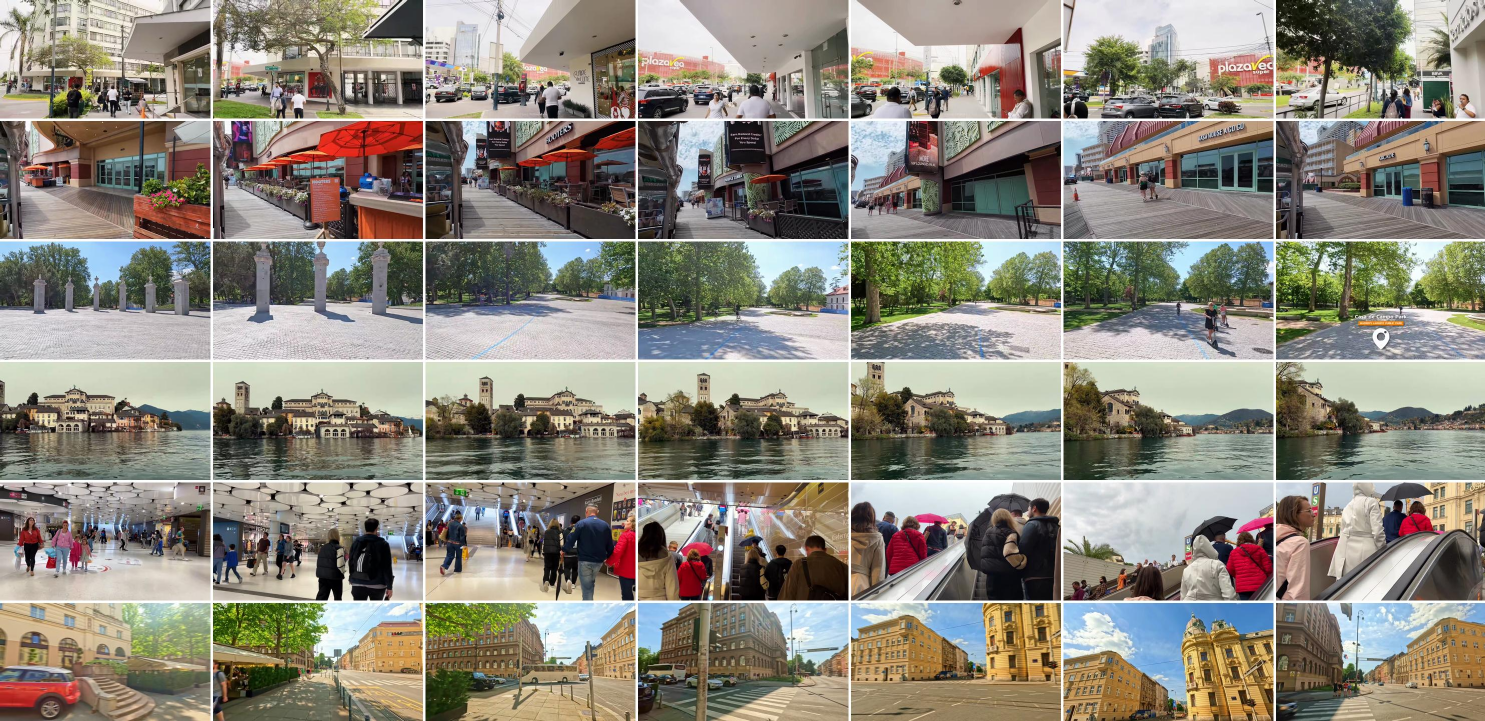}
    \caption{\textbf{Representative videos from the inherited Sekai subset.} Each row is one sequence observed at seven uniformly distributed relative timestamps. From top to bottom: walking in Lima, rail travel in Atlantic City, cycling in Madrid, boating in Orta San Giulio, escalator travel in Munich, and walking in Zagreb. The panel spans urban streets, public spaces, recreation, water, and transport environments.}
    \label{fig:sekai1_video_cases}
\end{figure*}

\begin{figure*}[t]
    \centering
    \includegraphics[width=\textwidth]{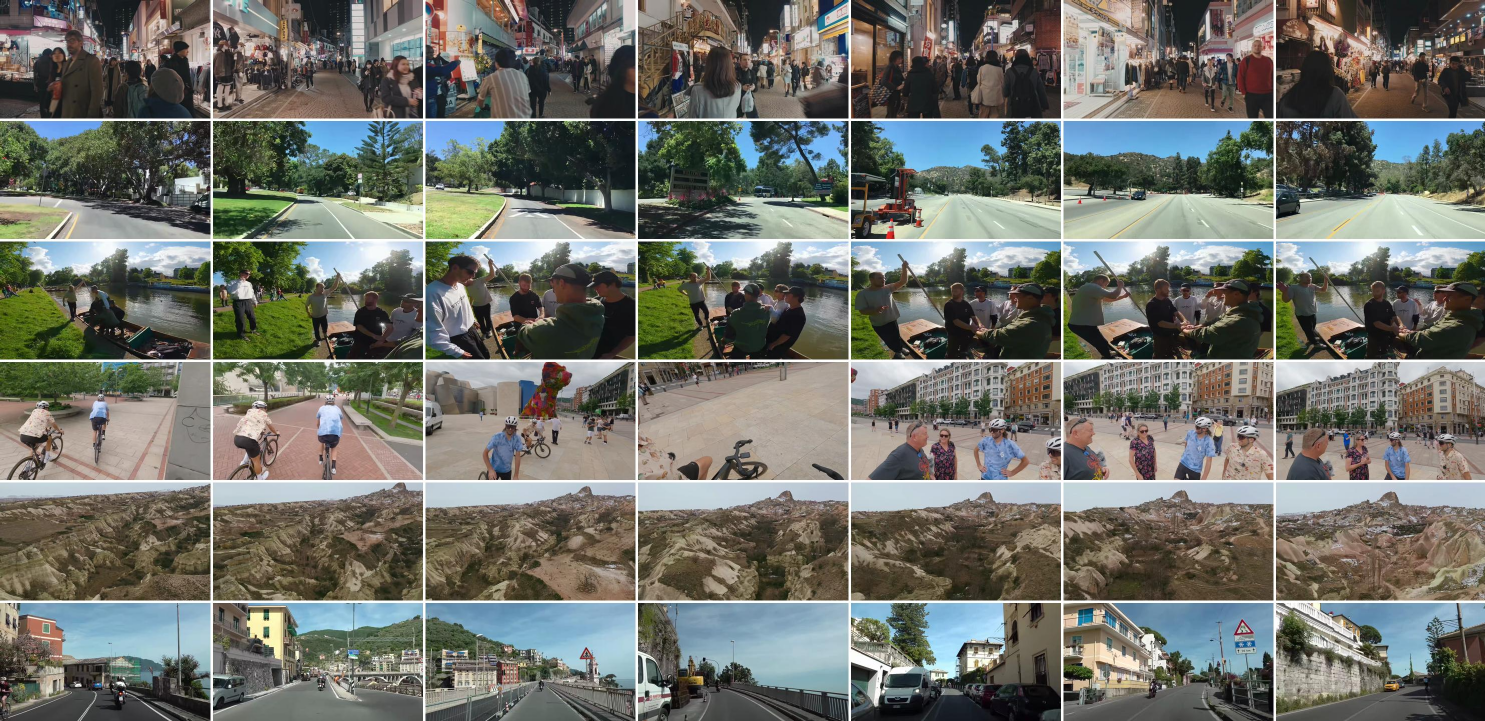}
    \caption{\textbf{Representative videos from the newly collected YouTube subset.} From top to bottom: night walking in Tokyo, driving in Los Angeles, boating in Cambridge, cycling in Bilbao, aerial capture over G\"oreme, and highway driving near Genoa. Seven frames per row summarize the evolution of each sequence over its complete duration.}
    \label{fig:sekai2_video_cases}
\end{figure*}

\begin{figure*}[t]
    \centering
    \includegraphics[width=\textwidth]{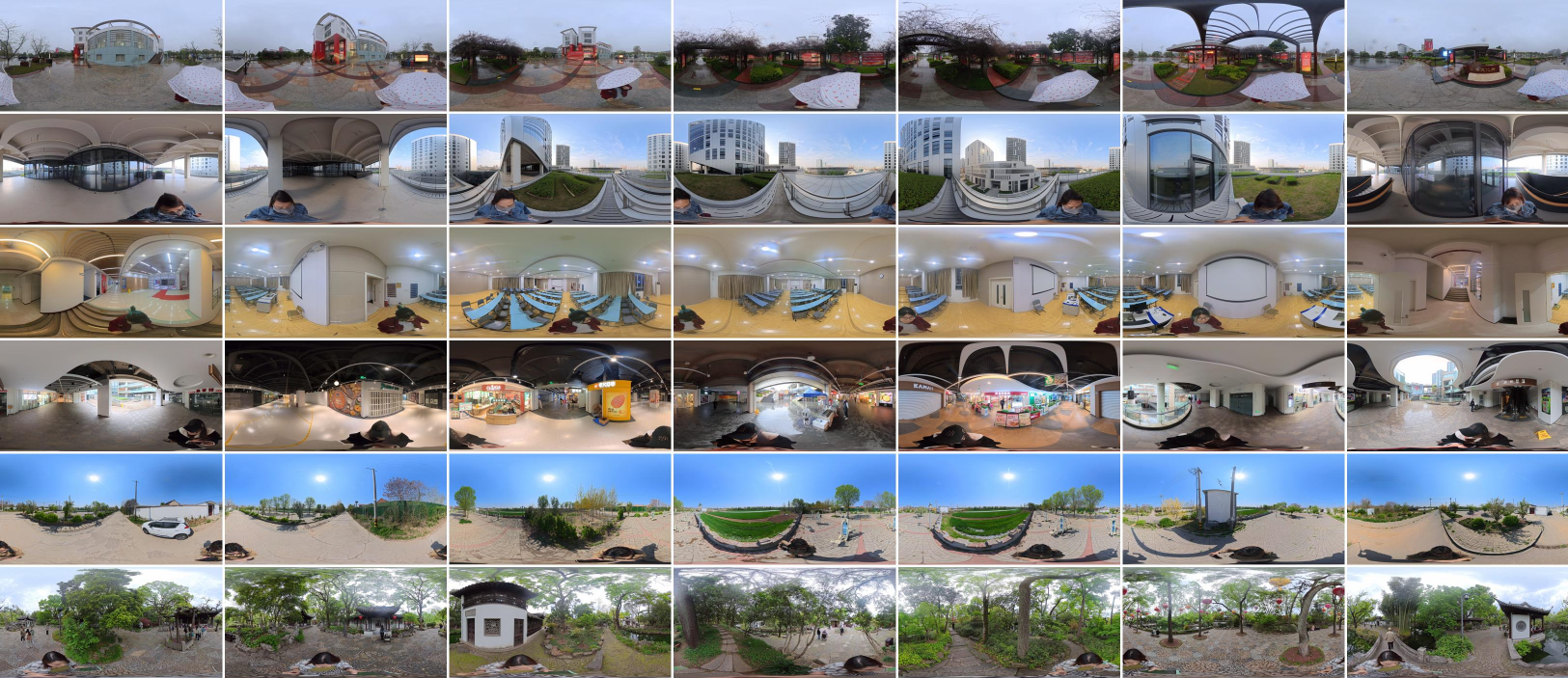}
    \caption{\textbf{Representative self-captured panoramic videos.} Frames preserve the native $2{:}1$ equirectangular field of view. From top to bottom, the sequences exhibit serpentine traversal, obstacle avoidance, stair traversal, a sharp turn, a second serpentine path, and a substantial elevation change. Complete azimuthal coverage keeps both the travel direction and surrounding context visible throughout each motion.}
    \label{fig:panoramic_video_cases}
\end{figure*}

Figures~\ref{fig:sekai1_video_cases}--\ref{fig:panoramic_video_cases} show six representative sequences from each of the three collection streams. Each row corresponds to one video and contains seven frames sampled at $4\%$, $19\%$, $35\%$, $50\%$, $66\%$, $82\%$, and $96\%$ of its duration. Sampling across the full sequence makes changes in viewpoint and scene layout visible, rather than reducing a long video to a single attractive keyframe.

The inherited Sekai cases span pedestrian, rail, cycling, boat, and escalator footage across several countries. The newly collected YouTube cases complement them with night walking, road travel, boating, cycling, and aerial capture in urban, natural, and transport environments. Finally, the panoramic cases retain their native $2{:}1$ equirectangular view and expose complete surroundings during non-linear motions, including serpentine traversal, obstacle avoidance, stairs, sharp turns, and elevation changes. Together, the strips illustrate both appearance diversity and sustained temporal evolution across the sources; later case studies analyze the associated structured captions, pose trajectories, and panoramic reconstructions.

\subsection{Multimodal Long-Horizon Annotation}
\label{app:multimodal_case}

\begin{figure*}[t]
    \centering
    \includegraphics[width=0.86\textwidth]{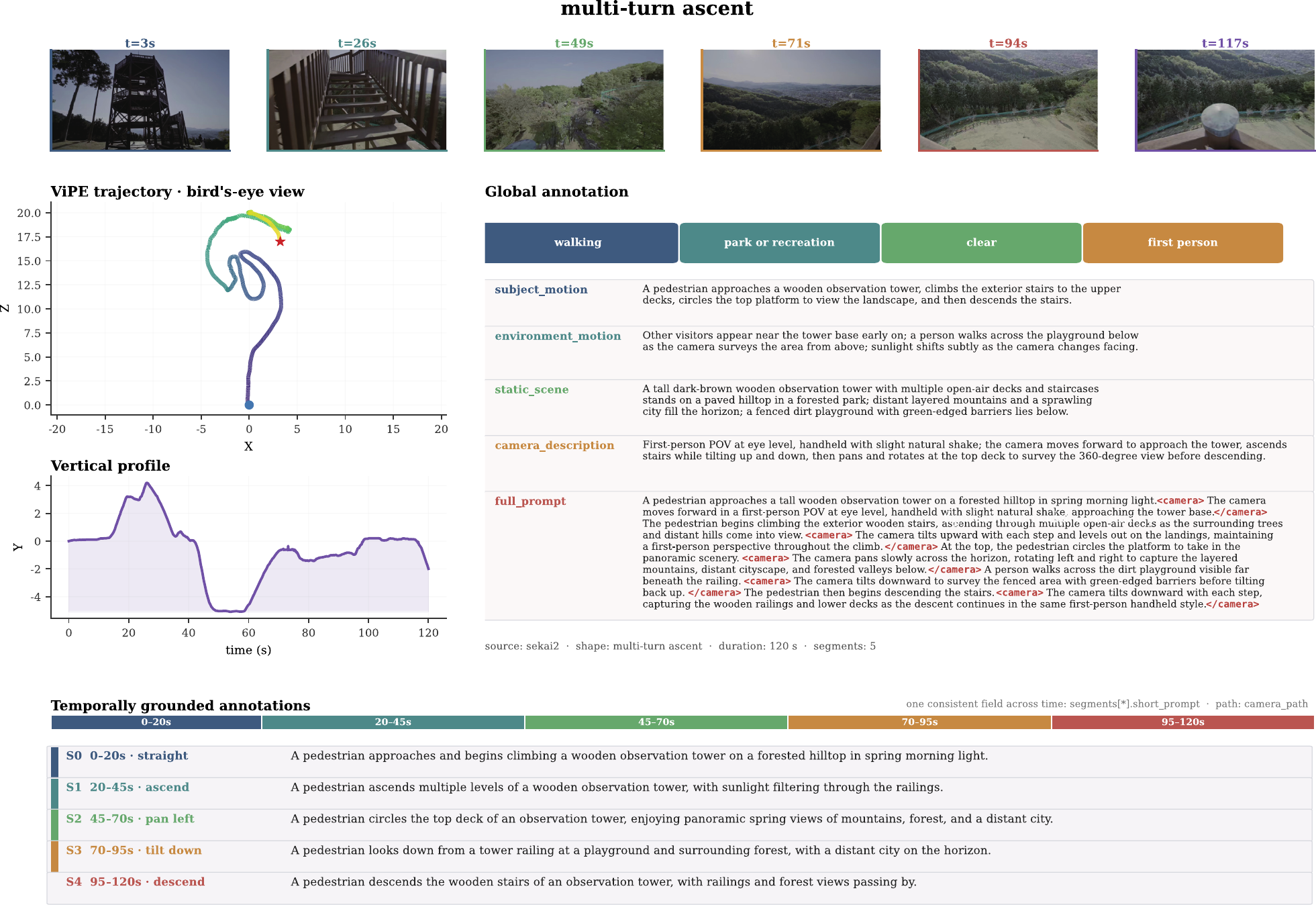}
    \vspace{4pt}
    \caption{\textbf{Video--pose--caption case study: multi-turn ascent (newly collected YouTube subset).}
    A $120$-second first-person ascent and descent of an observation tower. RGB frames, the ViPE trajectory and vertical profile, and hierarchical annotations align with camera-path transitions from \texttt{straight} to \texttt{ascend}, \texttt{pan\_left}, \texttt{tilt\_down}, and \texttt{descend}.}
    \label{fig:multimodal_case}
\end{figure*}

Fig.~\ref{fig:multimodal_case} presents a $120$-second first-person sequence in which a pedestrian approaches, ascends, surveys, and descends a wooden observation tower. Uniformly sampled frames capture the evolving visual context, while the ViPE trajectory and vertical profile jointly represent the horizontal and vertical components of the camera motion. Five temporally grounded segments align these geometric changes with the discrete camera paths \texttt{straight}, \texttt{ascend}, \texttt{pan\_left}, \texttt{tilt\_down}, and \texttt{descend}. The example illustrates why a single clip-level caption is insufficient: it can summarize the overall activity, but cannot precisely localize the successive motion phases and their associated scene changes.

\begin{figure*}[t]
    \centering
    \includegraphics[width=0.86\textwidth]{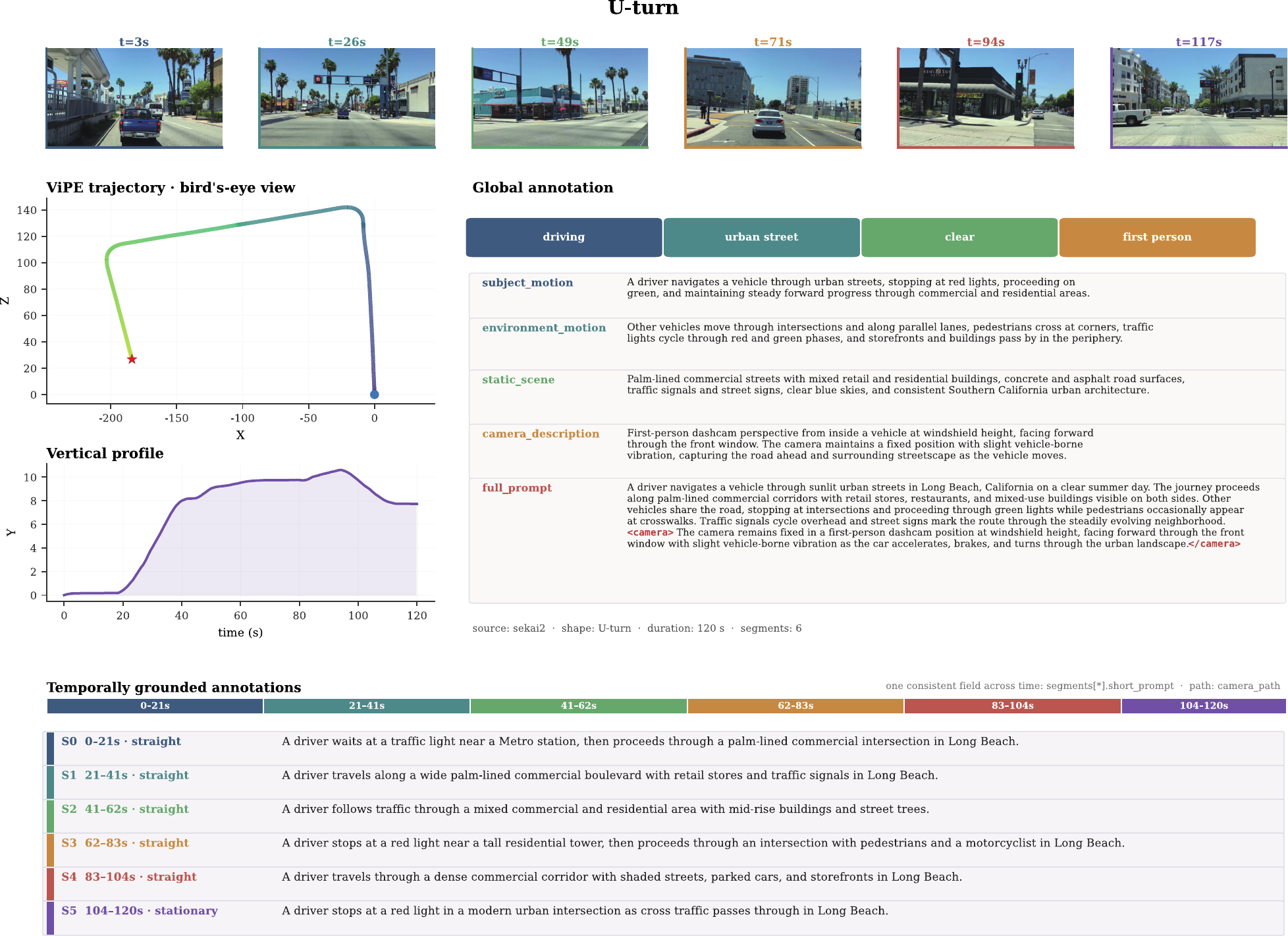}
    \vspace{4pt}
    \caption{\textbf{Case study: U-turn (newly collected YouTube subset).} A $120$s street-level walk with a mid-clip trajectory reversal. The six grounded segments switch from \texttt{straight} to \texttt{turn} and back, matching the heading change in the RGB row.}
    \label{fig:case_uturn}
\end{figure*}

\begin{figure*}[t]
    \centering
    \includegraphics[width=0.86\textwidth]{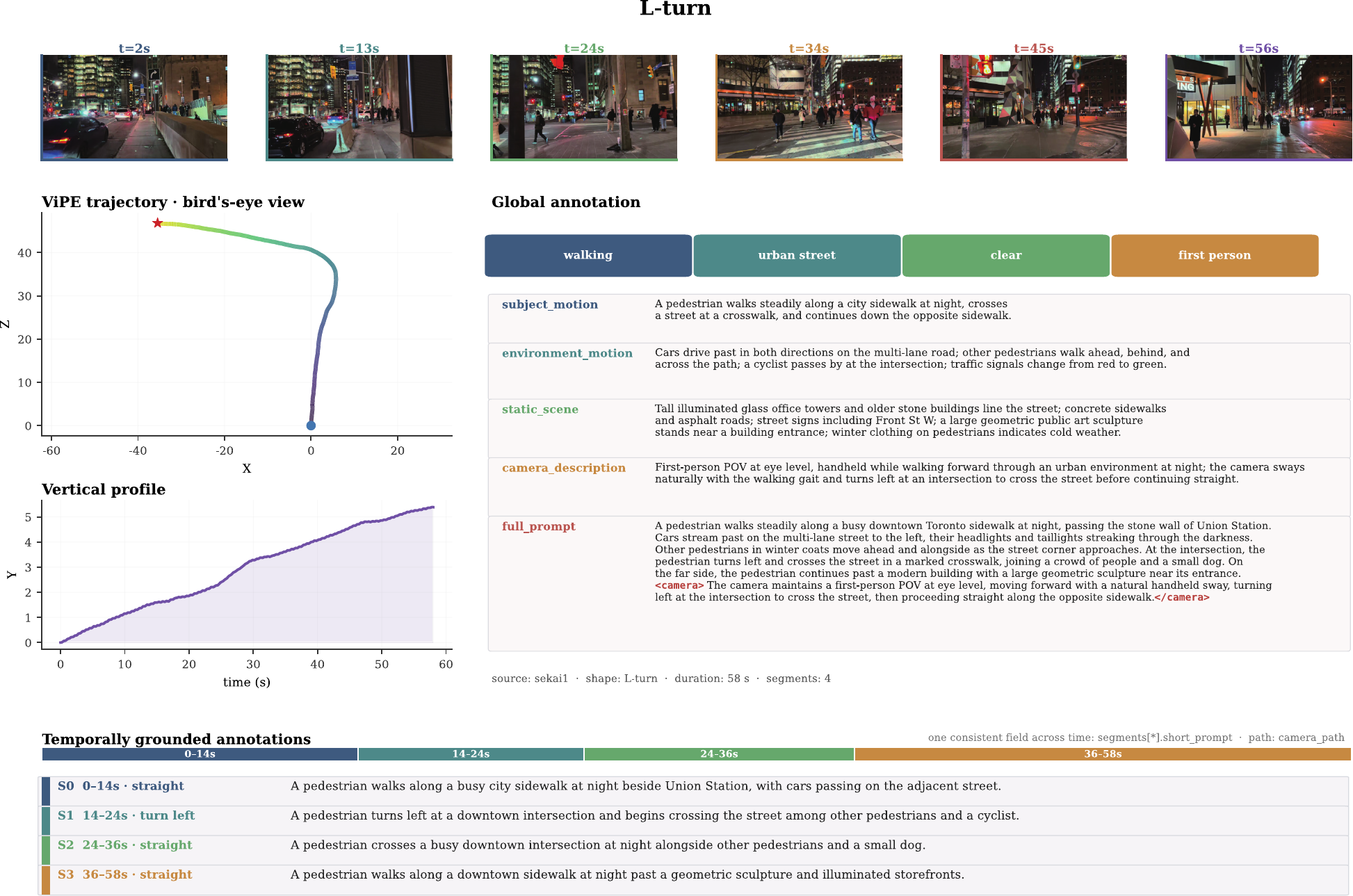}
    \vspace{4pt}
    \caption{\textbf{Case study: L-turn (inherited Sekai subset).} A $60$-second walking clip with two opposite bends. Inherited clips are half the length of the newly collected ones and are typically summarized by four segments rather than six to eight.}
    \label{fig:case_scurve}
\end{figure*}

\begin{figure*}[t]
    \centering
    \includegraphics[width=0.82\textwidth]{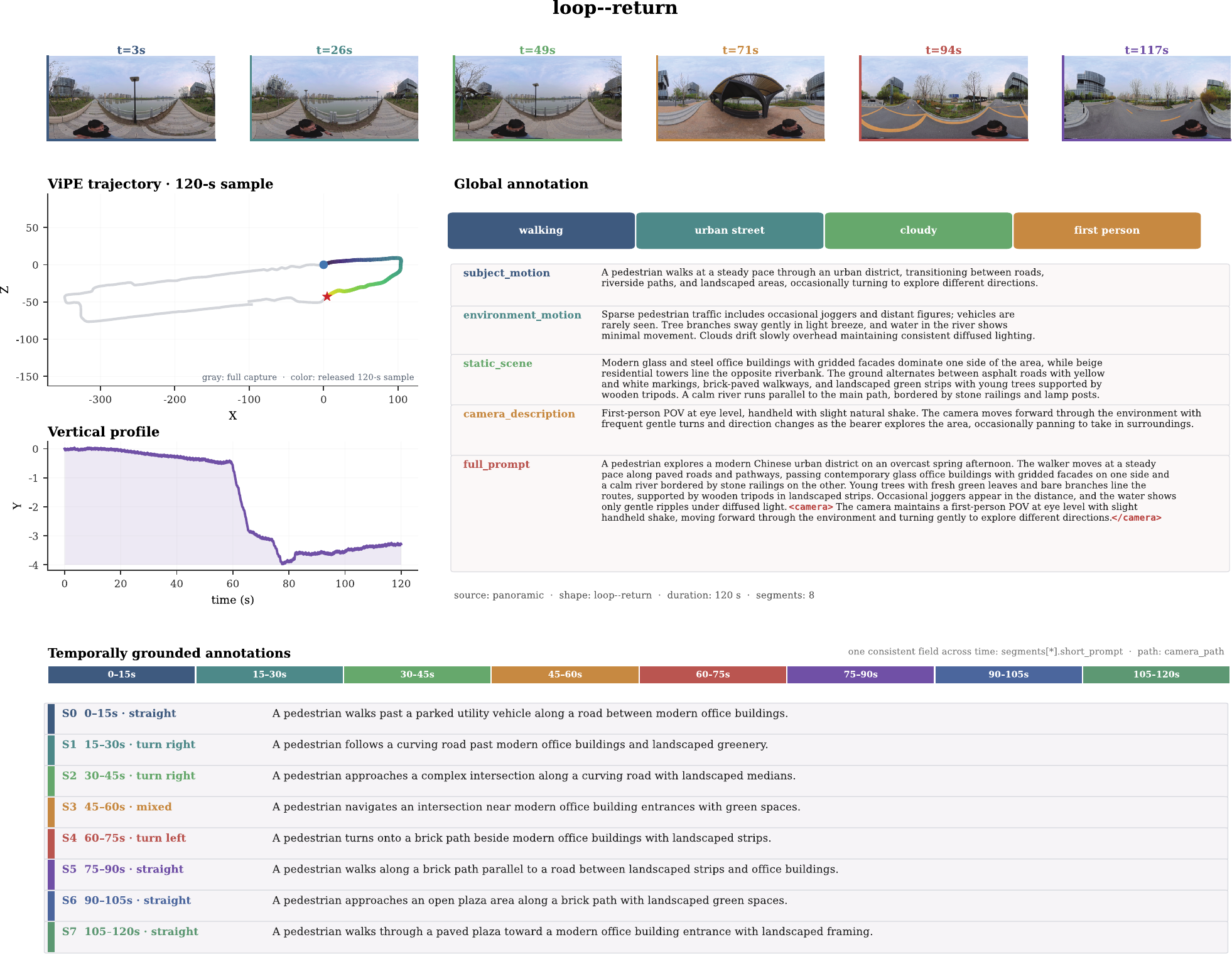}
    \vspace{4pt}
    \caption{\textbf{Case study: loop and return (self-captured panoramic subset).} A $120$-second window of a panoramic capture that returns to its starting area. Frames keep the native $2{:}1$ equirectangular field of view, and the eight grounded segments alternate between \texttt{straight} and \texttt{turn} along the route.}
    \label{fig:case_loop}
\end{figure*}

\begin{figure*}[t]
    \centering
    \includegraphics[width=0.82\textwidth]{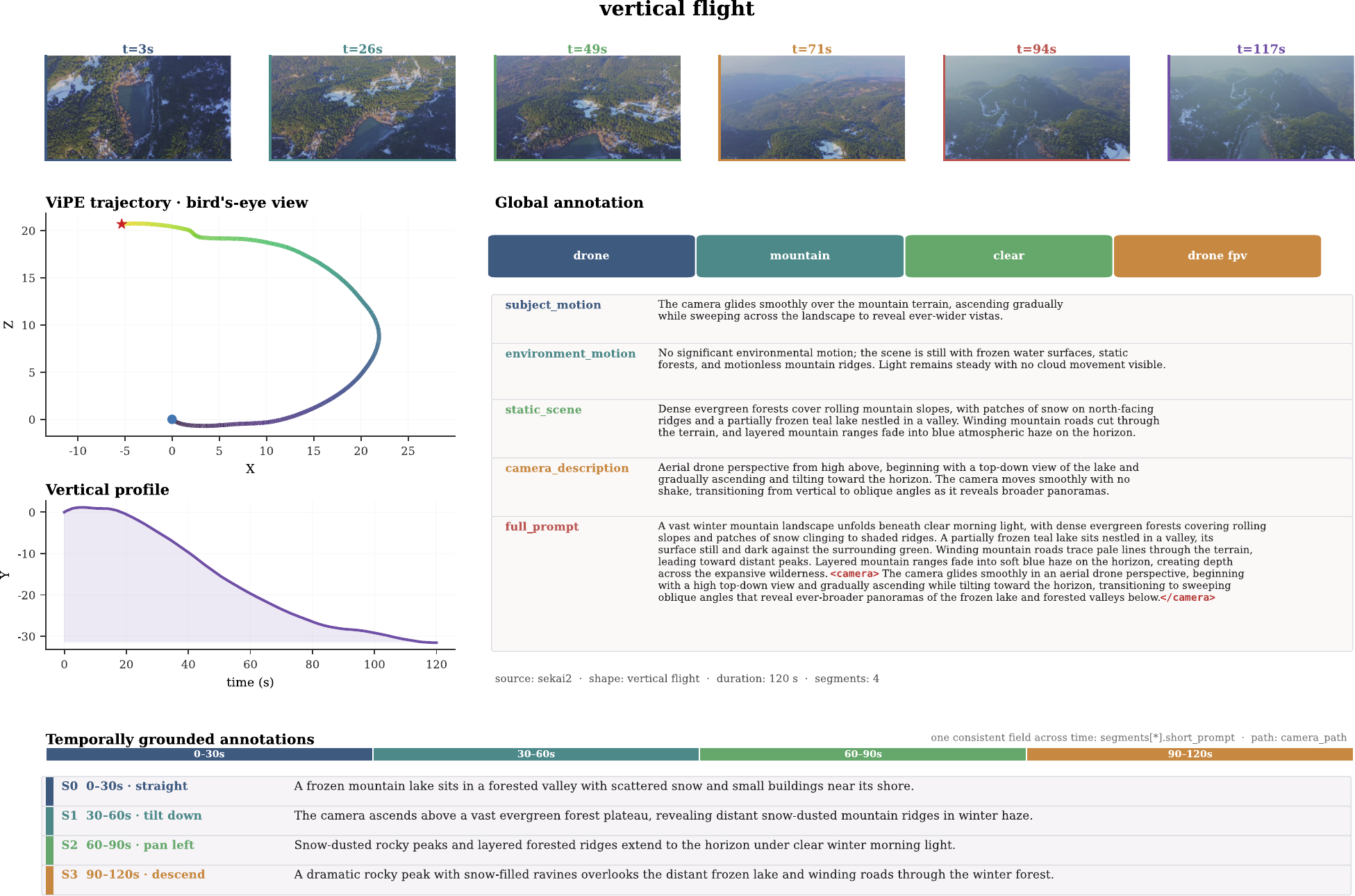}
    \vspace{4pt}
    \caption{\textbf{Case study: vertical flight (newly collected YouTube subset).} A $120$-second aerial sequence. The vertical profile carries most of the motion, which the ground-plane view alone would not reveal, and the scene description changes with altitude while the camera path remains predominantly straight.}
    \label{fig:case_vertical}
\end{figure*}

\begin{figure*}[t]
    \centering
    \includegraphics[width=0.86\textwidth]{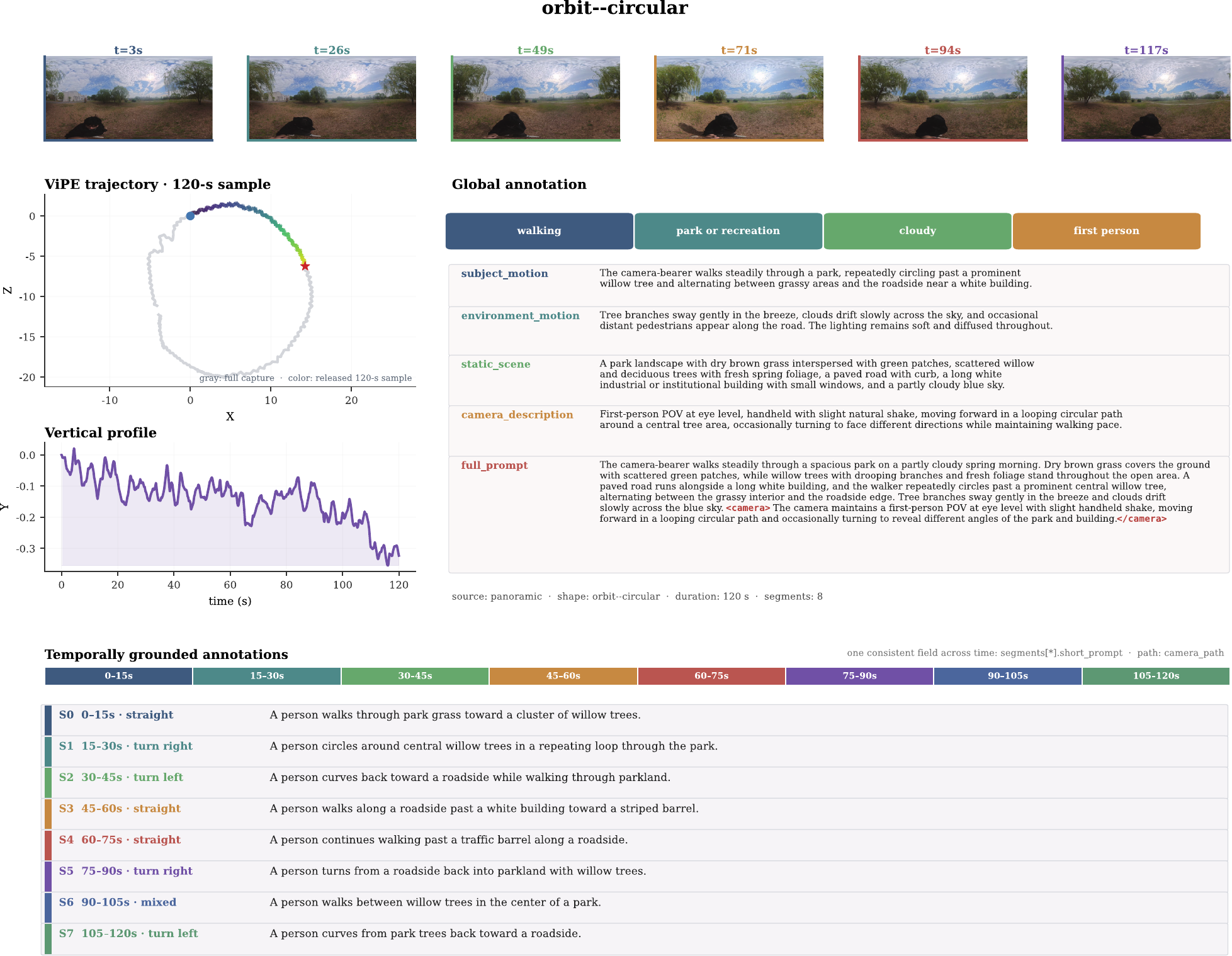}
    \vspace{4pt}
    \caption{\textbf{Case study: circular orbit (self-captured panoramic subset).} A $120$-second window in which the operator circles an open area. Complete azimuthal coverage keeps both the travel direction and the surrounding context visible throughout the motion, and the trajectory closes without a visible endpoint offset.}
    \label{fig:case_orbit}
\end{figure*}

Figures~\ref{fig:case_uturn}--\ref{fig:case_orbit} extend this view to five further clips, selected to span the three sources and clearly distinct trajectory shapes rather than to show favourable cases. Each panel uses the same layout: six RGB observations sampled across the clip, the ground-plane ViPE trajectory with its vertical profile, the four factorized global descriptions with the derived full prompt, and the temporally grounded segments with their \texttt{camera\_path} labels. Read together, the five examples show the property that motivates segment-level annotation: within a single clip the camera path changes several times, and the associated captions change with it. The U-turn and L-curve cases contain direction reversals that a clip-level caption would have to average away; the aerial case couples a large vertical displacement with a scene transition that the ground-plane view alone would not expose; and the two panoramic captures close their routes, so the same structures reappear at both ends of the timeline.

\subsection{Camera-Trajectory Cases Across Data Sources}
\label{app:pose_trajectory_cases}

\begin{figure*}[t]
    \centering
    \includegraphics[width=1.0\textwidth]{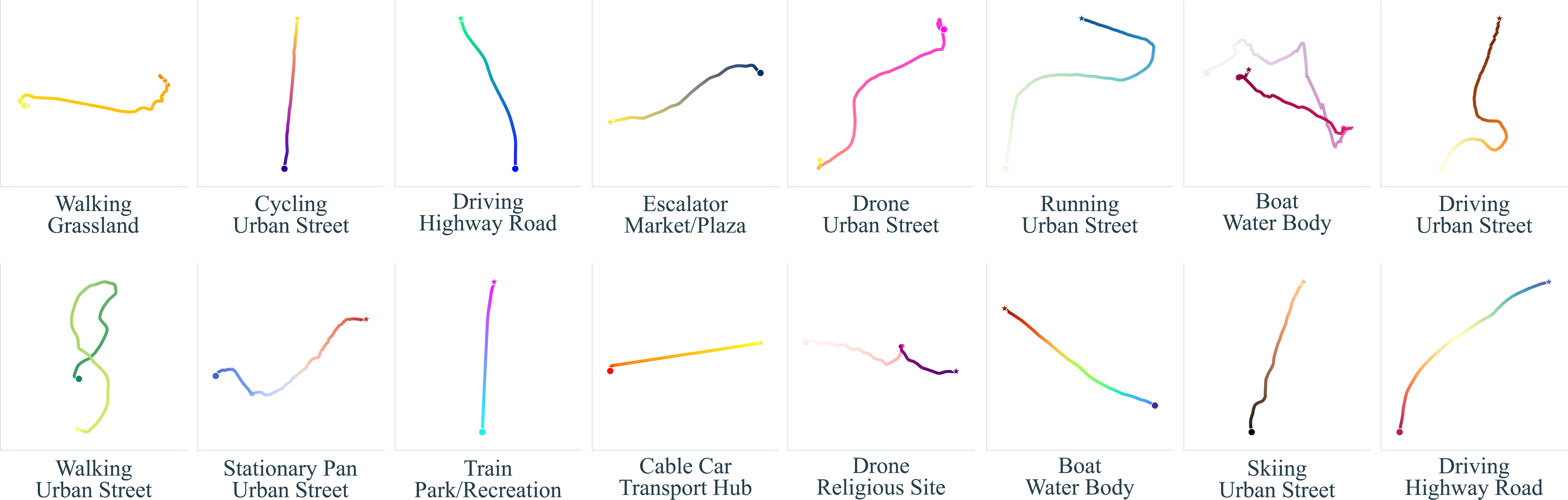}
    \vspace{4pt}
    \caption{\textbf{Representative camera trajectories from the inherited Sekai subset.} ViPE camera centers are projected onto the ground plane and colored by time. Circles and stars denote the trajectory start and end, respectively. The 20 deterministically selected cases emphasize clean but geometrically varied trajectories across the inherited Sekai subset.}
    \label{fig:sekai1_pose_cases}
\end{figure*}

\begin{figure*}[t]
    \centering
    \includegraphics[width=1.0\textwidth]{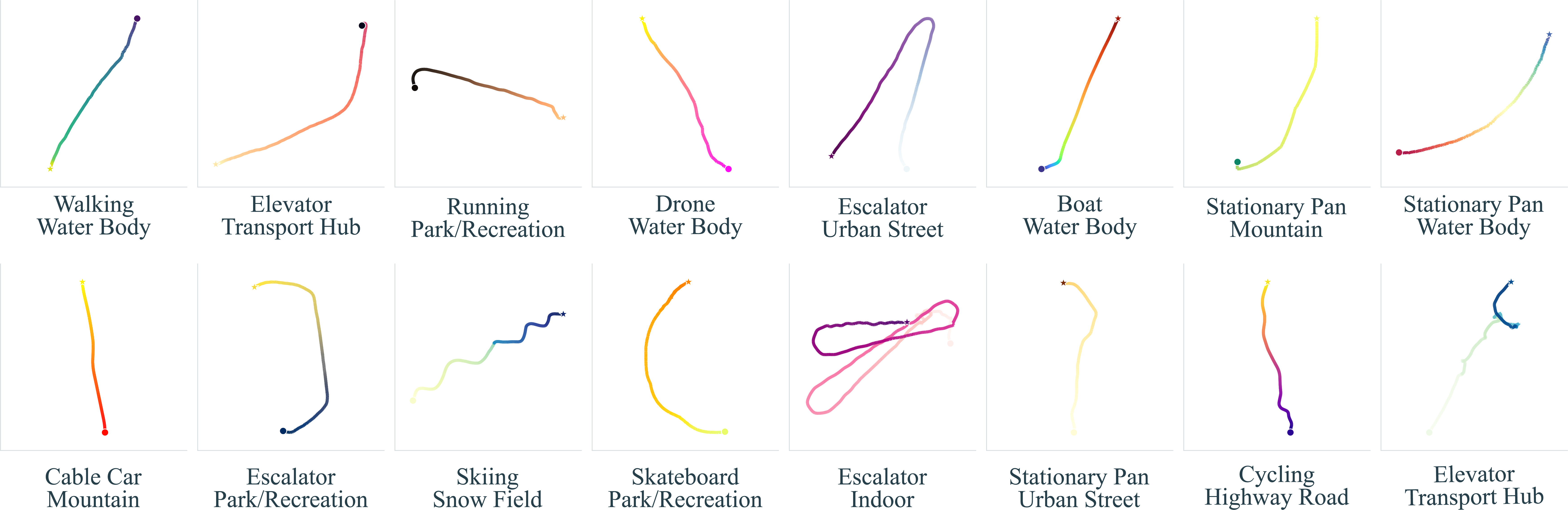}
    \vspace{4pt}
    \caption{\textbf{Representative camera trajectories from the newly collected YouTube subset.} This subset broadens both locomotion and camera regimes while retaining smooth pose tracks. The 20 cases are shown in the ViPE ground plane with temporal color progression, a circular start marker, and a star-shaped end marker.}
    \label{fig:sekai2_pose_cases}
\end{figure*}

Figures~\ref{fig:sekai1_pose_cases} and~\ref{fig:sekai2_pose_cases} present representative ViPE trajectories from the inherited Sekai subset and the newly collected YouTube subset, respectively. For each source, we deterministically select 20 cases after rejecting non-finite estimates, insufficient spatial extent, and trajectories dominated by isolated pose jumps. Farthest-point sampling over trajectory-geometry descriptors suppresses near-duplicates and prevents the gallery from being dominated by straight paths. The resulting cases cover walking, driving, rail travel, cycling, aerial motion, boats, cable cars, escalators, and other camera regimes. Despite this variation, the trajectories are temporally smooth and retain recognizable global structures such as long traversals, gradual bends, sharp turns, returns, and winding paths. The two source-specific panels therefore provide qualitative evidence for both pose continuity and geometric coverage; they complement, rather than replace, the quantitative trajectory-quality evaluation.

\subsection{Panoramic Full-Accumulation Reconstruction}
\label{app:panoramic_reconstruction_cases}

\begin{figure}[t]
    \centering
    \includegraphics[width=\textwidth]{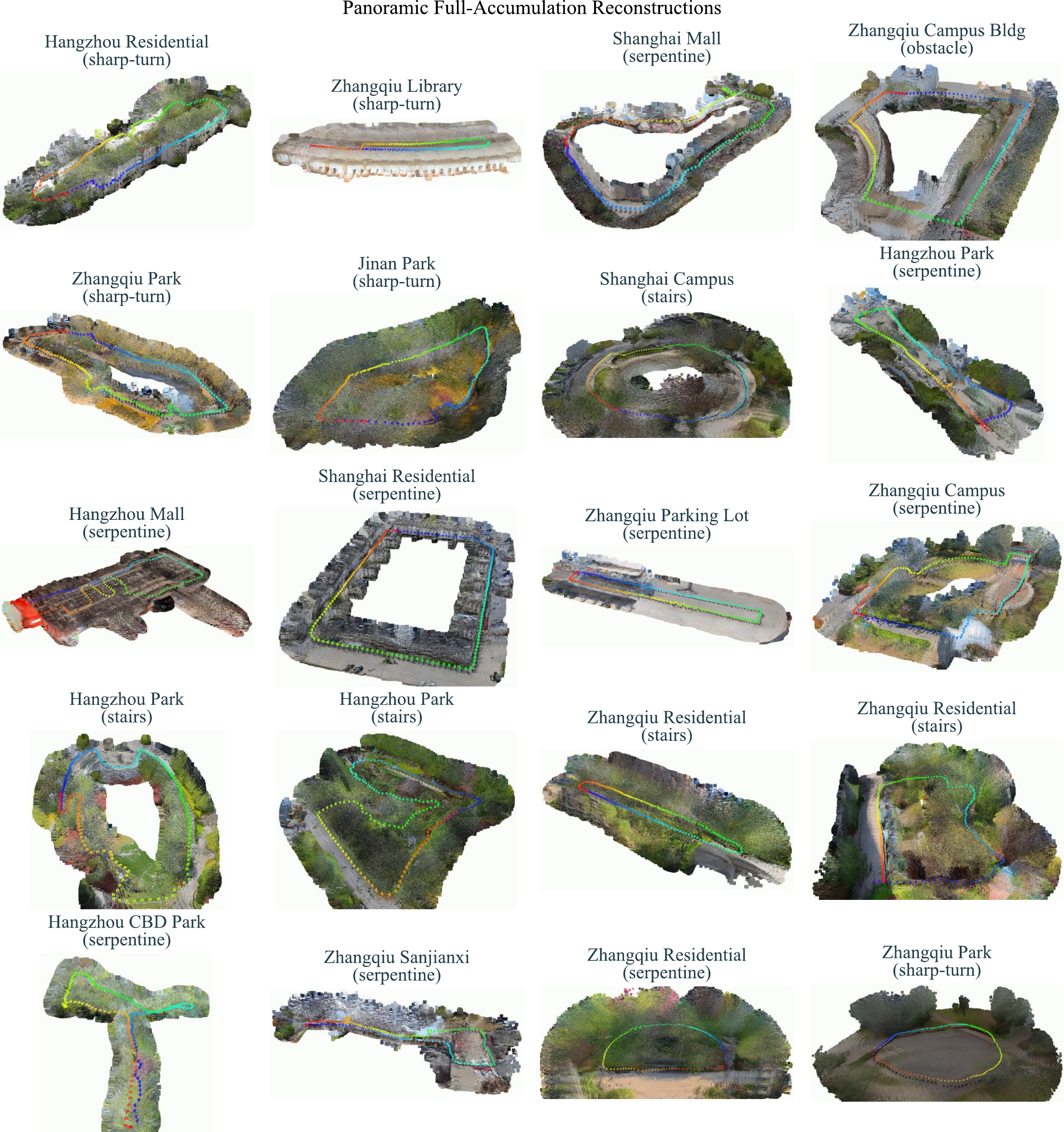}
    \vspace{4pt}
    \caption{\textbf{Panoramic full-accumulation reconstruction case studies (part I).} Each panel accumulates observations over an entire capture using the loop-refined camera poses; the colored curve shows the temporal camera trajectory. The examples span residential, campus, park, mall, and parking environments with sharp-turn, serpentine, stair, and obstacle-avoidance motion.}
    \label{fig:pano_full_accumulation_1}
\end{figure}

\begin{figure}[t]
    \centering
    \includegraphics[width=\textwidth]{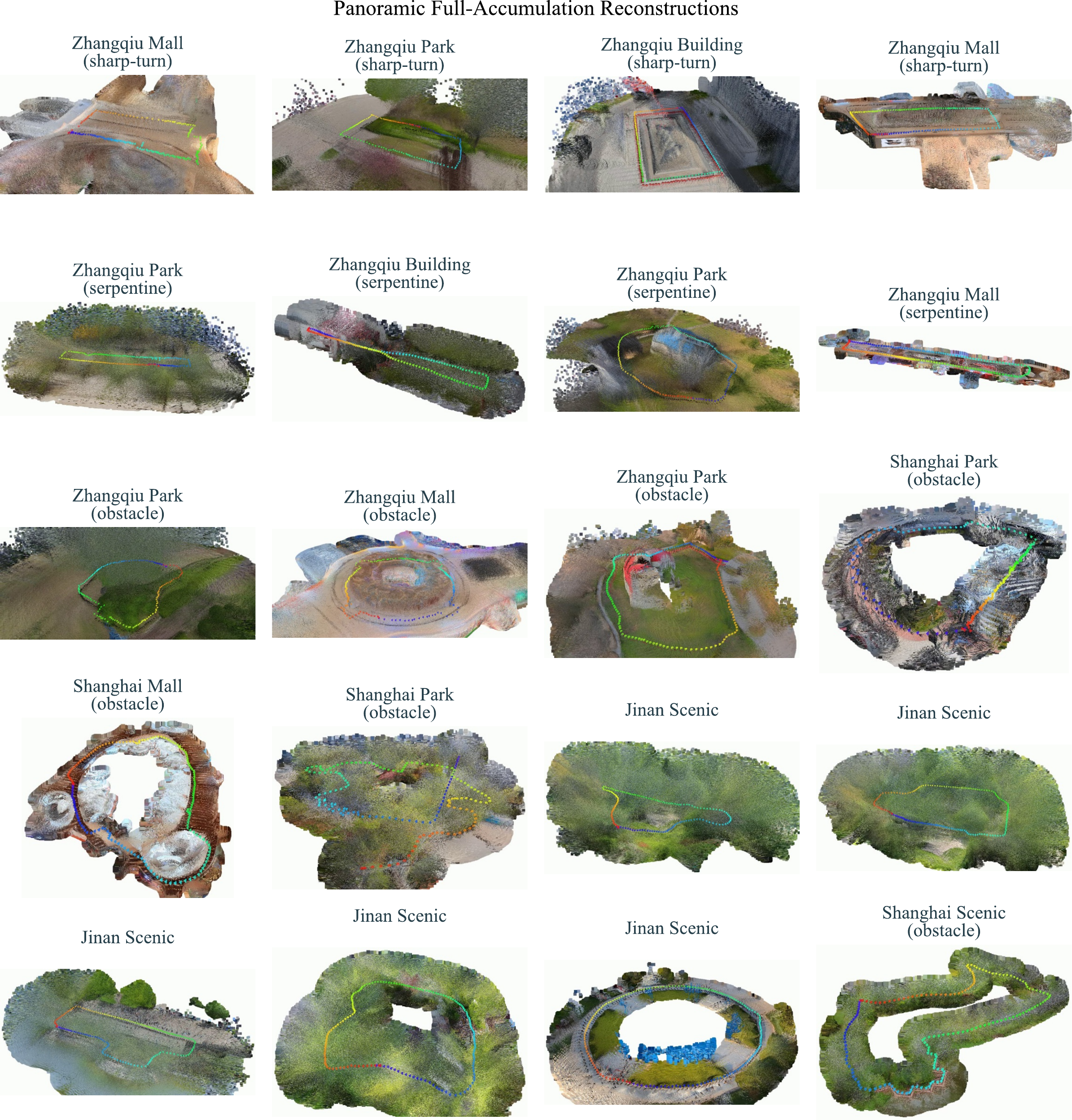}
    \vspace{4pt}
    \caption{\textbf{Panoramic full-accumulation reconstruction case studies (part II).} The additional examples demonstrate coherent long-range accumulation across buildings, parks, malls, and scenic areas, including strongly non-linear paths and repeated viewpoints.}
    \label{fig:pano_full_accumulation_2}
\end{figure}

Figures~\ref{fig:pano_full_accumulation_1} and~\ref{fig:pano_full_accumulation_2} visualize full-sequence reconstructions obtained by accumulating panoramic observations along the loop-refined trajectories. The examples cover indoor and outdoor environments and deliberately challenging motions, including sharp turns, serpentine paths, stair traversal, and obstacle avoidance. Across these settings, the accumulated geometry remains spatially coherent over extended trajectories: repeated structures align rather than forming visibly separated copies, and the estimated paths follow the reconstructed walkable regions while preserving their non-linear shape. Closed and near-closed paths are especially informative because small local pose errors would otherwise accumulate into conspicuous endpoint displacement and duplicated scene structure. These qualitative results complement the loop-verification statistics above by showing that the refinement is effective across diverse capture conditions, rather than only on a small set of simple circular walks. We emphasize that full accumulation is used here as a diagnostic of global trajectory consistency, not as an additional released 3D supervision signal.

\end{document}